\documentclass{article}

\usepackage{framed,multirow}
\usepackage[round]{natbib}
\usepackage{geometry}
\usepackage{authblk}

\usepackage{amssymb}
\usepackage{latexsym}
\usepackage{amsmath}
\usepackage{bbold}      
\usepackage[algoruled,linesnumbered]{algorithm2e}
\usepackage{diagbox}

\usepackage{url}
\usepackage{xcolor}
\usepackage{hyperref}

\usepackage{comment}    
\usepackage{makecell}   

\usepackage{tikz}
\usetikzlibrary{calc,fit,arrows,arrows.meta, backgrounds}
\pgfdeclarelayer{background}
\pgfdeclarelayer{foreground}
\pgfsetlayers{background,foreground}

\newcommand{\be}{\begin{equation}}
\newcommand{\ee}{\end{equation}}
\newcommand{\bc}{\begin{center}}
\newcommand{\ec}{\end{center}}
\newcommand{\bd}{\begin{description}}
\newcommand{\ed}{\end{description}}
\newcommand{\bi}{\begin{itemize}}
\newcommand{\ei}{\end{itemize}}
\newcommand{\pa}{\partial}
\newcommand{\bs}{\boldsymbol}
\newcommand{\bsl}{\bs{\lambda}}

\newcommand{\bu}{\bs{u}}
\newcommand{\bt}{\bs{\theta}}
\newcommand{\buf}{\bs{u}_f}
\newcommand{\buc}{\bs{u}_c}
\newcommand{\blf}{\bsl_f}
\newcommand{\blc}{\bsl_c}

\newcommand{\refeq}[1]{Equation (\ref{#1})}

\newcommand{\tx}{\text}

\newcommand{\tit}{\textit}

\newcommand{\tbf}{\textbf}
\newcommand{\bmat}{\begin{pmatrix}}
\newcommand{\emat}{\end{pmatrix}}
\newcommand{\bsmat}{\left(\begin{smallmatrix}}
\newcommand{\esmat}{\end{smallmatrix}\right)}
\newcommand{\bes}{\begin{equation}\begin{split}}
\newcommand{\ees}{\end{split}\end{equation}}

\newcommand{\revone}[1]{\textcolor{black}{#1}}
\newcommand{\revonemath}[1]{\textcolor{black}{#1}}
\newcommand{\revtwo}[1]{\textcolor{black}{{#1}}}
\newcommand{\revtwomath}[1]{\textcolor{black}{#1}}

\begin{document}
\title{A physics-aware, probabilistic machine learning framework for coarse-graining high-dimensional systems in the Small Data regime}
\author[1]{Constantin Grigo}
\author[1]{Phaedon-Stelios Koutsourelakis\footnote{Corresponding author \\ p.s.koutsourelakis@tum.de, constantin.grigo@tum.de}}
\affil[1]{Department of Mechanical Engineering, Technical University of Munich, Boltzmannstr. 15, 85748 Munich, Germany}

\maketitle

\begin{abstract}

The automated construction of coarse-grained  models  represents a pivotal component in computer simulation of physical systems and is a key enabler in various analysis and design  tasks related to uncertainty quantification.
Pertinent methods are severely inhibited by the high-dimension of the parametric input and the limited number of training input/output pairs that can be generated when computationally demanding forward models are considered. Such cases are frequently encountered in the  modeling of {\em random} heterogeneous media where the scale of  the microstructure necessitates   the use of high-dimensional random vectors and  very fine discretizations of the governing equations.
The present paper proposes a probabilistic  Machine Learning framework that is capable of operating in the presence of \textbf{ Small Data} by exploiting aspects of the physical structure of the problem as well as contextual knowledge. As a result, it can perform comparably well under {\em extrapolative} conditions.
 It unifies the tasks of dimensionality and model-order reduction through an encoder-decoder scheme that simultaneously  identifies a sparse set of  salient lower-dimensional microstructural features  and calibrates an inexpensive, coarse-grained model which is predictive of the output. Information loss is accounted for and quantified in the form of probabilistic   predictive estimates. The learning engine is based on Stochastic Variational Inference. We demonstrate how the variational objectives can  be used not only to train the coarse-grained model, but also to suggest refinements that lead to improved predictions.
 
 {\let\thefootnote\relax\footnote{\flushleft{Source code as well as examples can be found at:} \href{https://github.com/congriUQ/physics_aware_surrogate}{github.com/congriUQ/physics\_aware\_surrogate}}}
 
\end{abstract}



\section{Introduction}
The present paper is concerned with the data-driven construction of coarse-grained descriptions and predictive surrogate models in the context of random heterogeneous materials. 
This is of paramount importance for carrying out various uncertainty quantification (UQ) tasks, both in terms of forward and backward uncertainty propagation. More generally, the development of efficient surrogates is instrumental in enabling {\em many-query} applications in the presence of complex models of physical and engineering systems. 
The significant  progress that has been materialized in the fields of statistical or machine learning \cite{ghahramani_probabilistic_2015}  in the last few years, particularly in the context of Deep Learning \cite{LeCun2015,Goodfellow2016}  has found its way  in physical systems where training data for the surrogates are generated by forward simulations  of the reference, high-fidelity model.
 We note however that such problems exhibit (at least) two critical differences as  compared to typical machine learning applications.
Firstly, the dimension of the (random) inputs and outputs encountered in complex physical models is generally much larger than speech- or image-related data and associated tasks. Furthermore, the dependencies between these two sets of variables cannot be discovered with brute-force  dimensionality reduction techniques. Secondly and most importantly, these are not {\em Big Data} problems. The abundance of data that can be rightfully assumed in many data-driven contexts, is contrary to the nature of the problem we are trying to address, that is  learning surrogate  models, or more generally enabling UQ tasks, with the least possible high-fidelity output data. Such regimes could be more appropriately be referred to as {\em Small Data} or better yet {\em Tall Data} in consideration of the dimensionality  above \cite{koutsourelakis_special_2016}.

The development of  surrogates for UQ purposes in the context of continuum thermodynamics where pertinent models are based on PDEs and ODEs has a long history and some of the most well-studied methods have been based on  (generalized) Polynomial Chaos expansions (gPC) \cite{Wiener1938}, \cite{Ghanem1991}, \cite{Xiu2002} which have become popular in the last decade also due to the emergence of non-intrusive,  sparse-grid stochastic collocation approaches \cite{Xiu2005, Ma2009, Lin2009}. Despite recent progress in connecting gPC with domain decomposition methods \cite{Lin2010, Tipireddy2017, Tipireddy2018}, these approaches typically struggle in problems with high-dimensional stochastic input, as is the case with random heterogeneous materials   \cite{Torquato1993} encountered in problems such as  flow through  porous media \cite{Bilionis2013, Atkinson2019, Mo2018b} or electromagnetic wave propagation through random permittivity fields \cite{Chauviere2006}.

Another well-established strategy makes use of statistical couplings between the reference model and  less expensive, lower-fidelity models  in order to enable {\em multi-fidelity } methods   \cite{Kennedy2000}. In the context of PDEs, such lower-fidelity or reduced-order models may readily be constructed by utilizing potentially much coarser spatio-temporal discretizations which, in conjunction with appropriate statistical models, are capable of providing fully probabilistic predictions \cite{Koutsourelakis2009, Perdikaris_2017}. The  most striking benefit of such methods is that they automatically retain some of the underlying physics of the high-fidelity model and are therefore able to produce accurate predictions even when the data is scarce and the stochastic input dimensionality is high. Another category of algorithms falling under the class of reduced-order models is given by reduced-basis (RB) methods \cite{Hesthaven2016, Quarteroni2016} where, based on a small set of `snapshots' i.e. forward model evaluations, the solution space dimensionality is reduced by projection onto the principal `snapshot' directions. The reduced-order solution coefficients are then found either by standard Galerkin projection \cite{Rowley2004, Cui2015,afkham_structure_2017} or regression-based strategies \cite{Yu2018, Guo2018, Hesthaven2018}. Apart from issues of efficiency and stability,  both gPC and RB approaches in their standard form are generally treated in a non-Bayesian way and therefore only yield point estimates instead of full predictive posterior distributions.

A more recent trend is to view surrogate modeling as a supervised learning problem that can be handled by pertinent statistical learning tools, e.g. Gaussian Process (GP) regression \cite{Rasmussen2006, Bilionis2017} which provide closed-form predictive distributions by default. Although recent advances have been made towards multi-fidelity data fusion \cite{Kennedy2000, Raissi2017, Perdikaris2015} and incorporation of background physics \cite{Raissi2018, Yang2018, Lee2018, Tipireddy2018b} via Gaussian Processes, the poor scaling behavior with stochastic input dimension remains one of the main challenges for GP models.

More recently,  deep neural networks (DNNs) \cite{LeCun2015, Goodfellow2016} have found their way into surrogate modeling of complex computer codes \cite{Tripathy2018, Zhu2018, Mo2018b, Yang2019}. One of the most promising  developments in the integration of such tools in physical modeling are physics-informed neural networks \cite{Raissi2017b, Raissi2019, Tartakovsky2018} which are trained by minimization of a loss function augmented by the residuals of the governing equations \cite{Raissi2017b} and \cite{raissi2017physics}. In \cite{Zhu2019}, this methodology is generalized and extended to cases without any labeled data, i.e. without computing the output of the reference, high-fidelity  model. However, such models transfer much of the numerical burden from the generation of training data to the repeated evaluation of the residual during model training. While promising results have been obtained, we note  that several challenges remain with regards to the incorporation of prior physical knowledge in the form of symmetries/invariances etc. into DNNs as well as their capabilities under extrapolative conditions.

In this paper, we present a novel, fully Bayesian strategy for surrogate modeling designed for problems characterized by
\begin{itemize}
    \item  high input dimension, i.e. the ``curse of dimensionality'' -- in the problems examined, we consider input data (random material microstructures) of effective dimensionality $\gtrsim 10^4$; and
    \item  small number of training data $N$, i.e. forward or fine grained model (FGM) input/output pairs that can practicably be generated in order to train any sort of reduced-order model or surrogate -- in the examples presented here, we use $N \lesssim 100$ training samples. 
\end{itemize}
In the context of random media and in order to achieve the aforementioned objectives,  it is of primary importance (a) to extract only the microstructural features pertinent for reconstruction of the fine-grained model (FGM) output and (b) to construct a surrogate that retains as much as possible the underlying physics of the problem. Whilst point (a) should effectively reduce the input dimension, (b) may exclude a myriad of potential model outputs violating essential physical principles of the FGM. Consequently, the core unit of the surrogate model established here is based on a coarse-grained model (CGM) operating on simplified physics  valid on a larger scale than the characteristic length of the microstructural features in the FGM. The CGM serves as a stencil that automatically retains the primary physical characteristics of the FGM. It is parametrized by a latent, low-dimensional set of variables that locally encode the effective material properties of the underlying random heterogeneous microstructure and are found by appropriate sparsity-inducing statistical learning \cite{schoeberl_predictive_2017}. Finally, a decoder distribution maps the CGM outputs to the FGM solution space, yielding fully probabilistic model predictions. 
Instead of viewing the encoding step (dimension reduction) separately from the surrogate modeling task \cite{Ma2011, Xing2016}, both are integrated and trained simultaneously, ensuring that the encoded, latent representation of the microstructure is maximally predictive  of the sought FGM response instead of the original fine-scale microstructure itself. 
Model training is carried out using modern Stochastic Variational Inference (SVI) techniques \cite{Paisley2012, Hoffman2013} in combination with efficient stochastic optimization algorithms \cite{Kingma2014}. A hierarchical Bayesian treatment of the model parameters  obviates any need for fine-tuning or additional input by the analyst \cite{Bishop2000}.

The rest of the paper is organized as follows. Section \ref{sec:methodology} lays the methodological foundations for the proposed model, including a presentation of the pertinent physical structure, associated state variables and governing equations. 
Moreover, a detailed definition of both encoding and decoding steps is given, as well as a clear specification of the prior model and the training/prediction algorithms. In section \ref{sec:experiments}, we provide numerical evidence that the proposed model can lead to accurate predictions even when trained on a few tens of FGM data.  Furthermore, we present  a methodological framework for adaptively adjusting the CGM complexity in order to enhance its predictive ability.  Section \ref{sec:conclusion} summarizes the main findings and presents further extensions and possible applications of the proposed model.

\section{Methodology}
\label{sec:methodology}

\label{sec:model}
We present a general, probabilistic framework to coarse-graining which we specialize in the subsequent sections. The starting point is an expensive, deterministic, fine-grained model (FGM) with high-dimensional, random inputs $\blf$ and high-dimensional outputs $\buf=\buf(\blf)$. Due to the high dimensionality $\dim (\bs \lambda_f)$ of the input uncertainties $\bs \lambda_f$ and the small number of  forward evaluations  that can \revone{practically} be performed,  any naive regression approach, i.e. learning a direct map from $\bs \lambda_f$ to $\bs u_f$ is doomed to fail.

\begin{figure}[!t]
\centering
\includegraphics[width=\textwidth]{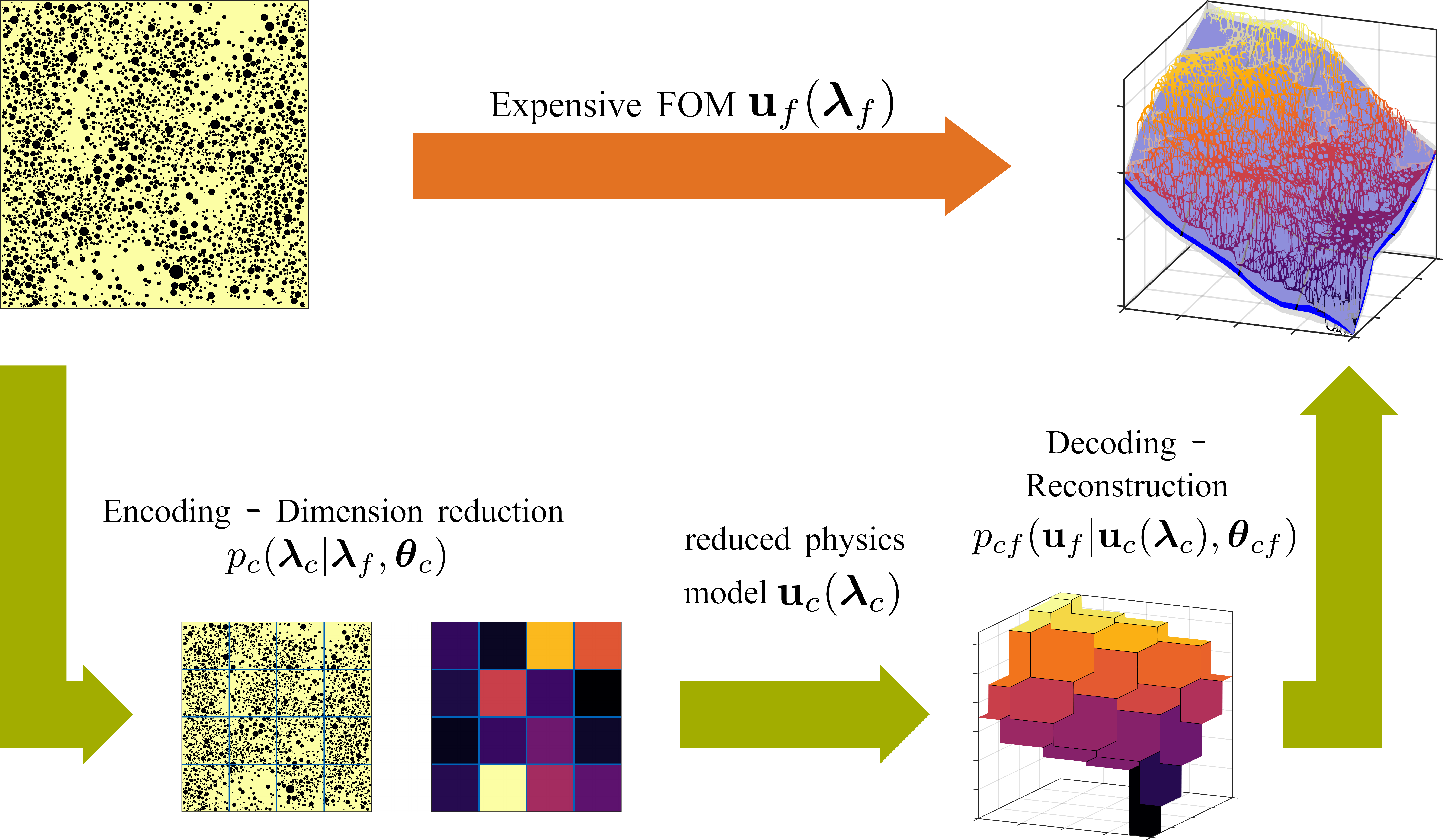}
\caption{Graphical illustration of the three-step process described in section \ref{sec:model}. Beginning from the top left: For the given microstructure $\bs \lambda_f$, a low-dimensional latent space representation $\bs \lambda_c$ is found via $p_c(\bs \lambda_c|\bs \lambda_f, \bs \theta_c)$. In the next step, $\bs \lambda_c$ serves as the input to a CGM based on simplified physics and/or coarser spatial resolution. The CGM output $\bs u_c(\bs \lambda_c)$ is then used to reconstruct the FGM solution using $p_{cf}(\bs u_f|\bs u_c(\bs \lambda_c), \bs \theta_{cf})$.}
\label{fig:overview}
\end{figure}

We seek instead a lower-dimensional, effective representation $\bs \lambda_c$ of the high-dimensional stochastic \revone{inputs} $\bs \lambda_f$ which encodes as much as possible information on the respective FGM output $\bs u_f$, see Figure \ref{fig:overview}. We note that this differs from classical dimension reduction approaches in the sense that the encoded representation $\bs \lambda_c$ should be maximally predictive of the FGM output $\bs u_f(\bs \lambda_f)$ -- and not for the reconstruction of $\bs \lambda_f$ itself \cite{Tishby2000}.
Rather than assigning an abstract, statistical meaning to $\blc$, we associate them with a less-expensive, physical,  coarse-grained  model (CGM). In particular, the latent variables $\blc$ represent the inputs of the CGM. We denote by $\buc = \bs u_c(\bs \lambda_c)$ the corresponding CGM outputs which we subsequently attempt to link with the FGM response $\bs u_f(\bs \lambda_f)$, see Figure  \ref{fig:overview}.  
 The CGM effectively represents the bottleneck through which information that the FGM input $\blf$  provides about the FGM output $\buf$ is squeezed. While several possibilities exist, independently of the final choice for the CGM, information loss will generally  take place during the coarse-graining process, which in turn results in predictive  uncertainty that we aim at quantifying.
Moreover, tools for comparing different CGMs as well as strategies of refining the CGM adaptively in order to improve predictive accuracy are developed in a later section.

The general framework described above can be effected by a model that integrates the following three key components  (see Figure \ref{fig:overview} and \cite{Grigo2017, Grigo2019}):
\begin{itemize}
    \item \textbf{Encoding - Dimension reduction}: A mapping from the high - dimensional microstructural description $\bs \lambda_f$ to a much lower-dimensional quantity $\bs \lambda_c$ retaining as much as possible information on the corresponding response $\bs u_f(\bs \lambda_f)$ -- this mapping is carried out probabilistically by the conditional distribution $p_c(\bs \lambda_c| \bs \lambda_f, \bs \theta_c)$ parametrized by $\bs \theta_c$.
    \item \textbf{Model reduction}: A skeleton of a coarse-grained model (CGM) that employs the dimension-reduced quantities $\bs \lambda_c$ as input and yields an output $\bs u_c$. Such a model can be based on  simplified physics and/or coarser spatial discretization. In our case, this is constructed by employing a coarse discretization of the Darcy-flow governing equations as it will be explained in detail later.  We denote by $p_{\tx{CGM}}(\bs{u}_c | \bs{\lambda}_c)$ the generally stochastic input-output map implied.
    \item \textbf{Decoding - Reconstruction}: A probabilistic mapping from the CGM output $\bs u_c$ back onto the original FGM output $\bs u_f$ -- this mapping is mediated by the distribution $p_{cf}(\bs u_f| \bs u_c, \bs \theta_{cf})$ parametrized by $\bs \theta_{cf}$.
\end{itemize}
The combination of the aforementioned three densities yields
\begin{equation}
\begin{split}
    p(\bs u_f| \bs \lambda_f, \bs \theta_{cf}, \bs \theta_c) &= \int \underbrace{p_{cf}(\bs u_f| \revonemath{\bs u_c}, \bs \theta_{cf})}_{\tx{decoder}} \underbrace{p_{\tx{CGM}}(\bs u_c|\bs \lambda_c, \bs \theta_c)}_{\tx{CGM}} \underbrace{p_c(\bs \lambda_c|\bs \lambda_f, \bs \theta_c)}_{\tx{encoder}} d\bs \lambda_c d\bs u_c.
    \label{eq:singleLikelihoodgen}
\end{split}
\end{equation}
Assuming a deterministic CGM, i.e. $p_{\tx{CGM}}(\bs u_c|\bs \lambda_c) = \delta(\bs u_c - \bs u_c(\bs \lambda_c))$, this simplifies to
\begin{equation}
\begin{split}
    p(\bs u_f| \bs \lambda_f, \bs \theta_{cf}, \bs \theta_c) &= \int p_{cf}(\bs u_f|\bs u_c(\bs \lambda_c), \bs \theta_{cf}) p_c(\bs \lambda_c|\bs \lambda_f, \bs \theta_c)d\bs \lambda_c.
    \label{eq:singleLikelihood}
\end{split}
\end{equation}
The resulting parametrized density can be viewed as the {\em likelihood}  of a FGM input/output pair $\{\bs \lambda_f,  \bs u_f\}$ and is thus the key quantity in  the proposed probabilistic surrogate. We note that the three constitutive densities are highly modular and can be adapted to the particulars of the FGM and the CGM adopted. In the following, the FGM consists of a Stokes-flow simulator through a random heterogeneous medium, see section \ref{sec:stokes}. The CGM is described by Darcy's equations of flow through a permeable medium and  is explained in section \ref{sec:darcy}. We present the essential components of the proposed model as well as the specialization of the variables and densities  above in section \ref{sec:components}. In section \ref{sec:modelTraining}, we present the learning engine that is capable of training the model with small data and in section \ref{sec:predictions}, we discuss how the trained model can be used to produce probabilistic predictions. We conclude this methodological section with a discussion on the numerical complexity (section \ref{sec:complexity}) as  well as a novel algorithm for the adaptive refinement of the coarse-grained model (section \ref{sec:aar}).

\subsection{The fine-grained model: Stokes flow through random porous media}
\label{sec:stokes}
\begin{figure}[!t]
\centering
\includegraphics[width=\textwidth]{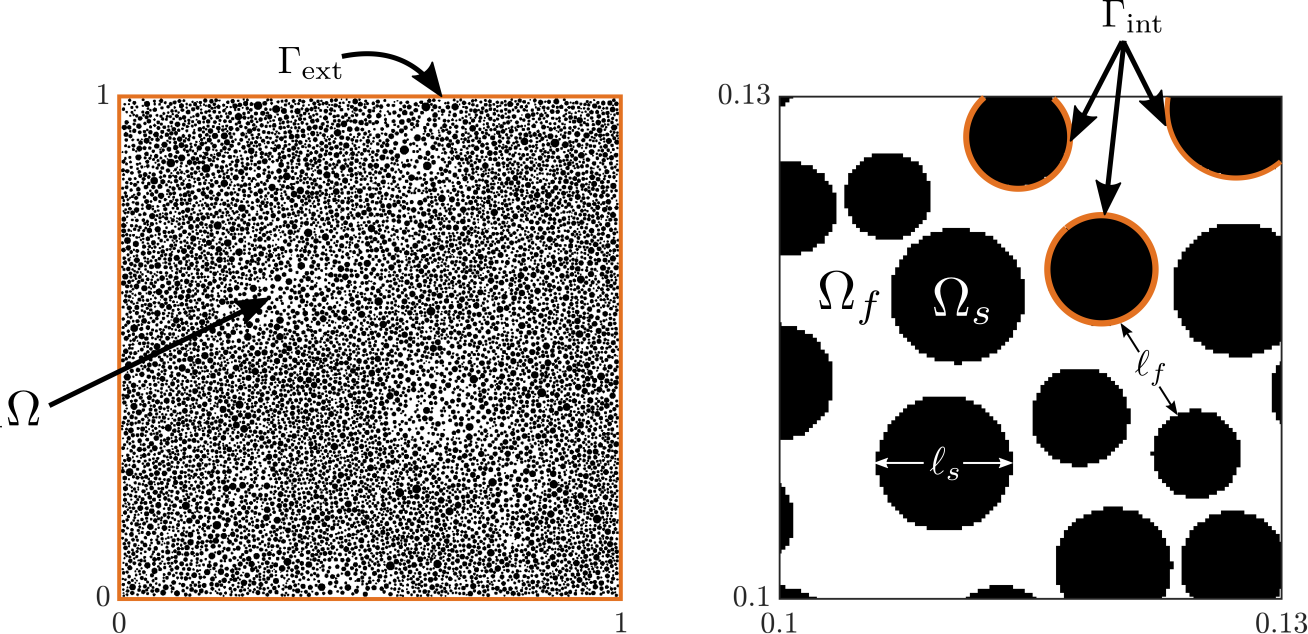}
\caption{Sample of a porous medium with randomly distributed, non-overlapping polydisperse spherical solid inclusions (black). Fluid can only flow in the pore space (white). The left side shows the whole unit square domain, whilst the right shows a zoomed segment for clarity.}
\label{fig:microstruct}
\end{figure}
Flow of incompressible Newtonian fluids through random porous media is characterized by low fluid velocities and is therefore dominated by viscous forces. In such regimes, the steady-state Navier-Stokes equations reduce to Stokes equations of momentum and mass conservation,
\begin{subequations}
\begin{align}
    \nabla_{\bs x} P - \mu \Delta_{\bs x} \bs V &= \bs f &&\text{for} \quad \bs x \in  \Omega_f, \\
    \nabla_{\bs x} \cdot \bs V &= 0 &&\text{for} \quad \bs x \in  \Omega_f,
\end{align}
\end{subequations}
where $P$ and $\bs V$ are pressure and velocity fields, $\bs f$ is an external force field, $\Omega_f = \Omega\backslash \Omega_s$ the pore part of the domain $\Omega = \Omega_f \cup \Omega_s$, where $\Omega_s$ denotes the part of impermeable solid material, and $\mu$ the fluid viscosity which we may set to $\mu = 1$ for convenience.

On the solid-fluid interface $\Gamma_{\tx{int}}$, the no-slip boundary condition
\addtocounter{equation}{-1}
\begin{subequations}
\addtocounter{equation}{2}
\begin{align}
    \bs V &= \bs 0 &&\text{for} \quad \bs x \in  \Gamma_{\tx{int}}
\end{align}
\end{subequations}
is imposed, whereas on the boundary $\Gamma_{\tx{ext}} = \pa \Omega = \pa \Omega_f \backslash \Gamma_{\tx{int}} = \Gamma_P \cup \Gamma_{\bs V}$ of the macroscopic system, we apply
\addtocounter{equation}{-1}
\begin{subequations}
\addtocounter{equation}{3}
\begin{align}
    \bs V &= \bs V_{bc}  &&\text{for} \quad \bs x \in  \Gamma_{\bs V}, \\
    \bs t &= (\nabla_{\bs x} \bs V_{bc} -  P_{bc} \bs I)\bs n &&\text{for} \quad \bs x \in  \Gamma_{P},
\end{align}
\label{eq:StokesPDE}
\end{subequations}
where $\bs t$ is the Cauchy traction and $\bs n$ the unit outward normal. 

For the remainder of the paper, we consider a unit square domain $\Omega = [0, 1]^2 \subset \mathbb R^2$ with randomly distributed non-overlapping polydisperse spherical exclusions, see Figure \ref{fig:microstruct} for a schematic representation. Such microstructures exhibit topologically fully connected pore spaces $\Omega_f$ and lead to unique solutions of \eqref{eq:StokesPDE} for the case where $\Gamma_{\tx{ext}} = \pa \Omega$. 

For the numerical solution of \refeq{eq:StokesPDE}, we employ triangular meshes with standard Taylor-Hood elements (i.e. quadratic and linear shape functions for velocity and pressure, respectively), implemented using the FEniCS finite element software package \cite{Fenics2012}. For generic two-phase media, $\bs \lambda_f$  consists of a list of binary variables representing the phase of each pixel/voxel.
For  the particular  case considered of randomly distributed non-overlapping polydisperse spherical exclusions, one could also represent the microstructure with a list of exclusion center coordinates and the corresponding radii. We denote by $\buf$ the discretized solution vector which can consist of pressures or velocities and write 
\begin{equation}
    {\bs u}_f(\bs \lambda_f) : \bs \lambda_f \mapsto {\bs u}_f
    \label{eq:FGM}
\end{equation}
 to indicate the map from the microstructure $\bs \lambda_f$ to the PDE response of interest.

\paragraph{Computational considerations}
Typically, the structure of the porous material is very complex with a characteristic length-scale $\ell$ much smaller than the macroscopic length scale $L = 1$ of the problem domain $\Omega$. This large difference in scales means that (a) from a modeling point of view,  a probabilistic description would be more realistic for  the  porous microstructure, and (b) the numerical solution of \eqref{eq:StokesPDE} requires discretizations fine enough to resolve all  microstructural details. Moreover, the uncertainty of microstructures $\bs \lambda_f \sim p(\bs \lambda_f)$ requires repeated  evaluation of the numerical solver \eqref{eq:FGM} to obtain accurate estimates (e.g. with Monte Carlo) of any statistic of interest relating to the output which quickly becomes impracticable due to the significant cost associated with each  FGM run.

\subsection{The coarse-grained model: Darcy equations for diffusion through a permeable medium}
\label{sec:darcy}
The macroscopic Darcy-like behavior of Stokes flow through random porous media \eqref{eq:StokesPDE} is a long-standing result from homogenization theory first shown for periodic microstructures \cite{Sanchez1980, Tartar1980}, later generalized for connected solid phase matrix \cite{Allaire1989}, the non-stationary process \cite{Allaire1992} and non-periodic media \cite{Whitaker1986}. In classical homogenization, the crucial step is to find a local boundary value problem on a Representative Volume Element (RVE) \cite{Sandstrom2013, Marchenko2006, Tartar2010}. The RVE size $r_0$, i.e. the radius of the averaging volume, must be large compared to the microstructural length scales $\ell_f, \ell_s$. On the other hand, the volume averaged quantities, i.e. the intrinsic phase averages $\bar P$ and $\bar{\bs V}$, should show sufficient variation over the problem domain $\Omega$, i.e. $L$ should be large compared to $r_0$. To summarize, Stokes to Darcy convergence is observed in the limit of
\begin{equation}
    \ell_f \ll r_0 \ll L
    \label{eq:lengthScales}
\end{equation}
which, according to \cite{Whitaker1986}, may slightly be relaxed to the conditions
\begin{equation}
    \left(\frac{r_0}{L} \right)^2 \ll 1, \qquad r_0 \gtrsim 5\ell_f
    \label{eq:lengthScales2}
\end{equation}
to safely assume Darcy-type flow. Using the definition for the intrinsic phase averages of pressure and velocity fields $P$ and $\bs V$
\begin{equation}
    \bar P = \frac{1}{|\Omega_\tx{f, RVE}|}\int_{\Omega_{f, \tx{RVE}}} P(\bs x) dV, \qquad \bar{\bs V} = \frac{1}{|\Omega_\tx{f, RVE}|}\int_{\Omega_{f, \tx{RVE}}} \bs V(\bs x) dV,
\end{equation}
where $|\Omega_{f, \tx{RVE}}|$ is the volume of the fluid part of the RVE, the effective Darcy constitutive behavior can be written as
\begin{subequations}
\begin{align}
\bar{\bs V} &= - \frac{\bs K}{\mu}(\nabla_{\bs x} \bar{P}) &&\tx{for} \quad \bs x \in \Omega, \\
\nabla_{\bs x} \cdot \bar{\bs V} &= \bs 0 &&\tx{for} \quad \bs x \in \Omega, \\
\bar{\bs V} \cdot \bs n &= \bs V_{bc} \cdot \bs n &&\tx{for} \quad \bs x \in \Gamma_{\bs V}, \\
\bar P &= P_{bc} &&\tx{for} \quad \bs x \in \Gamma_{P},
\end{align}
\label{eq:Darcy}
\end{subequations}
with the unit outward normal vector $\bs n$, the viscosity $\mu = 1$ and some unknown permeability tensor $\bs K$. Determining/estimating the effective permeability field $\bs K$ remains a substantial computational problem which in homogenization theory is usually approached by solving an RVE subscale problem \cite{Sandstrom2013}.
We emphasize that full Stokes/Darcy equivalence can only be assumed in the limit of infinite scale separation as defined by Equation \eqref{eq:lengthScales}. However, this does not rule out the possibility to use the Darcy equations  as a stencil of a machine learning model applied to fine-grained Stokes flow data even far  away from the scale separation limit, as we will see in the experiments section \ref{sec:experiments}.

\subsubsection{The Darcy permeability tensor field \texorpdfstring{$\bs K = \bs K(\bs x, \bs \lambda_c)$}{}}
\label{sec:DarcyK}

The material parameters of the CGM which provide the necessary closure pertain to the permeability tensor field $\bs K = \bs K(\bs x, \bs \lambda_c)$. We subdivide the problem domain $\Omega$ into  $N_{\tx{cells}, c} = \tx{dim}(\blc)$ of non-overlapping subregions/ cells $\left\{\Omega_m \right\}_{m=1}^{ N_{\tx{cells, c}} }$ such that $\cup_{m=1}^{ N_{\tx{cells, c}} } \Omega_m= \Omega$ and  $\Omega_{m_1} \cap \Omega_{m_2} = \emptyset$ for $m_1 \ne m_2$. In most cases examined in the experimental section \ref{sec:experiments}, the subregions $\Omega_m$ are squares of equal size, as can be seen in the bottom row of Figure \ref{fig:effPermeability}. Within each subregion $\Omega_m$, a constant permeability tensor $\bs K$ is assumed, i.e.
\begin{equation}
\bs K(\bs x, \bs \lambda_c) = \sum_{m= 1}^{N_{\tx{cells, c}}} \mathbb{1}_{\Omega_m}(\bs x) \bs K_m(\bs \lambda_c),
\label{eq:effPerm}
\end{equation}
where $\mathbb{1}_{\Omega_m}(\bs x)$  is the indicator function of subregion  $\Omega_m$. The local permeability tensors $\bs K_m(\bs \lambda_c)$ are positive definite matrices which, in the simplest case based on the isotropy assumption, take the form

\begin{equation}
\bs K_m(\bs \lambda_c) = \bs K_m(\lambda_{c, m}) = e^{\lambda_{c, m}} \bs I,
\label{eq:simpleK_l}
\end{equation}
where $\bs I$ is the identity matrix. We note that the number of cells $N_{\tx{cells}, c}$ does not need to coincide with the finite elements employed for the solution of the CGM. The former control the dimension of the reduced representation whereas the latter determine the computational cost in the solution of the CGM. Naturally, different discretizations or representations of the coarse model's spatially-varying properties can be adopted, e.g. $\blc$ could relate to  basis functions' coefficients in an appropriate expansion of $\bs K(\bs x, \bs \lambda_c)$.

Analogously to the FGM in \refeq{eq:FGM}, the Darcy-based, coarse-grained model (CGM) is defined as the deterministic mapping
\begin{equation}
    \bs u_c(\bs \lambda_c) ~ : ~ \bs \lambda_c \mapsto \bs u_c.
    \label{eq:CGM}
\end{equation}

\noindent \textbf{Remarks}
\bi
\item The low-dimensional latent variables $\bs \lambda_c$ encode the high dimensional descriptors $\bs \lambda_f$ of the fine scale porous domain of the Stokes flow FGM as an effective, Darcy-type permeability random field $\bs K = \bs K(\bs x, \bs \lambda_c)$. We re-emphasize  that it is extraneous if the latent representation $\bs \lambda_c$ is a faithful decoding of its fine scale analogue $\bs \lambda_f$ for the purpose of accurately reconstructing $\bs \lambda_f$ directly. Instead, $\bs \lambda_c$ must be maximally predictive (when solving the inexpensive CGM $\bs u_c(\bs \lambda_c)$) of the reconstruction of the FGM output $\bs u_f$. Hence, $\bs \lambda_c$ may be completely different from the reduced coordinates identified by typical dimensionality reduction  techniques applied directly on $\bs \lambda_f$. The encoder density $p_c(\bs \lambda_c|\bs \lambda_f, \bs \theta_c)$ should therefore be trained in such a way that it extracts all microstructural features from $\bs \lambda_f$ that are relevant in predicting the FGM output $\bs u_f$.

\item The decoder density $p_{cf}(\bs u_f|\revtwomath{\bs u_c(\bs \lambda_c)}, \bs \theta_{cf})$ attempts to reconstruct the FGM Stokes flow response $\bs u_f$ given the solution $\bs u_c(\bs \lambda_c)$ of the Darcy-based CGM \refeq{eq:Darcy}. While a general parametrization can be employed (e.g. making use of (deep) neural networks), this would necessitate either   huge amounts of data or very strong regularization assumptions. We exploit instead the spatial character of the problem, i.e. if the FGM and CGM outputs $u_{f, i}, u_{c, j}$ are directly related to spatial locations $\bs x_i, \bs x_j$ (e.g. if they represent pressure or velocity responses at these points), it is natural to assume that components $u_{c, j}$ corresponding to points $\bs x_j$ in the vicinity of $\bs x_i$ are most relevant for reconstruction of $u_{f, i}$ corresponding to point $\bs x_i$.

\item Independently of the values assigned to the parameters $\bt_c,\bt_{cf}$, predictions produced by the surrogate will be probabilistic as reflected by the density in \refeq{eq:singleLikelihood}.
This can be attributed to the epistemic uncertainty due to the unavoidable information loss during the coarse-graining process  as there is an upper bound on the mutual information $I(\blc, \buf) < I(\blf, \buf)$ when $\dim(\bs \lambda_c) \ll \dim(\bs \lambda_f)$ and no redundancies in $\bs \lambda_f$ \cite{Tishby2000}.
\ei
The following subsections are devoted to giving a clear description of the main building blocks $p_c$ and $p_{cf}$ as well as of the prior model we apply on the model parameters $\bs \theta_c$ and $\bs \theta_{cf}$.

 \subsection{Components of the proposed model}
 \label{sec:components}

\subsubsection{The encoder distribution \texorpdfstring{$p_c$}{}}
\label{sec:p_c}
\revone{The encoder distribution $p_c$ that effects the mapping from the high-dimensional $\blf$ to the lower-dimensional $\blc$ can assume various forms. Any successful strategy should be capable of extracting appropriate features from $\blf$ or, more generally, learning the right representation for predicting $\blc$. An obvious choice would be to exploit the expressive power of (deep) Neural Nets (NNs) and, given the spatial character of $\blf$, Convolutional Neural Networks (CNNs) in particular  \cite{fukushima_neocognitron_1980,lecun_gradient-based_1998,Krizhevsky2012}. 
A common misconception is that Deep Learning techniques are devoid of any feature engineering. While deep NNs are highly expressive, contain a lot of learnable parameters and reduce the need for feature engineering,  it is not true that data pre-processing or feature extraction are totally irrelevant\footnote{https://github.com/tensorflow/transform}. Several components, such as the number, size and type of layers or e.g. the type and number of pooling layers in CNNs are in fact hard-coded,  feature-extraction mechanisms.  More importantly, the basic premise in successful applications of these models is the availability of large amounts of data, from which the necessary information can be extracted. As mentioned earlier however, we operate in the Small Data regime.}

\revone{Based on this, we make use of  a large set of physically-motivated, feature functions $\varphi_{jm}(\blf)$ (discussed in detail in the sequel) in the context of the following model:}
\begin{equation}
\lambda_{c, m} = \sum_{j = 1}^{N_{\tx{features}, m}} \tilde{\theta}_{c, jm} \varphi_{jm}(\bs \lambda_f) + \sigma_{c, m} Z_m, \qquad Z_m \sim \mathcal N(0, 1),
\label{eq:p_cModel}
\end{equation}
In principle, the type and number of features can differ with every latent space component $m$. These feature functions are linearly combined with coefficients $\tilde{\bs \theta}_{c, m} = \left\{\tilde{\theta}_{c, jm} \right\}_{j = 1}^{N_{\tx{features}, m}}$ (potentially different for every component $m$ as well) and augmented with residual Gaussian white noise of variance $\sigma_{c, m}^2$, so that the set of free parameters for the distribution $p_c$ is $\bs \theta_c = \left\{\tilde{\bs \theta}_{c, m}, \sigma_{c, m}^2 \right\}_{m = 1}^{\dim(\bs \lambda_c)}$ \footnote{We use the notation $\bs \Sigma_c = \tx{diag}[\bs \sigma_c^2]$, where $\bs \sigma_c^2 = \left\{\sigma_{c, m}^2 \right\}_{m = 1}^{\dim{\bs \lambda_c}}$ is the vector of variances, wherever it is more convenient.}.

The distribution for $p_c$ is thus
\begin{equation}
    p_c(\bs \lambda_c|\bs \lambda_f, \bs \theta_c) = \prod_{m = 1}^{\dim(\bs \lambda_c)} \mathcal N(\lambda_{c, m}|\tilde{\bs \theta}_{c, m}^T \bs\varphi_m(\bs \lambda_f), \sigma_{c, m}^2),
    \label{eq:p_c}
\end{equation}
with $\bs \varphi_m(\bs \lambda_f) = \left\{\varphi_{jm}(\bs \lambda_f) \right\}_{j = 1}^{N_{\tx{features}, m}}$ being the vector of feature functions used to predict the latent space component $\bs \lambda_{c, m}$.

\paragraph{\revtwo{Feature functions:}}
\begin{figure}[!t]
\centering
\includegraphics[width=\textwidth]{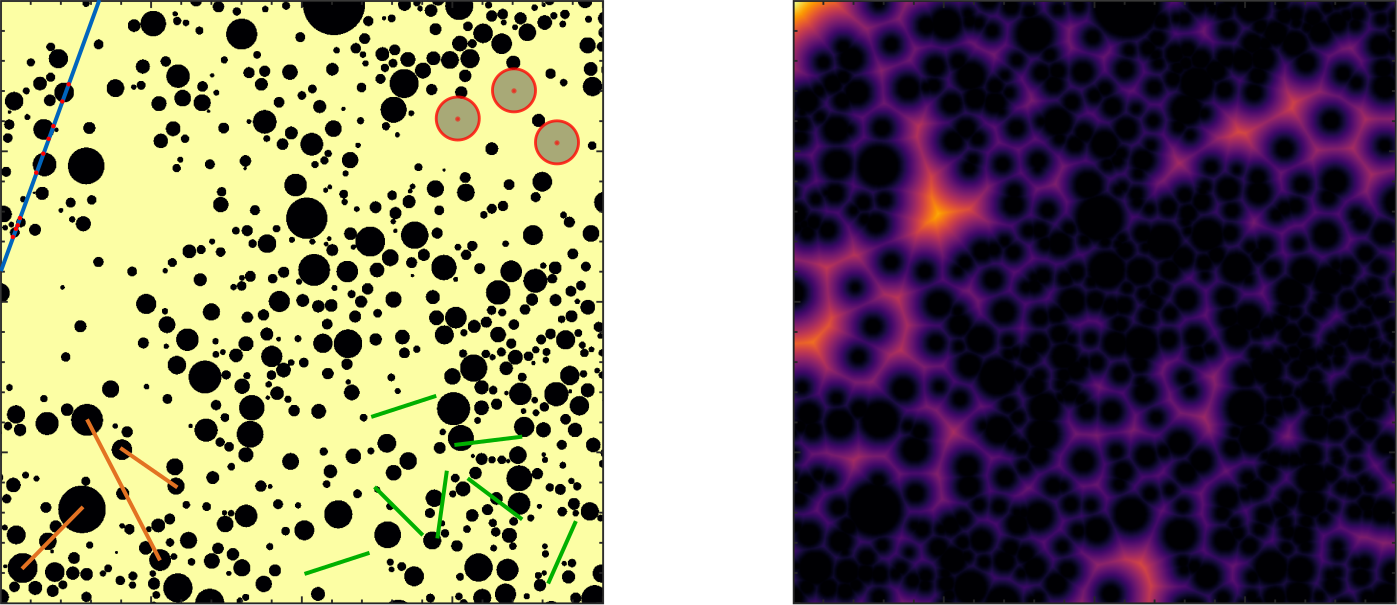}
\caption{\revtwo{Graphical illustration of certain classes of feature functions. Left: Depiction of the \tit{chord length density} (blue line with red dots), \tit{pore size density} (red/gray circles), \tit{mutual distances} (orange lines) and \tit{lineal path function} (green lines). Right: Euclidean distance transform. Explanations see text.}}
\label{fig:featureFunctions}
\end{figure}
The most critical aspect in the expressivity of the encoder pertains to the  feature functions $\varphi_m(\bs \lambda_f)$ employed. In this work, we make use of morphological measures \cite{Torquato2001, Lu1992, Torquato1993, Lowell2006}, quantities from classical fluid dynamics \cite{Sutera1993, Kozeny1927, Carman1937} or other physics \cite{Archie1947}, image analysis \cite{Soille1999} and low-dimensional autoencoder representations \cite{Doersch2016}. \revtwo{Some important classes of possible feature functions are depicted in Figure \ref{fig:featureFunctions} and briefly explained in the following list:
\begin{itemize}
\setlength{\itemsep}{0pt}
    \item \tit{Pore fraction}: The fraction of pore space $|\Omega_f|/|\Omega|$ over a certain region and several nonlinear functions thereof (e.g. $\exp, \log, $ certain powers);
    \item \tit{Interface area}: The interface area between matrix- and exclusion phase over a certain region and nonlinear functions thereof;
    \item \tit{Effective medium approximations}: There exist several approximation formula for effective material properties in diffusion problems (Bruggeman formula \cite{Bruggeman1935}, Maxwell-Garnett approximation \cite{Landauer1978}) which can be generalized to the infinite contrast limit of matrix/exclusion phase permeabilities;
    \item \tit{Chord length density}: The density of lengths of segments of an infinitely long line being in either matrix or exclusion phase, see blue line with red dots in Figure \ref{fig:featureFunctions}. A `segment' is a part of the line between two neighboring dots -- the feature function is given by evaluating the chord length density for several lengths $d$, see \ref{featureTable} for the length scales used in the experiments;
    \item \tit{Statistics of exclusion radii}: Mean, variance, max, min, etc. of exclusion radii;
    \item \tit{Statistics of mutual exclusion distances}: Mean, variance, max, min, etc. of mutual exclusion distances (center-to-center or edge-to-edge). See the orange lines in Figure \ref{fig:featureFunctions} for illustration;
    \item \tit{Pore-size density}: Density of the distance from a random point in $\Omega$ to the nearest matrix/exclusion interface. For illustration see the red/gray circles in Figure \ref{fig:featureFunctions}. Nearest-neighbor functions \cite{Torquato2001} are closely related. Again, the pore-size density is to be evaluated for several lengths $d$, see \ref{featureTable};
    \item \tit{Lineal-path function}: The probability of a random line of length $d$ being purely in matrix (or exclusion) phase. See the green lines in Figure \ref{fig:featureFunctions} for illustration and \ref{featureTable} for lengths $d$ used;
    \item \tit{2-point correlations}: The probability, that two random points of distance $d$ lie both in matrix (exclusion) phase -- see \ref{featureTable} for lengths $d$;
    \item \tit{Distance transforms}: For any point in $\Omega$, compute the distance to the closest exclusion (matrix phase), see right part of Figure \ref{fig:featureFunctions}. Features can be constructed from statistics of the distance transform.
\end{itemize}}

Ultimately, several hundreds of such functions can be deployed -- \revtwo{in Section \ref{sec:experiments}, we use the 150 listed in Table \ref{featureTable}}. In order to avoid overfitting in the \textbf{Small Data} regime we are operating (i.e. $N \lesssim 100$), we make use of a fully Bayesian, sparsity-inducing prior model \cite{Bishop2000} capable of automatically controlling model complexity by pruning all irrelevant features and identifying the most salient ones, see section \ref{sec:prior}.

\subsubsection{The decoder distribution \texorpdfstring{$p_{cf}$}{}}
\label{sec:p_cf}
The probabilistic coarse-to-fine mapping $p_{cf}(\bs u_f|\bs u_c, \bs \theta_{cf})$ should take into account the spatial characteristics of the problem, e.g. that FGM and CGM outputs $u_{f, i}, u_{c, j}$ may be given by physical quantities evaluated at spatial locations $\bs x_i, \bs x_j$. For simplicity, let us assume that $u_{f, i} = P(\bs x_i)$ is the pressure field solution of the FGM \refeq{eq:StokesPDE} interpolated at a regular fine scale grid $G^{(f)}$, $\bs x_i \in G^{(f)}$, and $u_{c, j} = \bar{P}(\bs x_j)$ its Darcy counterpart from \refeq{eq:Darcy} evaluated at a much coarser grid $G^{(c)}$, $\bs x_j \in G^{(c)}$. Then the model for $p_{cf}$ could look like
\begin{equation}
u_{f,i} = \sum_{j \in G^{(c)}} W_{ij} u_{c, j}(\bs \lambda_c) + \tau_{cf, i}^{-1/2}Z_i, \qquad Z_i \sim \mathcal N(0, 1),
\label{eq:p_cf}
\end{equation}
where $\bs W$ should have the structure of an interpolation matrix from $G^{(c)}$ to $G^{(f)}$ and $\tx{diag}[\bs \tau_{cf}]$ is a diagonal precision matrix to be learned from the data. The decoder distribution $p_{cf}$ can thus be written as
\begin{equation}
p_{cf}(\bs u_f|\bs u_c(\bs \lambda_c), \bs \theta_{cf}) = \mathcal N(\bs u_f| \bs W \bs u_c(\bs \lambda_c), \tx{diag}[\bs \tau_{cf}^{-1}]),
\label{eq:p_cfGauss}
\end{equation}
with model parameters $\bs \theta_{cf} = \left\{\bs W, \bs \tau_{cf}\right\}$. We emphasize that many different possibilities exist in terms of the parametrization  and architecture of $p_{cf}$. \revone{One should note here 
that, in contrast to $p_c$ which involves a map from the high-dimensional $\blf$ to a lower-dimensional $\blc$, $p_{cf}$ encompasses a map from the lower-dimensional $\bu_c$ to the high-dimensional $\bu_f$ which is obviously much more manageable.
As with $p_c$, the option of NNs comes to mind and given the spatial character of both $\bu_c$ and $\bu_f$ CNNs could be introduced. We have again decided against this option, due to the {\em Small Data} setting of the problem.}
\revone{Given the lower dimension of the inputs $\bu_c$, nonlinear models employing polynomials, kernels etc could be employed. An interesting alternative involves Bayesian non-parametric extensions. For instance, a Gaussian Process regression could be deployed for the probabilistic mapping $u_{c, j} = \bar{P}(\bs x_j) \mapsto u_{f, i} = P(\bs x_i)$, exploiting the spatial structure of the problem by making use of a stationary covariance kernel of the form $\tx{cov}(u_{f, i}, u_{c, j}) = \tx{cov}(P(\bs x_i), \bar{P}(\bs x_j)) = \tx{cov}(\bs x_i, \bs x_j) = \tx{cov}(|\bs x_i - \bs x_j|)$. Alternatively, } each $u_{f,i}$ can be expressed using an independent Gaussian Process in a fashion analogous to GP-LVMs \cite{lawrence_probabilistic_2005,titsias_bayesian_2010}. Such an extension is not pursued here due to the complexities associated with the inference and prediction steps \cite{damianou_variational_2015}.

\subsubsection{The prior model}
\label{sec:prior}
We advocate a fully Bayesian formulation where the (approximate) posterior densities on the model parameters $\bt_c, \bt_{cf}$ may be computed.  In view of the \textbf{Small Data} setting, we employ priors that can safeguard against overfitting, that do not require any fine-tuning or have ad hoc hyperparameters, but also promote interpretability of the results. To that end, in order to induce  sparsity in the feature functions of $p_c$ given in \refeq{eq:p_c}, we make use of the Automatic Relevance Determination (ARD) model \cite{Faul2003} corresponding to
\begin{equation}
    p(\tilde{\bs \theta}_c| \bs \gamma) = \prod_{m = 1}^{\dim(\bs \lambda_c)}\prod_{j = 1}^{N_{\tx{features}, m}} \mathcal N(\tilde{\theta}_{c, jm}| 0, \gamma^{-1}_{jm}),
    \label{eq:priorTheta}
\end{equation}
where the precision hyperparameters $\gamma_{jm}$  are equipped with a conjugate \tit{Gamma} hyperprior
\begin{equation}
    p(\gamma_{jm}) = Gamma(\gamma_{jm}| a, b) = b^a \gamma_{jm}^{a - 1} e^{-b\gamma_{jm}}/\varGamma(a).
    \label{eq:gammaHyperprior}
\end{equation}
This hierarchical model has been widely studied and the capability of pruning irrelevant parameters $\theta_{c,jm}$ has been shown when small values for the hyperparameters are employed\footnote{\label{ft:hyperparams}In our investigations, we set $a = b = c = d = e = f =  10^{-10}$ \revtwo{to avoid division by zero without affecting the converged values}.} \cite{Bishop2000, Tipping2001}.

Similarly, we apply conjugate, uninformative \tit{Gamma} hyperpriors\footnote{See footnote \ref{ft:hyperparams}} on the precision parameters $\bs \tau_c, \bs \tau_{cf}$ in the  encoder/decoder distributions $p_{c}$ and $p_{cf}$, i.e.
\begin{align}
    p(\tau_{c, m}) &= Gamma(\tau_{c, m}| c, d), \\
    p(\tau_{cf, i}) &= Gamma(\tau_{cf, i}| e, f).
\end{align}
A vague, uninformative  prior could also be used for $\bs{W}$ but as this was found to have no measurable effect on the results and in order to lighten the notation, we completely omit it.

\subsection{Model training}
\label{sec:modelTraining}
\begin{figure}[!t]
\centering
\includegraphics[width=.7\textwidth]{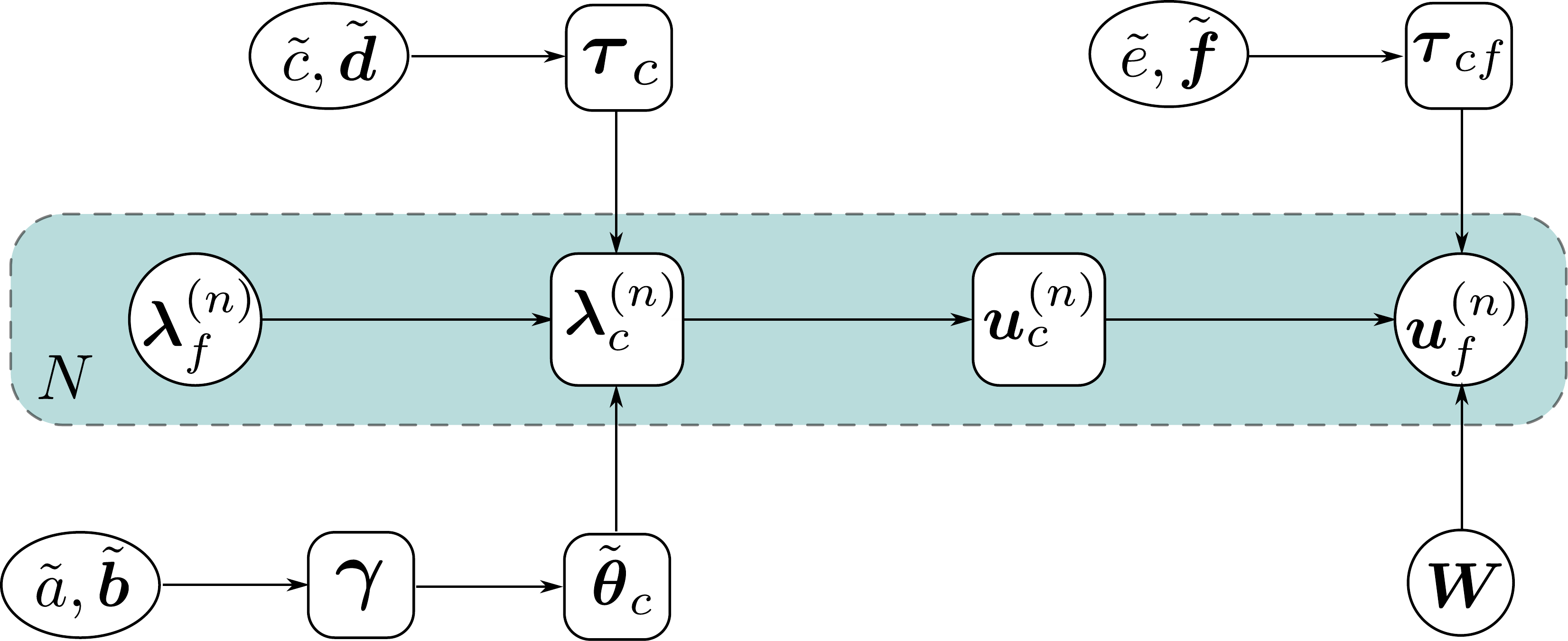}
\caption{Graphical representation of the Bayesian network defined by \refeq{eq:fullPosterior}. All internal vertices (with teal background) are latent variables.}
\label{fig:bayesnet}
\end{figure}

The probabilistic graphical model described  thus far is depicted in Figure \ref{fig:bayesnet}.
Given a dataset consisting of FGM input/output pairs (i.e. Stokes-flow model simulations)  $\mathcal D = \left\{\bs \lambda_f^{(n)}, \bs u_f^{(n)} \right\}_{n = 1}^N$, the model likelihood can be written as (see \refeq{eq:singleLikelihood}) 
\begin{equation}
 \begin{split}
 \mathcal L(\bs \theta_{cf}, \bs \theta_c) = \prod_{n = 1}^N &p(\bs u_f^{(n)}|\bs \lambda_f^{(n)}, \bs \theta_{cf}, \bs \theta_c)  \\
 = \prod_{n = 1}^N &\int \mathcal N(\bs u_f^{(n)}| \bs W \bs u_c(\bs \lambda_c^{(n)}), \tx{diag}[\bs \tau_{cf}]^{-1}) \cdot \\
 \prod_{m = 1}^{\dim(\bs \lambda_c)} &\left[ \mathcal N(\lambda_{c, m}^{(n)}|\tilde{\bs \theta}_{c, m}^T \bs\varphi_m(\bs \lambda_f^{(n)}), \tau_{c, m}^{-1}) \right] d\bs \lambda_c^{(n)},
 \end{split}
 \label{eq:likelihood}
 \end{equation}
 where $\tx{diag}[\bs \tau_{c}] = \tx{diag}[\bs \sigma_c^{-2}]$ is a diagonal precision matrix and $\bs \lambda_c^{(n)}$ are the latent variables that encode the coarse-grained material properties for each sample $n$.
Given the aforementioned priors, the posterior of the model parameters is
\begin{equation}
\begin{split}
    p(\tilde{\bs \theta}_c, \bs \tau_c, \bs \tau_{cf}| \mathcal D) \propto \prod_{n = 1}^N &\left[ \int \right. \mathcal N(\bs u_f^{(n)}| \bs W \bs u_c(\bs \lambda_c^{(n)}), \tx{diag}[\bs \tau_{cf}]^{-1}) \\
    \prod_{m = 1}^{\dim(\bs \lambda_c)} &\left[ \mathcal N(\lambda_{c, m}^{(n)}|\tilde{\bs \theta}_{c, m}^T \bs\varphi_m(\bs \lambda_f^{(n)}), \tau_{c, m}^{-1}) \right] d \bs \lambda_c^{(n)} \left. \vphantom{\int} \right] \\
    &\cdot \int p(\tilde{\bs \theta}_c|\bs \gamma)  p(\bs \gamma) d\bs \gamma \cdot p(\bs \tau_c)  p(\bs \tau_{cf})
    \label{eq:posterior}
\end{split}
\end{equation}
where $\tilde{\bs \theta}_{c} = \left\{\tilde{\bs \theta}_{c, m} \right\}_{m = 1}^{\tx{dim}(\bs \lambda_c)}$. 
The posterior over all variables $\bs \theta = \left\{\left\{\bs \lambda_c^{(n)} \right\}_{n = 1}^N, \right.$ $\left. \tilde{\bs \theta}_c, \bs \tau_c, \bs \tau_{cf}, \bs \gamma \vphantom{\left\{\bs \lambda_c^{(n)} \right\}_{n = 1}^N} \right\}$ is given by
\begin{equation}
\begin{gathered}
p(\bt| \mathcal D) = \cfrac{p(\bs \theta, \mathcal D) }{p(\mathcal D)} = \frac{1}{p(\mathcal D)} \prod_{n = 1}^N \Bigg[ \mathcal N(\bs u_f^{(n)}| \bs W \bs u_c(\bs \lambda_c^{(n)}), \tx{diag}[\bs \tau_{cf}]^{-1}) \cdot \\
\cdot \prod_{m = 1}^{\dim(\bs \lambda_c)}\left[ \mathcal N(\lambda_{c, m}^{(n)}|\tilde{\bs \theta}_{c, m}^T \bs\varphi_m(\bs \lambda_f^{(n)}), \tau_{c, m}^{-1}) \right] \Bigg]  ~
    p(\tilde{\bs \theta}_c|\bs \gamma)  p(\bs \gamma)  p(\bs \tau_c)  p(\bs \tau_{cf})
\end{gathered}
\label{eq:fullPosterior}
\end{equation}
where $p(\mathcal D)$ is the model evidence of the postulated coarse-graining model \cite{gelman_bayesian_2003}. This is a pivotal quantity in model validation that as we show later will serve as the main driver  for the refinement/enhancement of the CGM.

Evaluating the integrals in \refeq{eq:posterior} is analytically intractable, making necessary the use of suitable (approximate) inference methods that are capable of giving fast and accurate estimates of the above posterior.
In this work, we employ Stochastic Variational Inference (SVI, \cite{Paisley2012,Hoffman2013}) methods which produce closed-form approximations of the true posterior $p(\bt| \mathcal D)$ and simultaneously of the model evidence $p(\mathcal D)$. In contrast to sampling-based procedures (e.g. MCMC, SMC), SVI yields biased estimates at the benefit of computational efficiency and computable convergence objectives in the form of the Kullback-Leibler (KL) divergence between the approximation $Q(\bt)$ and the true posterior $p(\bt |\mathcal{D})$ \cite{blei2017variational}:
 \begin{equation}
    \tx{KL}(Q(\bs \theta) || p(\bs \theta|\mathcal D)) = - \int Q(\bs \theta) \log \frac{p(\bs \theta|\mathcal D)}{Q(\bs \theta)} d\bs \theta.
    \label{eq:KLdiv}
\end{equation}
We note that the $\log$ evidence $\log p(\mathcal D)$ can be decomposed as
\begin{equation}
\log p(\mathcal D) = \log \int p(\bs \theta, \mathcal D) d\bs \theta = \mathcal F(Q) + \tx{KL}(Q(\bs \theta) || p(\bs \theta|\mathcal D)),
\end{equation}
where
\begin{equation}
\mathcal F(Q) = \int Q(\bs \theta) \log \frac{p(\bs \theta, \mathcal D)}{Q(\bs \theta)} d\bs \theta.
\label{eq:elbo}
\end{equation}
Since $\tx{KL}(Q(\bs \theta) || p(\bs \theta|\mathcal D)) \ge 0$, $\mathcal F(Q)$ provides a rigorous lower bound to $\log p(\mathcal D)$, called the \tit{evidence lower bound} (ELBO). Hence minimizing $\tx{KL}(Q(\bs \theta) || p(\bs \theta|\mathcal D))$ $\ge 0$ is equivalent to maximizing the ELBO $\mathcal F(Q)$.

We employ a \tit{mean-field approximation} to the full posterior based on the following decomposition: \begin{equation}
\begin{split}
\label{eq:meanfield}
&Q(\left\{\bs \lambda_c^{(n)} \right\}_{n = 1}^N, \tilde{\bs \theta}_c, \bs \tau_c, \bs \tau_{cf}, \bs \gamma) = \\
&= \prod_{m = 1}^{\tx{dim}(\bs \lambda_c)} \left[Q_{\tau_{c, m}}(\tau_{c, m}) 
\prod_{j = 1}^{N_{\tx{features}, m}}\left[ Q_{\tilde{\theta}_{c, jm}}(\tilde{\theta}_{c, jm}) Q_{\gamma_{jm}}(\gamma_{jm}) \right] \right] \\
&~\quad \prod_{i = 1}^{\tx{dim}(\bs u_f)}\left[Q_{\tau_{cf, i}}(\tau_{cf, i}) \right] \prod_{n = 1}^N Q_{\bs \lambda_c^{(n)}}(\bs \lambda_c^{(n)}) \\
&\approx p(\left\{\bs \lambda_c^{(n)} \right\}_{n = 1}^N, \tilde{\bs \theta}_c, \bs \tau_c, \bs \tau_{cf}, \bs \gamma| \mathcal D).
\end{split}
\end{equation}
If $\bt_{k}$ denotes an arbitrary subset of the parameters $\bt$ above (i.e. if $\bs \theta_k$ is effectively one of the component variables $\tilde{\theta}_{c, jm},\tau_{c, m},\gamma_{jm},\tau_{cf, i},\bs \lambda_c^{(n)}$), then the optimal $Q_{\bs \theta_k}(\bs \theta_k)$ can be found by setting the first order variation of $\mathcal F(Q)$ to zero yielding \cite{beal_variational_2003}
\begin{equation}
Q_{\bs \theta_k}(\bs \theta_k) = \frac{\exp \left<\log p(\bs \theta, \mathcal D) \right>_{\ell \neq k}}{\int \exp \left<\log p(\bs \theta, \mathcal D) \right>_{\ell \neq k} d\theta_k},
\label{eq:qoptpartial}
\end{equation}
where $\left<~\cdot ~\right>_{\ell \neq k}$ denotes expectation w.r.t. all $Q_{\bs \theta_\ell}$'s except for $Q_{\bs \theta_k}$.
We emphasize that every $Q_{\bs \theta_k}$ implicitly depends on all other $Q_{\bs \theta_\ell}$'s, which means we need to self-consistently cycle over all $k$ and update $Q_{\bs \theta_k}$ given expected values w.r.t. all $Q_{\bs\theta_\ell}, ~ \ell\neq k$, until convergence is attained.
We provide complete details of the derivations in \ref{ap:optQ} and summarize here the closed-form update equations
\begin{equation}
    \label{eq:Qgamma}
    \begin{split}
    Q_{\gamma_{jm}}(\gamma_{jm}) &= Gamma(\gamma_{jm}|\tilde{a}, \tilde b_{jm}), \\
    \tilde{a} = a + \frac{1}{2},    &\qquad   \tilde{b}_{jm} = b + \frac{1}{2} \left<\tilde \theta_{c, jm}^2 \right>,
    \end{split}
\end{equation}
\vspace{3mm}
\begin{equation}
\begin{gathered}
    \label{eq:Qtau_c}
    Q_{\tau_{c, m}}(\tau_{c, m}) = Gamma(\tau_{c, m}| \tilde{c}, \tilde{d}_m), \\
    \tilde{c} = c + \frac{N}{2}, \qquad  \tilde{d}_m = d + \frac{1}{2}\sum_{n = 1}^N \left< \left(\lambda_{c, m}^{(n)} - \tilde{\bs \theta}_{c, m}^T \bs \varphi_m(\bs \lambda_f^{(n)}) \right)^2\right>,
\end{gathered}
\end{equation}
\vspace{3mm}
\begin{equation}
\begin{gathered}
    \label{eq:Qtau_cf}
    Q_{\tau_{cf, i}}(\tau_{cf, i}) = Gamma(\tau_{cf, i}|\tilde e, \tilde f_i), \\
    \tilde e = e + \frac{N}{2}, \qquad  \tilde f_i = f + \frac{1}{2} \sum_{n = 1}^N \left<\left[\bs u_f^{(n)} - \bs W\bs u_c(\bs \lambda_c^{(n)}) \right]_i^2 \right>,
\end{gathered}
\end{equation}
\vspace{3mm}
\begin{equation}
\begin{gathered}
    Q_{\tilde{\theta}_{c, jm}}(\tilde{\theta}_{c, jm}) = \mathcal N(\tilde{\theta}_{c, jm}| \mu_{\tilde{\theta}_{c, jm}}, \sigma^2_{\tilde{\theta}_{c, jm}}),
    \label{eq:Qtheta} \\
    \sigma_{\tilde{\theta}_{c, jm}}^2 = \left(\left<\tau_{c, m}\right>\sum_{n = 1}^N (\varphi_{jm}^{(n)})^2 + \left<\gamma_{jm}\right> \right)^{-1}, \\
    \mu_{\tilde{\theta}_{c, jm}} = \sigma_{\tilde{\theta}_{c, jm}}^2 \left<\tau_{c, m}\right> \sum_n \varphi^{(n)}_{jm}\left(\left<\lambda_{c, m}^{(n)}\right> - \sum_{k \neq j} \varphi_{km}^{(n)}\left<\tilde \theta_{c, km}\right>\right),
\end{gathered}
\end{equation}
where $\left<~\cdot ~\right>$ denotes expected values w.r.t. $Q(\left\{\bs \lambda_c^{(n)} \right\}_{n = 1}^N, \tilde{\bs \theta}_c, \bs \tau_c, \bs \tau_{cf}, \bs \gamma)$ as specified in \eqref{eq:meanfield}. Due to the choice of conjugate priors, most expected values are given in closed form,
\begin{equation}
\begin{gathered}
\left< \gamma_{jm} \right> = \frac{\tilde a}{\tilde{b}_{jm}},  \left<\tau_{c, m} \right> = \frac{\tilde{c}}{\tilde{d}_m}, \left<\tau_{cf, i} \right> = \frac{\tilde{e}}{\tilde{f}_i}, \\
\left<\tilde{\theta}_{c, jm} \right> = \mu_{\tilde{\bs \theta}_{c, jm}},  \left<\tilde{\theta}_{c, jm}^2\right> = \mu_{\tilde{\theta}_{c, jm}}^2 + \sigma^2_{\tilde{\theta}_{c, jm}}.
\end{gathered}
\label{eq:expectedValues}
\end{equation}

\begin{algorithm}[t]
\KwData{$\mathcal D = \left\{\bs \lambda_f^{(n)}, \bs u_f^{(n)}\right\}_{n = 1}^N$ \tcp*{Training data}}
\KwIn{$\tilde{b}_{jm} \leftarrow \tilde{b}_{jm}^{(0)},~ \tilde{d}_{m} \leftarrow \tilde{d}_m^{(0)},~ \tilde{f}_i \leftarrow \tilde{f}^{(0)}_i,~ \mu_{\tilde{\theta}_{c, jm}} \leftarrow \mu_{\tilde{\theta}_{c, jm}}^{(0)},~ \sigma_{\tilde{\theta}_{c, jm}}^2 \leftarrow (\sigma_{\tilde{\theta}_{c, jm}}^{(0)})^2$ \tcp*{Initialization}}

\vspace{3mm}
\KwOut{Variational approximation $Q(\bs \theta)$ to $p(\bs \theta|\mathcal D)$ minimizing KL-divergence eq. \eqref{eq:KLdiv}}

\vspace{2mm}
Evaluate all features $\{ \bs \varphi_m(\bs \lambda_f^{(n)}) \}_{n=1}^N$;

\vspace{2mm}
$\tilde a \leftarrow a + \frac{1}{2}$, \qquad $\tilde c = c + \frac{N}{2}$, \qquad $\tilde e = e + \frac{N}{2}$;

\While{(not converged)}{
    \For{$n \leftarrow 0$ \KwTo $N$}{
    \tcp{Fully parallelizable in $n$}
    Update $\tilde{Q}_{\bs \lambda_c^{(n)}}(\bs \lambda_c^{(n)}| \bs \mu_{\bs \lambda_c}^{(n)}, \bs \sigma_{\bs \lambda_c}^{(n)})$ according to eq. \eqref{eq:Qlambda}--\eqref{eq:VI}
    
    Estimate $\left< \lambda_{c,m}^{(n)}\right>_{\tilde{Q}_{\bs \lambda_c^{(n)}}}$, \quad $\left< (\lambda_{c, m}^{(n)})^2\right>_{\tilde{Q}_{\bs \lambda_c^{(n)}}}$,\quad $\left<u_{c, k}(\bs \lambda_{c}^{(n)}) \right>_{\tilde{Q}_{\bs \lambda_c^{(n)}}}$, \quad $\left<u_{c, k}^2(\bs \lambda_{c}^{(n)}) \right>_{\tilde{Q}_{\bs \lambda_c^{(n)}}}$
    }
    
    Update $Q_{\tilde{\bs \theta}_{c, m}}(\tilde{\bs \theta}_{c, m}|\bs \mu_{\tilde{\bs \theta}_{c, m}}, \bs \Sigma_{\tilde{\bs \theta}_{c, m}})$ according to \eqref{eq:Qtheta} and \eqref{eq:expectedValues};
    
    Update $Q_{\gamma_{jm}}(\gamma_{jm}| \tilde a, \tilde b_{jm})$ according to \eqref{eq:Qgamma} and \eqref{eq:expectedValues};
    
    Update $Q_{\tau_{c, m}}(\tau_{c, m}| \tilde c, \tilde d_m)$ according to \eqref{eq:Qtau_c} and \eqref{eq:expectedValues};
    
    Update $Q_{\tau_{cf, i}}(\tau_{cf, i}| \tilde e, \tilde{f}_i)$ according to \eqref{eq:Qtau_cf};
}

\Return{Variational approximation $Q(\bs \theta)$ to $p(\bs \theta|\mathcal D)$}
\caption{Model training.}
\label{alg:modelTraining}
\end{algorithm}

The optimal form of the remaining approximate posteriors, i.e. $Q_{\bs \lambda_c^{(n)}}$, is given by (based on \refeq{eq:qoptpartial})
\begin{small}
\begin{equation}
\begin{gathered}
Q_{\bs \lambda_c^{(n)}}(\bs \lambda_c^{(n)}) \propto  \\
\left<\mathcal N(\bs u_f^{(n)}| \bs W \bs u_c(\bs \lambda_c^{(n)}), \tx{diag}[\bs \tau_{cf}]^{-1}) \prod_{m = 1}^{\dim(\bs \lambda_c)}\left[ \mathcal N(\lambda_{c, m}^{(n)}|\tilde{\bs \theta}_{c, m}^T \bs\varphi_m(\bs \lambda_f^{(n)}), \tau_{c, m}^{-1})\right] \right>
\end{gathered}
\label{eq:Qlambda}
\end{equation}
\end{small}
which is analytically intractable due to the expectations with respect to the output of the CGM $\bs u_c(\bs \lambda_c^{(n)})$ (i.e. of the Darcy-flow simulator).
Instead of approximating the expectations above by Monte Carlo, we retain the variational character of the algorithm by employing approximations to the optimal $Q_{\bs \lambda_c^{(n)}}$ which take the form of multivariate Gaussians with a diagonal covariance, i.e.
\begin{equation}
\tilde{Q}_{\bs \lambda_c^{(n)}}(\bs \lambda_c^{(n)}| \bs \mu_{\bs \lambda_c}^{(n)}, \bs \sigma_{\bs \lambda_c}^{(n)}) \approx \mathcal N(\bs \lambda_c^{(n)}| \bs \mu_{\bs \lambda_c}^{(n)}, \tx{diag}[(\bs \sigma_{\bs \lambda_c}^{(n)})^2]),
\label{eq:qlambda}
\end{equation}
and find the optimal values of the parameters $\bs \mu_{\bs \lambda_c}^{(n)}, \bs \sigma_{\bs \lambda_c}^{(n)}$ via black-box variational inference \cite{Ranganath2014, Hoffman2013, Paisley2012}, i.e. by minimizing the KL-divergence between $\tilde{Q}_{\bs \lambda_c^{(n)}}$ and the optimal $Q_{\bs \lambda_c^{(n)}}$ in \refeq{eq:Qlambda},
\begin{equation}
\begin{split}
\min_{\bs \mu_{\bs \lambda_c}^{(n)}, \bs \sigma_{\bs \lambda_c}^{(n)}} & \tx{KL}\left(\left.\tilde{Q}_{\bs \lambda_c^{(n)}}(\bs \lambda_c^{(n)}|\bs \mu_{\bs \lambda_c}^{(n)}, \bs \sigma_{\bs \lambda_c}^{(n)}) \right|\left| Q_{\bs \lambda_c^{(n)}}(\bs \lambda_c^{(n)})\right.\right).
\end{split}
\label{eq:optqlambda}
\end{equation}
The computation of the objective above as well as gradients with respect to $\bs \mu_{\bs \lambda_c}^{(n)}, (\bs \sigma_{\bs \lambda_c}^{(n)})^2$ involves expectations of the CGM outputs as well as their respective gradients $\pa \bs u_c(\bs \lambda_c^{(n)})/\pa \bs \lambda_c^{(n)}$. The latter are efficiently computed using adjoint formulations  (see e.g. \cite{Heinkenschloss2008, Constantine2014}). In order to minimize the noise in the expectations involved, we apply the  the reparametrization trick \cite{Kingma2013} and supply the stochastic gradients to the \tit{adaptive moment estimation} optimizer (\tit{ADAM} \cite{Kingma2014}). We note that this stochastic optimization problem needs to be run for every data point $n$ in every training iteration, but is entirely parallelizable in $n$.
After optimization, the expected values $\left<\lambda_{c, m}^{(n)}\right>, \left<(\lambda_{c, m}^{(n)})^2 \right>$ are readily given by the first and second moments of the resulting Gaussian specified in Equation \eqref{eq:qlambda}, whereas the expected values $\left<u_{c, k}(\bs \lambda_c^{(n)}) \right>, \left<u_{c, k}^2(\bs \lambda_c^{(n)}) \right>$ can efficiently be obtained by direct Monte Carlo.
Full details about the performed black-box VI are given in \ref{ap:bbvi}.
All steps of the training process are summarized in Algorithm \ref{alg:modelTraining}.

\subsection{Model predictions}
\label{sec:predictions}
A main feature of the coarse-grained model presented in this work is the capability to produce probabilistic predictions that  quantify uncertainty both due to limited training data and due to limited model complexity, i.e. the information loss happening during the coarse-graining process going  $\bs \lambda_f \to \bs \lambda_c \to \buc \to \buf$. In particular, given training data $\mathcal{D}$ and a new FGM input $\blf$ (which is not included in the training data $\mathcal D$), the predictive posterior density $p_{\tx{pred}}(\bs u_f|\bs \lambda_f, \mathcal D)$ for the corresponding FGM output $\buf$ is
\begin{equation}
\begin{gathered}
    p_{\tx{pred}}(\bs u_f|\bs \lambda_f, \mathcal D) = \int \underbrace{p(\bs u_f|\bs \lambda_f, \bs \theta)}_{\tx{\refeq{eq:singleLikelihood}}} p(\bs \theta|\mathcal D) d\bs \theta \\
    = \int p_{cf}(\bs u_f|\bs u_c(\bs \lambda_c), \bs \theta_{cf}) p_c(\bs \lambda_c|\bs \lambda_f, \bs \theta_c)d\bs \lambda_c p(\bs \theta_{cf}, \bs \theta_c|\mathcal D) d \bt_c~ d\bt_{cf},
\end{gathered}
\label{eq:p_pred0}
\end{equation}
with the decoding/encoding distributions $p_{c}, p_{cf}$ as described in sections \ref{sec:p_c}, \ref{sec:p_cf}. We make use of the variational approximation (section \ref{sec:modelTraining}) to the posterior
\begin{equation}
p(\bs \theta|\mathcal D) = p(\tilde{\bs \theta}_c, \bs \tau_c, \bs \tau_{cf}|\mathcal D) \approx Q_{\tilde{\bs \theta}_c}(\tilde{\bs \theta}_c) Q_{\bs \tau_c}(\bs \tau_c) Q_{\bs \tau_{cf}}(\bs \tau_{cf}),
\end{equation}
in order to rewrite the predictive posterior as
\begin{small}
\begin{equation}
\begin{split}
    p_{\tx{pred}}(\bs u_f|\bs \lambda_f, \mathcal D) &= \int p_{cf}(\bs u_f|\bs u_c(\bs \lambda_c), \bs \tau_{cf}) p_c(\bs \lambda_c|\bs \lambda_f, \tilde{\bs \theta}_c, \bs \tau_c)d\bs \lambda_c \\
    & \qquad \cdot Q_{\tilde{\bs \theta}_c}(\tilde{\bs \theta}_c) d \tilde{\bs \theta}_c \cdot Q_{\bs \tau_c}(\bs \tau_c) d\bs \tau_c \cdot Q_{\bs \tau_{cf}}(\bs \tau_{cf}) d\bs \tau_{cf} \\
    &= \int  p_{cf}(\bs u_f|\bs u_c(\bs \lambda_c), \bs \tau_{cf}) \left[\int p_c(\bs \lambda_c|\bs \lambda_f, \tilde{\bs \theta}_c, \bs \tau_c) Q_{\tilde{\bs \theta}_c}(\tilde{\bs \theta}_c) d \tilde{\bs \theta}_c \right] d\bs \lambda_c \\
    &\qquad \cdot Q_{\bs \tau_c}(\bs \tau_c) d\bs \tau_c \cdot Q_{\bs \tau_{cf}}(\bs \tau_{cf}) d\bs \tau_{cf}.
    \label{eq:p_pred}
\end{split}
\end{equation}
\end{small}
By drawing samples from the approximate posterior, the integration above can be performed efficiently with Monte Carlo since the cost of each sample is very small as it entails solely CGM output evaluations.
We note though that some of the integrations involved in the computation of lower-order statistics can also be done (semi)-analytically. In particular, we note that the predictive posterior mean $\bs \mu_{\tx{pred}}(\bs \lambda_f)$ is given by:
\be
\begin{split}
\bs \mu_{\tx{pred}}(\bs \lambda_f) & = \int \buf ~  p_{\tx{pred}}(\bs u_f|\bs \lambda_f, \mathcal D) ~d\buf = \bs{W}\left<\buc(\blc)\right> \\
& \approx  \frac{1}{ N_{\tx{samples}}} \bs{W} \sum_{s = 1}^{N_{\tx{samples}}} \bs u_c(\bs \lambda_c^{(s)})
\end{split}
\label{eq:mu_pred}
\ee
where $\left<\buc(\blc)\right>$ denotes the predictive posterior average of the CGM output computed with Monte Carlo by ancestral  sampling. Similarly, one can compute the posterior predictive covariance of the FGM output vector $\buf$, which we denote by $\tx{Cov}_{\tx{pred}}(\blf)$, as
\be
\begin{split}
 &\tx{Cov}_{\tx{pred}}(\blf) = \\
 &= \tx{diag}\left[\left<\bs \tau_{cf}^{-1}\right> \right] + \bs{W} \left<\left(\buc(\blf) - \left<\buc(\blf)\right>\right) \left(\buc(\blf)-\left<\buc(\blf)\right>\right)^T\right> \bs{W}^T \\
 &\approx \frac{1}{ N_{\tx{samples}}} \bs{W} \left( \sum_{s = 1}^{N_{\tx{samples}}} \left(\bs u_c(\bs \lambda_c^{(s)})-\left<\buc\right>\right) \left(\bs u_c(\bs \lambda_c^{(s)})-\left<\buc\right>\right)^T~ \right) \bs{W}^T \\
 &+ \tx{diag}\left[\frac{\tilde{\bs f}}{\tilde{e}-1}\right],
\end{split}
\label{eq:cov_pred}
\ee
where $\left<\left(\buc(\blf)-\left<\buc(\blf)\right>\right)\left(\buc(\blf) - \left<\buc(\blf)\right>\right)^T\right>$ 
denotes the predictive posterior covariance of the CGM output vector $\buc$ which can be estimated by Monte Carlo and $\left<\tau_{cf,i}^{-1}\right>$ correspond to the posterior averages of reciprocal precisions which can be computed in closed form based on \refeq{eq:Qtau_cf}.

The computational steps to generate predictive samples from the generally non-Gaussian $p_{\tx{pred}}(\bs u_f|\bs \lambda_f, \mathcal D)$ for an unseen microstructural input $\bs \lambda_f$ is summarized in Algorithm \ref{alg:prediction}.
\begin{figure}[t]
\centering
\begin{tikzpicture}[auto,node distance=0cm,>=stealth']
\scriptsize
\tikzset{block/.style= {draw, rectangle, rounded corners, minimum height=2em,minimum width=4em}}

\begin{pgfonlayer}{foreground}
\node[] (titleOffline) {Training/offline stage};
\node[block, node distance=12.5mm, fill=white, align=center, below of = titleOffline] (genData)
{Generate training data $\mathcal D$ \\
$\begin{gathered}
\bs \lambda_f^{(n)} \sim p(\bs \lambda_f) \\
\bs u_f^{(n)} = \bs u_f(\bs \lambda_f^{(n)}) \\
\mathcal D = \left\{\bs \lambda_f^{(n)}, \bs u_f^{(n)} \right\}_{n = 1}^N
\end{gathered}$
};

\node[block, below of=genData, node distance=21mm, fill=white, align=center] (evalphiTrain)
{Evaluate features \\ $\varphi_{jm}(\bs \lambda_f^{(n)})$, sec. \ref{sec:p_c}};

\draw[->]     (genData) -- (evalphiTrain);

\node[block, below of=evalphiTrain, node distance=11mm, fill=white] (trainModel)
{Train surrogate, sec. \ref{sec:modelTraining}};

\draw[->]     (evalphiTrain) -- (trainModel);

\node[block, below of=trainModel, node distance=10.5mm, fill=white, align=left] (MAP)
{Output: approximate \\ posterior $Q(\bs \theta) \approx p(\bs \theta| \mathcal D)$};

\draw[->]     (trainModel) -- (MAP);

\node[right of=MAP, node distance=19mm] (refPointA) {};

\draw[-]     (MAP) -- (refPointA.center);

\node[right of = titleOffline, node distance=45mm] (titleOnline) {\qquad\qquad\qquad\qquad\qquad\qquad\quad Prediction/online stage};

\node[block, below of=titleOnline, xshift=-3.5mm, node distance=7.5mm, align=center, fill=white] (sample_theta)
{Average over parameters, \\
$\int p(\bs u_f|\bs \lambda_f, \bs \theta) p(\bs \theta|\mathcal D) d\bs \theta$, \\
sec. \ref{sec:predictions}};

\node[left of=sample_theta, node distance=22.5mm] (refPointB) {};

\draw[-]    (refPointA.center) -- (refPointB.center);

\draw[->]    (refPointB.center) -- (sample_theta);

\node[block, right of =sample_theta, node distance=36mm, fill=white, align=center] (evalphi)
{Evaluate features \\ $\varphi_{jm}(\bs \lambda_{f})$};

\node[block, right of=evalphi, node distance=20mm, align=center] (unseenLambda)
{Unseen \\ $\bs \lambda_{f}$};

\draw[->]   (unseenLambda) -- (evalphi);

\node[block, below of =sample_theta, xshift=29.5mm, node distance=15mm, fill=white, align=center] (sample_lambda)
{Sample $\bs \lambda_c^{(s)} \sim p_c(\bs \lambda_c| \bs \lambda_f, \bs \theta_c)$};

\node[below of=sample_theta, node distance=3.5mm, xshift=14mm] (refPointC0) {};

\node[below of=sample_theta, node distance=13mm, xshift=14mm] (refPointC) {};

\node[below of=evalphi, node distance=13mm] (refPointD) {};

\draw[->]   (refPointC0) -- (refPointC);

\draw[->]   (evalphi) -- (refPointD);

\node[block, below of =sample_lambda, xshift=6.5mm, node distance=14mm, fill=white, align=center] (coarseSolver)
{Solve CGM \\
$\bs u_c^{(s)} = \bs u_c(\bs \lambda_c^{(s)})$};

\node[below of=sample_lambda, node distance=2mm, xshift=6.5mm] (refPoint1) {};

\draw[->]   (refPoint1) -- (coarseSolver);

\node[block, below of =coarseSolver, xshift=0mm, node distance=15mm, fill=white, align=center] (reconstruction)
{Project back \\
$\bs u_f^{(s)} = \bs W \bs u_c^{(s)}$};

\draw[->]   (coarseSolver) -- (reconstruction);
\node[below of=reconstruction, node distance=6.5mm] (refPointE) {};

\draw[-]    (reconstruction) -- (refPointE.center);

\node[block, left of =coarseSolver, yshift=-5mm, node distance=34mm, align=center, fill=white] (repeat)
{Repeat and update \\ QoI $f$ (e.g., $\bs \mu_{\tx{pred}}$), \\ $\left<f\right> \approx \frac{1}{N_{\tx{samples}}} \sum_{s = 1}^{N_{\tx{samples}}} f(\bs u_f^{(s)})$};

\node[left of=refPointE, node distance=35mm] (refPointF) {};

\node[above of=refPointF, node distance=10mm] (refPointF2) {};

\node[above of=refPointF2, node distance=13mm] (refPointF3) {};

\draw[-]    (refPointE.center) -- (refPointF.center);
\draw[-]    (refPointF.center) -- (refPointF2.center);,

\node[above of=refPointF3, node distance=12.5mm] (refPointG) {};

\draw[-]    (refPointF3.center) -- (refPointG.center);

\node[right of=refPointG, node distance=10mm] (refPointH) {};

\draw[->]    (refPointG.center) -- (refPointH.center);
\end{pgfonlayer}{foreground}

\begin{pgfonlayer}{background}
\node[above of = evalphi] (outerOnline) {};
\node[fit= (outerOnline) (evalphi) (sample_theta) (sample_lambda) (coarseSolver) (reconstruction) (repeat) (refPointE), fill=teal!25, opacity=.7, dashed, draw, inner sep=0.1cm, rounded corners] (Box)
{};
\node[above of = genData] (outerOffline) {};
\node[fit= (outerOffline) (genData) (evalphiTrain) (trainModel) (MAP), fill=teal!25, opacity=.7, dashed, draw, inner sep=0.1cm, rounded corners] (Box)            {};
\end{pgfonlayer}{background}

\end{tikzpicture}
\caption{Model workflow training (left) and prediction phases (right).}
\label{fig:workflow}
\end{figure}
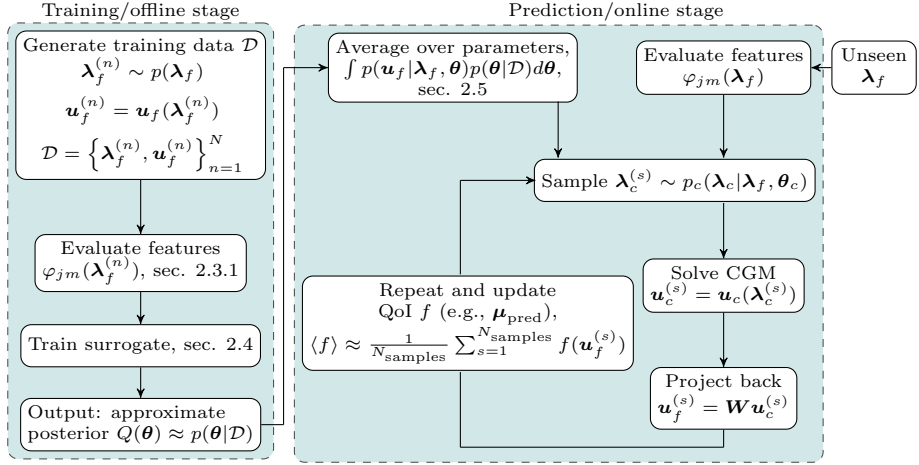
The training (offline) and prediction (online) stages are represented graphically in the workflow diagram in Figure \ref{fig:workflow}.

\begin{algorithm}[h]
\KwData{$\bs \lambda_{f, \tx{new}}$ \tcp*{Test microstructure $\bs \lambda_{f, \tx{new}}$}}
\KwIn{$p(\tilde{\bs \theta}_c, \bs \tau_c, \bs \tau_{cf}|\mathcal D) \approx Q_{\tilde{\bs \theta}_c}(\tilde{\bs \theta}_c) Q_{\bs \tau_c}(\bs \tau_c) Q_{\bs \tau_{cf}}(\bs \tau_{cf})$ \tcp*{(Approximate) posterior}}

\vspace{3mm}
\KwOut{Predictive sample $\bs u_f^{(s)} \sim p_{\tx{pred}}(\bs u_f|\bs \lambda_{f, \tx{new}}, \mathcal D)$, eq. \eqref{eq:p_pred};}

\vspace{2mm}
Evaluate all feature functions $\bs \varphi_m(\bs \lambda_{f, \tx{new}})$;

Sample $\bs \tau_{cf}^{(s)} \sim Q_{\bs \tau_{cf}}(\bs \tau_{cf}),~ \bs \tau_c^{(s)} \sim Q_{\bs \tau_c},~ \tilde{\bs \theta}_c^{(s)} \sim Q_{\tilde{\bs \theta}_c}(\tilde{\bs \theta}_c)$;

Sample $\bs \lambda_c^{(s)} \sim p_c(\bs \lambda_c|\bs \varphi_m(\bs \lambda_{f, \tx{new}}), \tilde{\bs \theta}_c^{(s)}, \bs \tau_c^{(s)})$;

Solve CGM $\bs u_c^{(s)} = \bs u_c(\bs \lambda_c^{(s)})$;

Draw predictive sample $\bs u_f^{(s)} \sim p_{cf}(\bs u_f|\bs W, \bs \tau_{cf}^{(s)})$;

\Return{\tx{Predictive sample} $\bs u_f^{(s)}$}
\caption{Generation of predictive samples.}
\label{alg:prediction}
\end{algorithm}

\subsubsection{Model performance metrics}
\label{sec:perfMetrics}
Apart from computational efficiency, the key objective of any surrogate model is its predictive accuracy, i.e. its closeness to the true FGM  solution $\bs u_f(\bs \lambda_f)$. In order to assess this, we consider separate test datasets $\mathcal D_{\tx{test}} =$ $\left\{\bs \lambda_f^{(n)}, \bs u_f^{(n)} \right\}_{n = 1}^{N_{\tx{test}}}$, where the $\bs \lambda_f^{(n)} \sim p(\bs \lambda_f)$ are drawn from the same distribution $p(\bs \lambda_f)$ that was used for the training set $\mathcal D$. We report metrics that reflect both the accuracy of point estimates of the predictive posterior  (e.g. $\bs{\mu}_{\tx{pred}}(\blf)$ in \refeq{eq:mu_pred}) as well as the whole density $p_{\tx{pred}}(\buf |\blf, \mathcal{D})$ (\refeq{eq:p_pred}).

In particular, we compute the \tit{coefficient of determination} $R^2$ \cite{Zhang2017} 
\begin{equation}
    R^2 = 1 - \frac{\sum_{n = 1}^{N_{\tx{test}}} \Vert \bs u_f^{(n)} - \bs \mu_{\tx{pred}}(\bs \lambda_f^{(n)}) \Vert^2}{\sum_{n = 1}^{N_{\tx{test}}} \Vert \bs u_f^{(n)} - \bar{\bs u}_f\Vert^2},
    \label{eq:R2}
\end{equation}
where $\bar{\bs u}_f = \frac{1}{N_{\tx{test}}}\sum_{n = 1}^{N_{\tx{test}}} \bs u_f^{(n)}$ is the sample mean of the FGM output $\bs u_f$ over the test dataset. The denominator normalizes the point deviation in the numerator by the actual variability of the true outputs.
In the case of perfect prediction, $\bs \mu_{\tx{pred}}(\bs \lambda_f^{(n)}) = \bs u_f^{(n)} ~\forall n$, the second term in \eqref{eq:R2} vanishes and $R^2 = 1$ which is the largest value it can attain. Smaller values of $R^2$ indicate either larger deviation of the mean prediction from the truth or smaller variability in the $\buf$ components.

 The second metric we employ is the \tit{mean log likelihood} $MLL$ \cite{Zhu2018} 
given by
 \begin{equation}
    \begin{split}
        MLL &= \frac{1}{N_{\tx{test}} \tx{dim}(\bs u_f)} \sum_{n = 1}^{N_{\tx{test}}} \log p_{\tx{pred}}(\bs u_f(\bs \lambda_f^{(n)})|\bs \lambda_f^{(n)}, \mathcal D). \\
  \end{split}
    \label{eq:MLLref}
\end{equation}
We note that when $\log p_{\tx{pred}}(\bs u_f |\bs \lambda_f, \mathcal D)$ degenerates to a discrete density  centered at the true $\buf(\blf), ~\forall \blf$ i.e. not only it predicts perfectly the FGM output but there is no predictive uncertainty, then the $MLL$ attains the highest possible value of $0$. In the other extreme if for one or more of the test samples $\bs \lambda_f^{(n)}$, $p_{\tx{pred}} \to 0$ at the true  $\buf(\blf)$, then $MLL$ attains its lowest possible value of $-\infty$.
Hence, $MLL$ attempts to measure the quality of the whole predictive posterior $p_{\tx{pred}}$. 
Given that the integration in \refeq{eq:p_pred} is analytically intractable and no closed-form expression for $p_{\tx{pred}}$ can be established, we evaluate $MLL$ under the assumption that $p_{\tx{pred}}$ can be adequately approximated by a multivariate Gaussian with mean $\bs{\mu}_{\tx{pred}}$ (\refeq{eq:mu_pred}) and a diagonal covariance which consists of the diagonal entries of $\tx{Cov}_{\tx{pred}}(\blf)$ (\refeq{eq:cov_pred}) which we represent with the vector $\bs \sigma^2_{\tx{pred}}(\blf)$.
In this case:
\begin{equation}
    \begin{split}
        MLL &= \frac{1}{N_{\tx{test}} \tx{dim}(\bs u_f)} \sum_{n = 1}^{N_{\tx{test}}} \log p_{\tx{pred}}(\bs u_f(\bs \lambda_f^{(n)})|\bs \lambda_f^{(n)}, \mathcal D) \\
        &\approx -\frac{1}{2}\log 2\pi - \frac{1}{N_{\tx{test}} \tx{dim}(\bs u_f)}\sum_{n = 1}^{N_{\tx{test}}}  \Biggl(\frac{1}{2} \sum_{i = 1}^{\tx{dim}(\bs u_f)} \log \sigma_{\tx{pred, i}}^2(\blf^{(n)}) \\
        &\quad - \frac{1}{2} \sum_{i = 1}^{\tx{dim}(\bs u_f)} \frac{\left(u_{f, i}(\bs \lambda_f^{(n)}) - \mu_{\tx{pred}, i}(\bs \lambda_f^{(n)}) \right)^2}{\sigma_{\tx{pred, i}}^2(\blf^{(n)})} \Biggr)
    \end{split}
    \label{eq:MLL}
\end{equation}
In the numerical experiments discussed in section \ref{sec:experiments},  $N_{\tx{test}} = 1024$ is used, leading to negligible Monte Carlo error due to variation of test samples.

\subsection{Numerical complexity analysis}
\label{sec:complexity}
\begin{table}[t]
\begin{small}
\begin{center}
\begin{tabular}{l||c|c|p{16mm}|c|c|c}
\begin{tabular}{c} Quantity \\ \hline Phase  \end{tabular} & $N$ & $\tx{dim}(\bs u_f)$ & \centering $\tx{dim}(\bs \lambda_f)$ & $\tx{dim}(\bs u_c)$ & $\tx{dim}(\bs \lambda_c)$ & $\tx{dim}(\tilde{\bs \theta}_{c, m})$ \\
\hline
\hline
training & $N$ & $\tx{dim}(\bs u_f)$ & \centering $1 \ldots$ $(\tx{dim}(\bs \lambda_f))^2$ & $\tx{dim}(\bs u_c)$ & $\tx{dim}(\bs \lambda_c)$ & \begin{tabular}{c} $\tx{dim}(\tilde{\bs \theta}_{c, m})$ \end{tabular} \\
\hline
prediction & $1$ & $\tx{dim}(\bs u_f)$ & \centering $1 \ldots$ $(\tx{dim}(\bs \lambda_f))^2$ & $\tx{dim}(\bs u_c)$ & $\tx{dim}(\bs \lambda_c)$ & $\tx{dim}(\tilde{\bs \theta}_{c, m})$
\end{tabular}
\end{center}
\end{small}
\caption{Computational complexity of training and prediction stages. The scaling with $\tx{dim}(\bs \lambda_f)$ is dependent on the applied set of feature functions $\bs \varphi_m(\bs \lambda_f)$ and reaches from $\mathcal O(1)$ (e.g. for $\varphi_m(\bs \lambda_f) = 1$) to $\mathcal O\left((\tx{dim}(\bs \lambda_f))^2\right)$ (e.g. $\varphi_m(\bs \lambda_f) = $ mean mutual distance of exclusion centers).}
\label{tab:complexity}
\end{table}
For analysis of the numerical complexity of the proposed model, it is vital to distinguish between training (offline) and prediction (online) phases, see also Figure \ref{fig:workflow}. There are basically six quantities that are relevant for the computational cost of both stages. These are the number of training data $N$, the dimension of the FGM input $\tx{dim}(\bs \lambda_f)$, the dimension of the FGM output  $\tx{dim}(\bs u_f)$, the dimension of CGM input $\tx{dim}(\blc)$, the dimension of the CGM output $\tx{dim}(\bs u_c)$ and the number of feature functions per latent space variable $\tx{dim}(\tilde{\bs \theta}_c)$.

As it can be seen in Algorithm \ref{alg:modelTraining}, the complexity of the training stage depends linearly on the number of training data $N$ (for-loop), but is fully parallelizable over every data point. The prediction stage only sees the final (approximate) posterior $p(\tilde{\bs \theta}_c, \bs \tau_c, \bs \tau_{cf}|\mathcal D) \approx Q_{\tilde{\bs \theta}_c}(\tilde{\bs \theta}_c) Q_{\bs \tau_c}(\bs \tau_c) Q_{\bs \tau_{cf}}(\bs \tau_{cf})$, so that the scaling is independent of $N$. Both training and prediction phases scale linearly with the fine scale output dimension $\tx{dim}(\bs u_f)$, since they only involve vector operations. 
For the same reason, the scaling with respect to $\tx{dim}(\bs \lambda_c)$ is linear. 
Scaling with $\tx{dim}(\bs u_c)$ depends on the CGM solver i.e. in the worst case, for the linear Darcy model with a direct 
 solver, the scaling is  
$\mathcal O\left(\tx{dim}((\bs u_c))^3\right)$. We apply a sparse banded solver, which scales as $\mathcal O\left(\tx{dim}((\bs u_c))\right)$.
 Scaling with the dimension of microstructural inputs $\tx{dim}(\bs \lambda_f)$ is dependent on the set of applied feature functions $\bs \varphi_m$  and ranges from $\mathcal O(1)$ (e.g. constant feature, i.e. $\varphi(\bs \lambda_f) = 1$) to $\mathcal O\left((\tx{dim}(\bs \lambda_f))^2\right)$ (e.g. mean mutual distance of circular exclusions).
 Finally, scaling with respect to  $\tx{dim}(\tilde{\bs \theta}_{c, m})$ is $\mathcal O\left((\tx{dim}(\tilde{\bs \theta}_{c, m}))\right)$ see \refeq{eq:Qtheta}. 
 Furthermore, in the prediction stage, it is possible to omit components of $\tilde{\bs \theta}_{c, m}$ that have been pruned out by the applied sparsity prior, thereby avoiding the cost connected to the evaluation of corresponding feature functions. 
 The scaling characteristics of training and prediction phases is summarized in table \ref{tab:complexity}.

\subsection{Automatic  adaptive refinement}
\label{sec:aar}
As already mentioned in section \ref{sec:modelTraining}, the evidence lower bound (ELBO) is an approximation to the model's $\log-$evidence. The latter balances the model's fit to the data with the model's complexity and 
 serves  as a parsimonious metric of how well the model can explain the training data \cite{murray_note_2005,rasmussen_occams_2001}. This quantity can therefore be used not only for scoring different competing models which have been trained on the same data, but also as the objective to any refinement of an existing model. 
 According to \refeq{eq:elbo}, the ELBO is given by the expected value
\begin{equation}
\mathcal F(Q) = \left<\log p(\bs \theta, \mathcal D) - \log Q(\bs \theta) \right>_{Q(\bt)},
\label{eq:elbo_exp}
\end{equation}
where $p(\bs \theta, \mathcal D)$ is  given in \refeq{eq:fullPosterior} and $Q(\bs \theta)$ is the  variational approximation to the posterior $p(\bt | \mathcal{D})$ as defined in \eqref{eq:meanfield}. Since all of the expected values needed in the right side of the above equation are estimated during model training, the ELBO $\mathcal F(Q)$ can be evaluated explicitly and be used to also  monitor convergence behavior during  training. In order to keep this section uncluttered, the final expression for the ELBO is given in \ref{ap:elbo}.

While one can envision many ways of refining the coarse-grained model, we focus here on the latent variables $\blc$. These constitute the bottleneck through which information that the FGM input $\blf$  provides about the FGM output $\buf$ is squeezed. Physically, they relate to the effective permeability of the Darcy-based CGM and assuming a model according to equations \eqref{eq:effPerm}--\eqref{eq:simpleK_l}, they are inherently tied with the piecewise-constant discretization adopted, i.e. each component $\lambda_{c,m}$ relates to the Darcy permeability in a square subregion $\Omega_m$ of the problem domain.

The question we would like to address is the following: If  one was to refine the piecewise-constant representation of the Darcy permeability field in equations \eqref{eq:effPerm}--\eqref{eq:simpleK_l} by subdividing one square cell $\Omega_m$ of the existing partition  into smaller subregions, then  which $\Omega_m$  would yield the best possible model? We note that such a refinement would lead to an increase in the dimension $\tx{dim}(\blc)$ of the bottleneck variables which, in principle, should lead to more information form $\blf$ being retained. 

Given a fixed refinement mechanism of an $\Omega_m$, say by subdividing into 4 equal squares (in two-dimensions), a brute force solution strategy would be to refine one of the existing $\Omega_m$ at a time, train the new model, compute its ELBO and compare with the others. Obviously such a strategy is computationally cumbersome particularly when the number of existing subregions $N_{\tx{cells,c}}$ is high.
The cost would be even greater if one was to consider all possible combinations of two or more subregions that could be refined at the same time.

We thus propose a procedure based on the decomposition of the ELBO $\mathcal F(Q)$ into contributions from different subregions $\Omega_m$ (and therefore $\lambda_{c,m}$) according to $\mathcal F(Q) = \sum_{m = 1}^{N_{\tx{cells,c}}} \mathcal F_m(Q) + \mathcal H(Q)$, where with $\mathcal H(Q)$ we denote terms that cannot be assigned to any of the subregions $\Omega_m$.
As the goal is ultimately to increase the ELBO of the model, we propose selecting for refinement the cell $m$ with the {\em lowest} contribution $\mathcal F_m(Q)$ and therefore contributes the least in explaining the data.   This strategy is analogous to the one advocated in \cite{ghahramani_variational_2000} for  refining components in a mixture of factor analyzers. Hence we use $\mathcal F_m(Q)$ as the scoring function of each subregion/cell. 
The derivation of $\mathcal F_m(Q)$ is contained in \ref{ap:elbo}.
and we include here only the final expression which, for a model with tied $\gamma_{jm} = \gamma_j$, is
\begin{equation}
\mathcal F_m(Q) = \sum_{n = 1}^N \log \sigma_{\lambda_{c, m}}^{(n)} - \tilde{c} \log \tilde{d}_m + \sum_{j = 1}^{\tx{dim}(\bs \gamma)} \log \sigma_{\tilde{\theta}_{c, jm}}.~~~
\label{eq:cellScore}
\end{equation}
where $(\sigma_{\lambda_{c, m}}^{(n)})^2$ is the posterior variance of $\lambda_{c,m}^{(n)}$ (see equations \eqref{eq:qlambda} and \eqref{eq:optqlambda}), $\tilde{c}, \tilde{d}_m$ are given in \refeq{eq:Qtau_c} and $\sigma_{\tilde{\theta}_{c, jm}}^2$ is the posterior variance of the parameters $\theta_{c,jm}$  given in \refeq{eq:Qtheta}.

Keeping in mind that  the subregion/cell $\Omega_m$ with the {\em smallest} $\mathcal F_m$ is proposed for refinement, the scoring function defined by Equation \eqref{eq:cellScore} favours cells where the posterior variances $(\sigma_{\lambda_{c, m}}^{(n)})^2$ of the $\lambda_{c, m}^{(n)}$ tend to be small, i.e. tight (approximate) posterior distribution $Q_{\lambda_{c, m}^{(n)}}(\lambda_{c, m}^{(n)})$, whereas at the same time, the expected decoder misfit $\sum_{n = 1}^N \left< \left(\lambda_{c, m}^{(n)} - \tilde{\bs \theta}_{c, m}^T \bs \varphi_m(\bs \lambda_f^{(n)}) \right)^2\right>$ (see Equation \eqref{eq:Qtau_c}) should be large but there should be high certainty about the feature coefficients $\tilde{\bs \theta}_{c, m}$, i.e. low $\sigma_{\tilde{\theta}_{c, jm}}^2$.
We finally note that once the cell $m$ with the lowest $\mathcal F_{m}(Q)$ has been identified and split, training is restarted  with all model parameters initialized to their previously converged values. We illustrate  these steps by example in section \ref{sec:expref}.

\section{Numerical examples}
\label{sec:experiments}
The numerical experiments carried out in this paper are aimed at replacing the expensive Stokes flow FE simulation defined by Equations \eqref{eq:StokesPDE}--\eqref{eq:FGM} by an inexpensive,  Bayesian, coarse-grained model built around the Darcy-flow skeleton as presented in sections \ref{sec:darcy}--\ref{sec:complexity}. We begin this section with a clear specification of the FGM data generation  and the model distributions $p_c, p_{cf}$. Subsequently, we show that close to the homogenization limit (i.e. infinite scale separation) defined by \refeq{eq:lengthScales}, the proposed coarse-grained model is capable to accurately reconstruct the FGM solution by learning (from a small number of FGM training samples) the requisite effective properties {\em without} ever solving any of the equations homogenization theory describes. Thereafter, we investigate the coarse-grained model's predictive quality  as a function of the  number of training data $N$ and the latent space dimensionality $\tx{dim}(\bs \lambda_c)$ for microstructural data far from the homogenization limit (without scale separation). We show that the model indeed exhibits feature sparsity in learning the effective Darcy permeability field $\bs K(\bs x, \bs \lambda_c)$. We then apply the surrogate to an uncertainty propagation (UP) problem and finally demonstrate the performance of the proposed  adaptive refinement objective and associated algorithm. Source code as well as examples can be found at: \href{https://github.com/congriUQ/physics_aware_surrogate}{https://github.com/congriUQ/physics\_aware\_surrogate}

\subsection{Experimental setup}
\subsubsection{Fine scale microstructural data}
\label{sec:data}
\begin{figure}[!t]
\centering
\includegraphics[width=.9\textwidth]{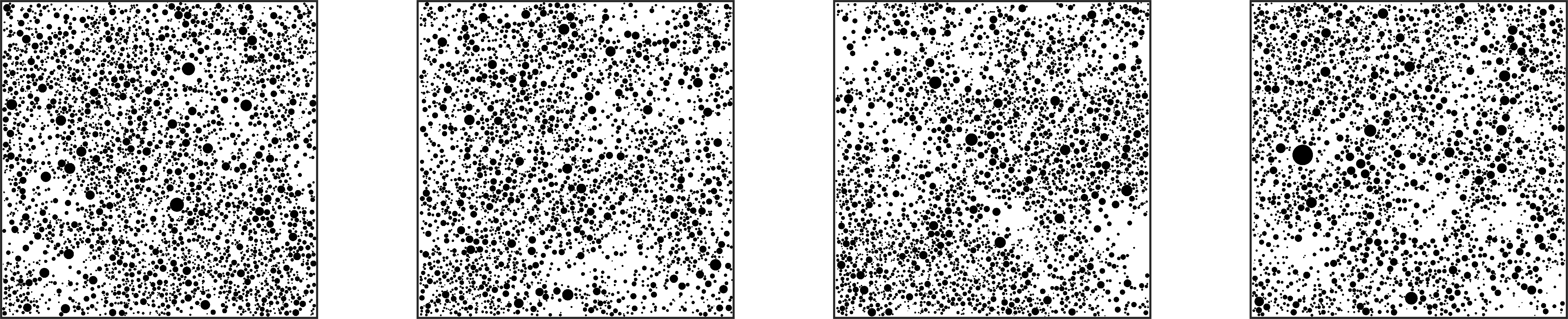}
\caption{Microstructures generated as described in section \ref{sec:data}. The corresponding parameters are $\mu_{\tx{ex}} = 7.8$, $\sigma_{\tx{ex}} = 0.2$, $\sigma_r = 0.3$, $l_s = 1.2$, $l_{\bs x} = 0.08$ and $l_r = 0.05$.}
\label{fig:microstructures}
\end{figure}
Throughout the following experiments, we use randomly distributed, non - overlapping polydisperse spherical exclusions from a unit square domain $\Omega = [0, 1]^2$ as depicted  in Figure \ref{fig:microstruct}. The total number of exclusions $N^{(n)}_{\tx{ex}}$ of a material sample $n$ as well as the radii $r_{\tx{ex}, i}^{(n)}, i = 1, \ldots, N^{(n)}_{\tx{ex}}$, are sampled from a log-normal distribution,
\begin{align}
    \label{eq:Nex}
    N_{\tx{ex}}^{(n)} &\sim \tx{round}\left[\tx{Lognormal}\left(\mu_{\tx{ex}}, \sigma_{\tx{ex}}^2\right)\right], \\
    r_{\tx{ex}, i}^{(n)} &\sim \tx{Lognormal}\left(\mu_r^{(n)}(\bs x), \sigma_r^2\right).
    \label{eq:rex}
\end{align} 
The coordinates of the center of each exclusion  $\bs x_{\tx{ex}, i}^{(n)}$ are drawn from a Gaussian process with squared exponential kernel of length scale $l_{\bs x}$ warped with the logistic sigmoid function $S(z) = 1/(1 + e^{-l_s z})$,
\begin{equation}
    \bs x_{\tx{ex}, i}^{(n)} \sim \frac{1}{N_{\rho^{(n)}}}\rho^{(n)}_{\bs x_{\tx{ex}}}(\bs x), \qquad \rho^{(n)}_{\bs x_{\tx{ex}}}(\bs x) \sim S\left(GP(0, k_{\bs x}(\bs x - \bs x'))\right),
\end{equation}
where $N_{\rho^{(n)}} = \int \rho^{(n)}_{\bs x_{\tx{ex}}}(\bs x) \bs x$ and a squared exponential covariance $k_{\bs x} =$ \newline $\exp\left\{-\left\vert\bs x - \bs x'\right\vert^2/l_{\bs x}^2 \right\}$\footnote{Note that the resulting distribution is distorted by the fact that a sampled exclusion is rejected if it overlaps with an exclusion which has been inserted before.}. Also, the (location-dependent) $\log$-normal mean of exclusion radii $\mu_r(\bs x)$ is drawn from a Gaussian process for every sample according to
\begin{equation}
    \mu_r^{(n)}(\bs x) \sim GP(0, k_r(\bs x - \bs x')),
\end{equation}
with $k_{r} = \exp\left\{-\left\vert\bs x - \bs x'\right\vert^2/l_{r}^2 \right\}$ the corresponding covariance kernel. Four microstructural samples generated according to the described distributions are depicted in Figure \ref{fig:microstructures}. After generation of a microstructure, the domain is pixelized into a grid of size $256\times 256$. The corresponding binary vector representing the phase of each pixel makes up the FGM input $\blf$, i.e.:
\be
dim(\blf)=65,536.
\ee
The sampled microstructres are  passed to the FEniCS mshr \cite{Fenics2012} module to generate a triangular finite element mesh containing roughly $256\times 256 = 65,536$ vertices.
The output of interest, i.e. $\buf$, consists of the the interpolated pressure values on a regular grid of $129\times 129$ i.e.:
\be 
dim(\buf)=16,641.   
\ee
 Depending on the total number of exclusions $N_{\tx{ex}}^{(n)}$, generation of a single mesh takes in between $\sim 2$ hours (for $N_{\tx{ex}}^{(n)} \approx 1,000$) up to 10 days (for $N_{\tx{ex}}^{(n)} \approx 20,000$) of computation time on a single Intel Xeon E5-2620 (2.00 GHz) CPU. The average  solution for each FGM (i.e. Stokes-flow) PDE-solve takes 1512s $\pm$ 5.7s on the same hardware.

We use boundary conditions on the FGM pressure $P_{bc}(\bs x)$ and velocity fields $\bs V_{bc}(\bs x)$ of the  form
\begin{equation}
\begin{split}
    P_{bc}(\bs x) &= 0, \\
    \bs V_{bc}(\bs x) &= \bmat a_x + a_{xy} y \\ a_y + a_{xy} x \emat, 
\end{split}
\qquad \qquad
\begin{split}
    &\tx{for} \quad \bs x \in \Gamma_P = \bs 0, \\
    &\tx{for} \quad \bs x \in \Gamma_{\bs V} = \pa \Omega\backslash \bs 0,
\end{split}
\label{eq:bc}
\end{equation}
where $\Gamma_P = \bs 0$ denotes the origin of the domain, i.e. the point $\bs x = \bs 0$. The boundary condition coefficients $a_x, a_y$ and $a_{xy}$ may assume different values in the following experiments and are therefore specified later.

\subsubsection{The CGM  \texorpdfstring{$\bs u_c(\bs \lambda_c)$}{}}
\label{sec:CGM}
For the Darcy-type reduced order model $\bs u_c(\bs \lambda_c)$, we use a finite-element solver to Equations \eqref{eq:Darcy} as described in section \ref{sec:darcy}, operating on a regular $16\times 16$ square grid endowed with bilinear shape functions. The coarse-grained random permeability field $\bs K(\bs x, \bs \lambda_c)$ of the CGM Darcy solver is assumed to have the form defined by equations \eqref{eq:effPerm}--\eqref{eq:simpleK_l} throughout this section.

\subsubsection{The encoder distribution \texorpdfstring{$p_c$}{}}
For the encoder distribution $p_c(\bs \lambda_c| \bs \lambda_f, \bs \theta_c)$, we assume a model as defined by equations \eqref{eq:p_cModel}--\eqref{eq:p_c}, with the restriction that
\begin{equation}
\begin{split}
\lambda_{c, m} &= \sum_{j = 1}^{N_{\tx{glob}}} \tilde{\theta}_{c, jm} \varphi_{\tx{glob},j}(\bs \lambda_f) + \sum_{j = N_{\tx{glob}} + 1}^{N_{\tx{glob}} + N_{\tx{loc}}} \tilde{\theta}_{c, jm} \varphi_{\tx{loc},j}(\bs \lambda_f^{(m)}) + \sigma_{c, m} Z_m, \\
Z_m &\sim \mathcal N(0, 1),
\end{split}
\label{eq:p_cExperiments}
\end{equation}
where $\bs \lambda_f^{(m)}$ denotes the subset of the microstructure $\bs \lambda_f$ which corresponds to the permeability field cell $\Omega_m$. \refeq{eq:p_cExperiments} essentially states that the same set of features $\bs \varphi_m(\bs \lambda_f) = \bs \varphi(\bs \lambda_f) = \left\{\bs \varphi_{\tx{glob}}(\bs \lambda_f), \bs \varphi_{\tx{loc}}(\bs \lambda_f^{(m)}) \right\}$ are used for every permeability field cell $m$, where $\bs \varphi_{\tx{glob}}, \bs \varphi_{\tx{loc}}$ denote features evaluated on the global domain and on a local cell permeability field cell $m$, respectively. It is noted though that the coefficients $\tilde{\theta}_{c, jm}$ are generally different for different cells $m$. A Table of all feature functions used in the experiments is given in \ref{sec:features}.

\paragraph{Prior on \texorpdfstring{$\tilde{\theta}_{c, jm}$}{}}
The prior defined in \refeq{eq:priorTheta} is restricted such that the precision hyperparameters $\gamma_{jm} = \gamma_{j}$, i.e. all permeability field cells $m$ share the same set of hyperparameters $\bs \gamma$. This leads to the same set of activated features, i.e. features with nonzero, but generally different $\tilde{\theta}_{c, jm}$ for different cells $m$, such that permeability cells are enabled to share information across the whole domain $\Omega$.

\subsubsection{The decoder distribution \texorpdfstring{$p_{cf}$}{}}
 The decoder distribution we use in the experiments is defined by equations \eqref{eq:p_cf}--\eqref{eq:p_cfGauss}. As the FGM model output $\bs u_f$, we choose to use the Stokes flow pressure response $u_{f, i} = P(\bs x_i)$ linearly interpolated onto a square rectangular grid $\bs x_i \in G^{(f)} = 129 \times 129$ in the unit square domain $\Omega = [0, 1]^2$. Although parameter uncertainty on the coarse-to-fine projection matrix $\bs W$ could readily be included to the model by specifying a suitable prior distribution $p(\bs W)$, it has proven to be beneficial (in terms of predictive quality) in the low-data regime $N \lesssim 100$ to fix $\bs W$ to the CGM Darcy solver shape function interpolant, i.e.
\begin{equation}
    W_{ij} = \psi_{c, j}(\bs x_i),
\end{equation}
where $\psi_{c, j}(\bs x)$ is the Darcy solver finite element shape function associated with the CGM degree of freedom $u_{c, j}$. The precisions $\tau_{cf, i}$ remain a free parameter and give the inverse reconstruction variance for $u_{f, i} = P(\bs x_i)$.

\subsection{Case 1: Scale-separation (homogenization limit) \& high data variability}
\label{sec:homLimit}
\begin{figure}[!t]
\centering
\includegraphics[width=\textwidth]{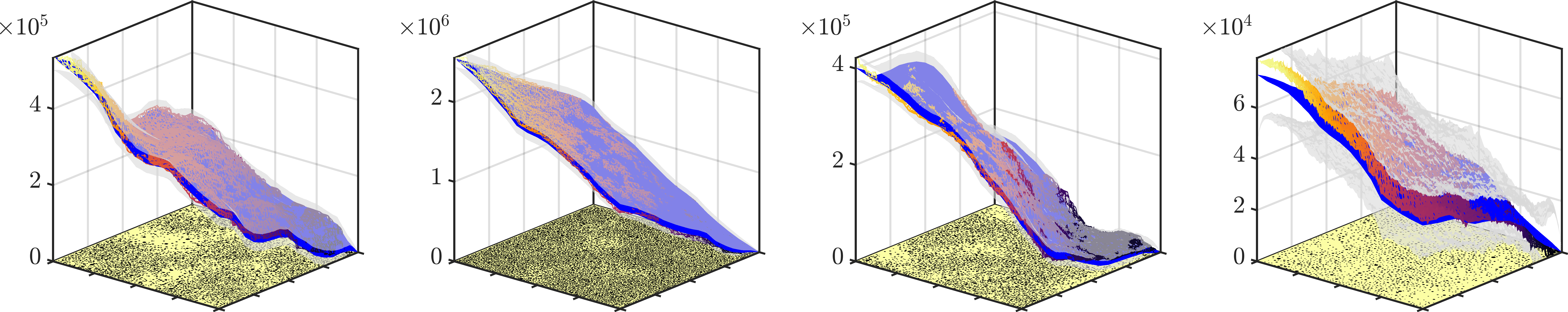}
\caption{Prediction examples for data with $\mu_{\tx{ex}} = 8.35$, $\sigma_{\tx{ex}} = 0.6$, $\sigma_r = 0.5$, $l_s = 1.5$, $l_{\bs x} = 0.1$ and $l_r = 0.05$, $N = 128$ training samples and a $\tx{dim}(\bs \lambda_c) = 4\times 4$ Darcy CGM. The colored surface is the true response $P(\bs x)$ of the test sample and the blue is the predictive mean $\bs \mu_{\tx{pred}}$ of the surrogate. The transparent gray surfaces are the predictive standard deviations $\bs \mu_{\tx{rped}}\pm \bs \sigma_{\tx{pred}}$. Note the different scales on the $z$-axis. Note also that the higher the exclusion numbers $N_{\tx{ex}}^{(n)}$ i.e. the smaller the pore phase length scales $\ell_f$ (see the two samples on the left), the better predictions are. This is because such data is closer to  the homogenization limit as defined by \refeq{eq:lengthScales}.}
\label{fig:predHomData}
\end{figure}

To reveal the full potential of the proposed approach, we use input data that contains microstructures $\bs \lambda_f$ that fulfill the homogenization limit \refeq{eq:lengthScales} as much as it is computationally affordable, whilst having large variability, i.e. high $\sigma_{\tx{ex}}^2, \sigma_r^2$, see equations \eqref{eq:Nex}--\eqref{eq:rex}.
We therefore sample FGM inputs $\blf$  from the model presented earlier  with the following parameter values:
\begin{align}
\nonumber
    \mu_{\tx{ex}} &= 8.35, & \sigma_{\tx{ex}} &= 0.6, \\
    \mu_r &= -5.53, & \sigma_r &= 0.5, \\
    \ell_s &= 1.5, & \ell_{\bs x} &= 0.1,
\nonumber
\end{align}
The boundary conditions employed are of the form given in \refeq{eq:bc} with
\be
a_x = a_y = 1, a_{xy} = 0.
\ee
We generate three training data sets of size $N = 128$ and report the averages of the performance metrics defined in \ref{sec:perfMetrics} over the three models trained. We find
\be 
R^2 = 0.995 \pm 0.004, \quad \tx{and} \quad MLL = -11.003 \pm 0.013
\ee
evaluated on a test set of size $N_{\tx{test}} = 1024$. Predictions for four random test samples are illustrated in Figure \ref{fig:predHomData}. 
We use $\tx{dim}(\bs \lambda_c) = 16$ corresponding to a uniform $4\times 4$ square discretization of the problem domain $\Omega$.
 
In \ref{sec:POD}, we report the results on the PCA decomposition of the FGM outputs $\buf$. One observes a rapid decay of the corresponding eigenvalues in Figure \ref{fig:POD} which would suggest that the problem could be addressed by methods that project the governing equations on the low-dimensional subspace spanned by the first few eigenvectors (e.g. reduced-basis (RB) methods \cite{Hesthaven2016, Quarteroni2016}).
We would like to emphasize though 
that it would be unclear how the Stokes/Darcy homogenization process should be accomplished for the high-dimensional random porous domain described by $\bs \lambda_f$, how to find the RB solution coefficients and if a standard Galerkin projection procedure would indeed be computationally more efficient. Moreover, we note that a RB-based Darcy CGM could readily be included as a substitute for the $\bs u_c(\bs \lambda_c)$ CGM as was defined in section \ref{sec:darcy}, which might be a possible extension to investigate in the future. Apart from that, we emphasize that fast decay of PCA eigenmodes of the \tbf{output} data $\bs u_f$ does not affect the high effective dimension of \tbf{inputs} $\bs \lambda_f$.

\subsection{Case 2: No scale separation (far from homogenization limit) and high output variability}
\label{sec:predPerformance}
\begin{figure}[t]
\centering
\includegraphics[width=\textwidth]{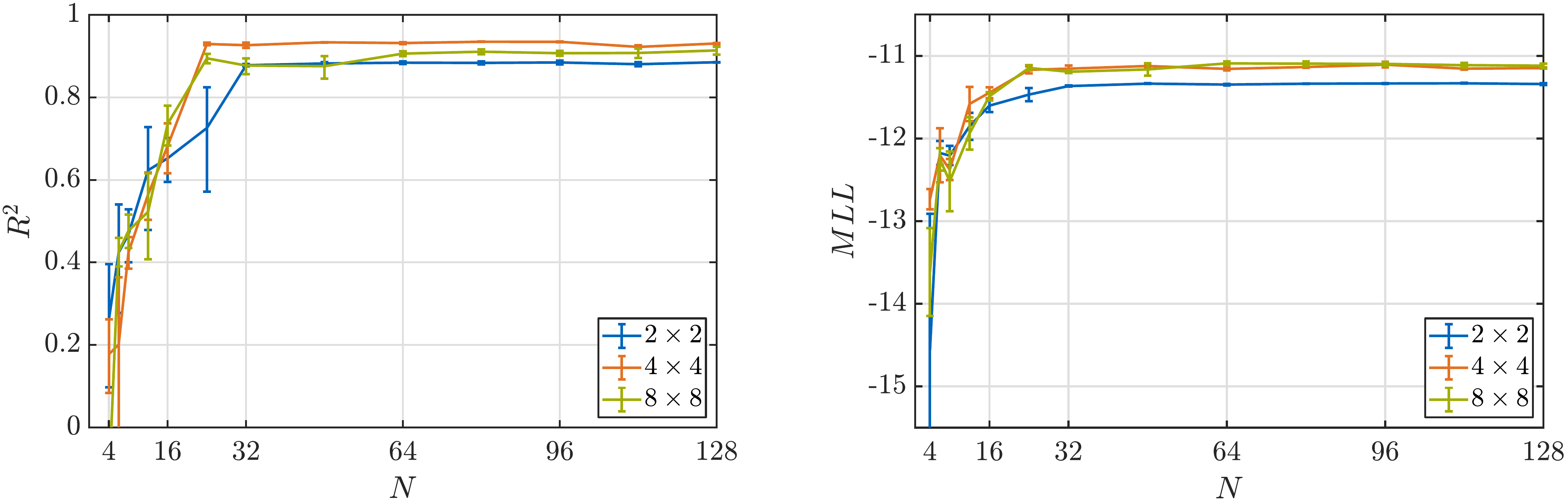}
\caption{Error measures \tit{coefficient of determination} $R^2$ and \tit{mean log likelihood} $MLL$ as defined in equations \eqref{eq:R2}--\eqref{eq:MLL} as a function  of the number of training samples $N$ and for different discretization of the effective permeability field $\bs K(\bs x, \bs \lambda_c)$. The error metrics are computed on $N_{\tx{test}} = 1024$ test samples. The error bars are due to randomization of training data sets (errors due to variation of the test set can be neglected).}
\label{fig:errorPlot}
\end{figure}

In this setting, we consider microstructures that lead to PCA decompositions of the FGM outputs $\buf$ that are not concentrated on a very small subset of eigenvectors.
In particular we generate microstructures based on the following parameters:
\begin{align}
\nonumber
    \mu_{\tx{ex}} &= 7.8, & \sigma_{\tx{ex}} &= 0.2, \\
    \mu_r &= -5.23, & \sigma_r &= 0.3, \\
    \ell_s &= 1.2, & \ell_{\bs x} &= 0.08,
\nonumber
\end{align}
see Figure \ref{fig:POD} for a PCA analysis of the output data $\left\{\bs u_f^{(n)} \right\}_{n = 1}^{2048}$, revealing the effective output dimension.
Boundary conditions are again fixed to $a_x = a_y = 1, a_{xy} = 0$. We consider three cases in terms of $\tx{dim}(\blc)$ which correspond to regular, square discretization of size   $2\times 2$ (i.e. $\tx{dim}(\blc) = 4$), $4\times 4$ (i.e. $\tx{dim}(\blc) = 16$) and $8\times 8$ (i.e. $\tx{dim}(\blc) = 64$). While the PDE discretization is independent of the representation of $\blc$, it needs to be able to resolve the corresponding Darcy permeability field. To that end, we employ a regular square mesh of $16\times 16$ bilinear elements for the CGM model, leading to $\tx{dim}(\buc) = 288$. 
One CGM evaluation requires $(1.04\pm 2) \times 10^{-3}~s$ of computational time on a single Intel Xeon E5-2620 (2.00 GHz) CPU, as opposed to the $1512 \pm 5.7s$ needed for each  FGM solve. For predictive purposes (section \ref{sec:predictions}), we use $N_{\tx{samples}} = 100$ for the estimation e.g. of $\bs \mu_{\tx{pred}}, \bs \sigma_{\tx{pred}}^2$ as defined in Equations \eqref{eq:mu_pred} and \eqref{eq:cov_pred}.

\begin{figure}[t]
\centering
\includegraphics[width=\textwidth]{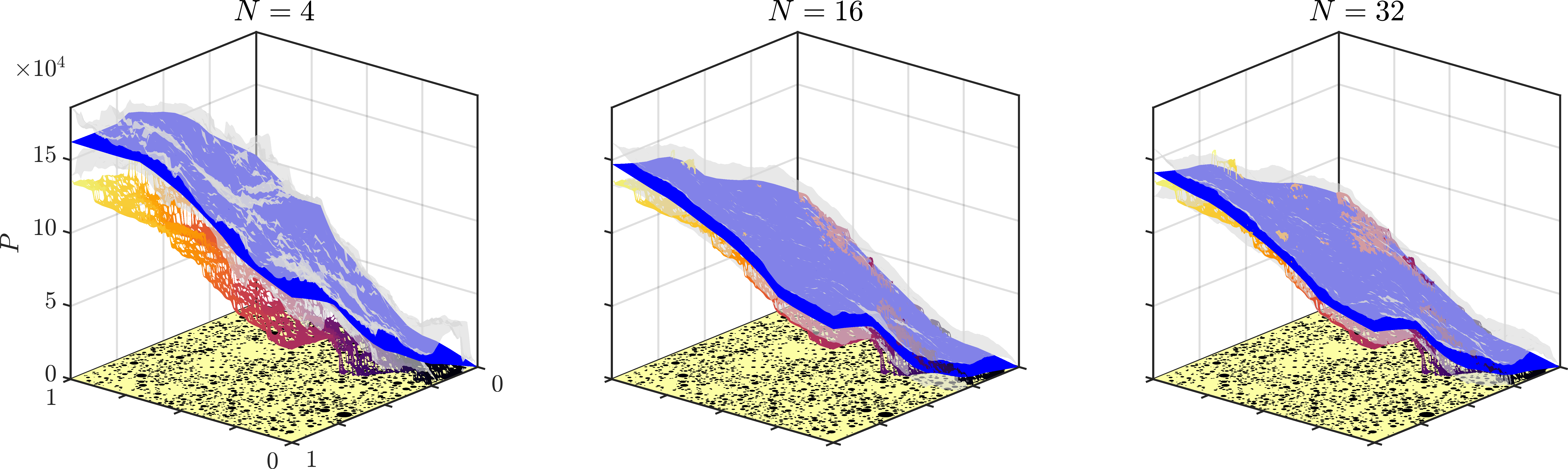}
\caption{Predictive examples for $N = 4, 16, 32$ training data samples with $\mu_{\tx{ex}} = 7.8$, $\sigma_{\tx{ex}} = 0.2$, $\sigma_r = 0.3$, $l_s = 1.2$, $l_{\bs x} = 0.08$ and $l_r = 0.05$, and a $\tx{dim}(\bs \lambda_c) = 4\times 4$ Darcy CGM. The colored surface is the true response $P(\bs x)$ of the test sample and the blue is the predictive mean $\bs \mu_{\tx{pred}}$ of the surrogate. The transparent gray surfaces give the predictive standard deviations $\bs \mu_{\tx{pred}}\pm \bs \sigma_{\tx{pred}}$. It can nicely be observed how predictions improve with increasing number of training data $N$.}
\label{fig:predVsData}
\end{figure}

Figure \ref{fig:errorPlot} depicts  both error metrics $R^2$ and $MLL$ as a function of the number of training samples $N$ for the different $\tx{dim}(\bs \lambda_c)$  as described above. 
One observes that with only $N \approx 32$ training samples (i.e. FGM simulations), the asymptotic values for $R^2$ and $MLL$ are attained. Furthermore, it is interesting to note that the highest $\tx{dim}(\bs \lambda_c) = 64$ achieves the second best $R^2$ score after $\tx{dim}(\bs \lambda_c) = 16$. This could be attributed to the higher information loss taking place using the local feature functions $\varphi_{\tx{loc}}$ on a finer grid in \refeq{eq:p_cExperiments}.

Figure \ref{fig:predVsData} shows three predictive means $\bs \mu_{\tx{pred}}$ (blue) $\pm$ one predictive standard deviation $\bs \sigma_{\tx{pred}}$ (transparent gray) on the same microstructure $\bs \lambda_f$, but using different number of training data $N = 4, 16, 32$, computed with $\tx{dim}(\bs \lambda_c) = 16$. One can observe how the true solution (colored surface) is better captured the more data $N$ is used for training.

\subsubsection{Predictive variance \texorpdfstring{$\bs \sigma^2_{\tx{pred}}$}{} \& \texorpdfstring{$L^2$}{}-error}
\begin{figure}[t]
\centering
\includegraphics[width=.8\textwidth]{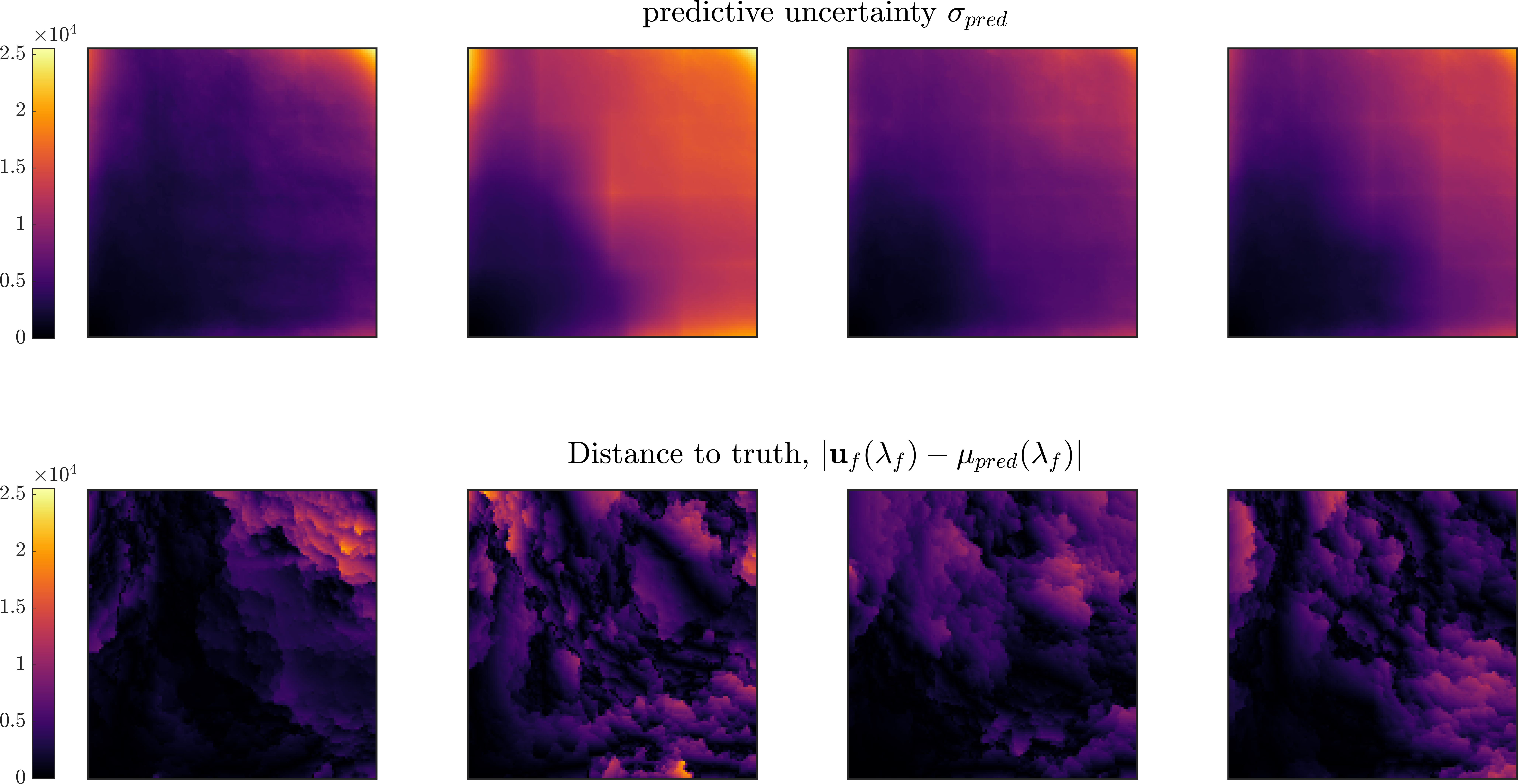}
\caption{Predictive uncertainty $\sigma_{\tx{pred}}$ (top row) and corresponding \revone{absolute} $L^2$-error $|\bs u_f(\bs \lambda_f) - \bs \mu_{\tx{pred}}(\bs \lambda_f)|$ for microstructural data drawn as in section \ref{sec:predPerformance}, but with boundary conditions according to $a_x = a_y = 0, a_{xy} = -1$. It can be observed that the predictive uncertainty $\bs \sigma_{\tx{pred}}$ is high/low in regions where the true \revone{absolute} $L^2$-error is high/low.}
\label{fig:predErr_L2err}
\end{figure}
In Figure \ref{fig:predErr_L2err}, the predictive uncertainty $\bs \sigma_{\tx{pred}}$ (top row) and the corresponding `true' $L^2$-error $|\bs u_f(\bs \lambda_f) - \bs \mu_{\tx{pred}}(\bs \lambda_f)|$ are plotted for four different microstructures $\bs \lambda_f$ drawn according to the same distribution as in section \ref{sec:predPerformance}, using a $4\times 4$ permeability field discretization and $N = 128$ training samples. To provoke different predictive errors in different regions in the domain, we use non-homogeneous flux boundary conditions of type \eqref{eq:bc} with $a_x = a_y = 0, a_{xy} = -1$. It can be observed that $\bs \sigma_{\tx{pred}}$ is indeed higher in samples/regions where the true $L^2$ error $|\bs u_f(\bs \lambda_f) - \bs \mu_{\tx{pred}}(\bs \lambda_f)|$ is higher, whereby it is incapable to resolve errors on a smaller scale than the one defined by the effective permeability field discretization.

\subsubsection{Effective Darcy permeabilities}
\begin{figure}[t]
\centering
\includegraphics[width=\textwidth]{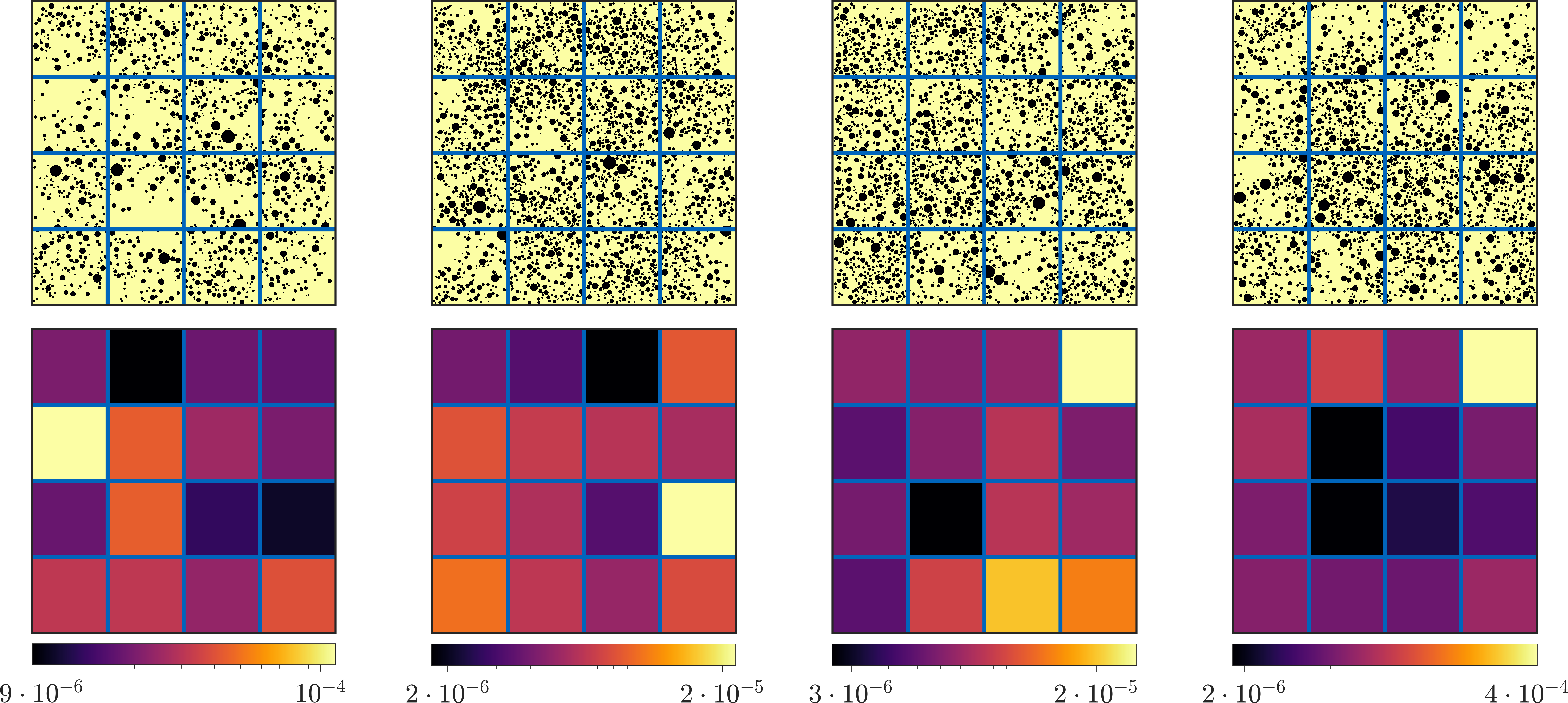}
\caption{Mean effective permeability field $\left<\bs K(\bs x, \bs \lambda_c) \right>_{Q_{\bs \lambda_c}}$ (bottom) and corresponding microstructures (top). The data is chosen randomly from an $N = 128$ training set as was used in section \ref{sec:predPerformance}.}
\label{fig:effPermeability}
\end{figure}
\begin{figure}[t]
\centering
\includegraphics[width=.9\textwidth]{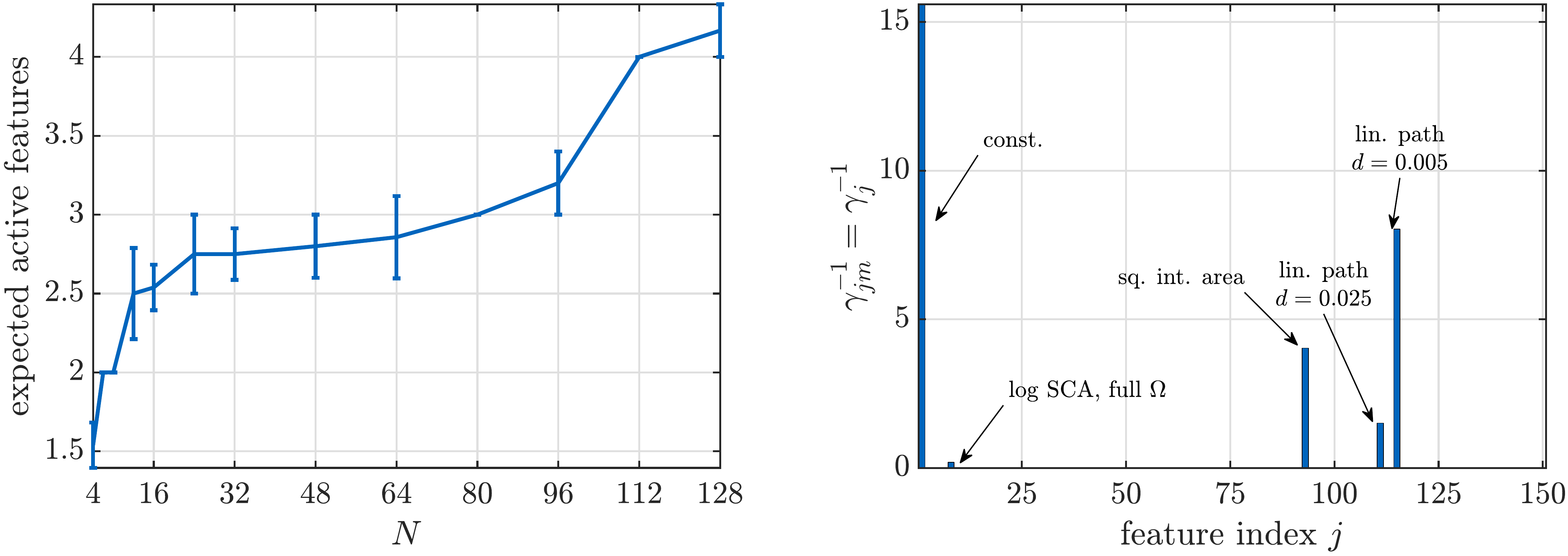}
\caption{Expected number of active features as a function of the number of training samples $N$ (left). We see that the prior model described in section \ref{sec:prior} effectively prunes most of the provided feature functions $\varphi$. The right side of the picture shows a bar plot of the converged values of the inverse precisions $<\gamma_{i}>^{-1}$ corresponding to feature $\varphi_i$. These were found using a $4\times 4$ effective permeability field discretization and 128 training samples with microstructures and boundary conditions as used in section \ref{sec:predPerformance}. Activated are: the constant feature/bias (i.e. $\varphi_i(\bs \lambda_f) = 1$), the $\log$ self-consistent approximation \cite{Torquato2001} (infinite contrast limit), the squared area of total solid/fluid interfaces, and fluid-phase lineal-path function evaluated at distances $d = 0.025$ and $d = 0.005$.}
\label{fig:activeFeatures}
\end{figure}

Further insight on the numerical homogenization process taking place in the $\bs \lambda_f \mapsto \bs \lambda_c$ encoding process is provided by Figure \ref{fig:effPermeability}, which shows the mean effective permeability field $\left<\bs K(\bs x, \bs \lambda_c) \right>_{Q_{\bs \lambda_c}}$ (bottom row, see section \ref{sec:CGM}) and the corresponding microstructures (top) for four randomly chosen training samples of a $4\times 4$ effective model trained on $N = 128$ samples using boundary conditions and microstructure distributions as in \ref{sec:predPerformance}. It is  clear that effective permeabilities are lower in cells with higher solid fraction/interface area.

Figure \ref{fig:activeFeatures} \revtwo{provides information on the activated feature functions. These are determined by the  posterior mean of the  precision parameters $\gamma_{jm} = \gamma_j$ that do not converge to large values (according to \refeq{eq:Qtheta} when $\left<\gamma_{jm}\right> \rightarrow \infty$ then $\sigma_{\tilde{\theta}_{c, jm}}^2 \rightarrow 0 $). We note that, apart from the bias term, the features activated pertain to the self-consistent approximation \cite{Torquato2001} (infinite contrast limit), the squared area of total solid/fluid interfaces, and void-phase lineal-path function which pertains to the connectivity of the void phase.}
%
In  Figure \ref{fig:activeFeatures}, $\tx{dim}(\bs \lambda_c) = 16$ is the dimension of the CGM's constitutive  parameter vector and a set of $N_{\tx{features}} = \tx{dim}(\bs \gamma) = 150$ feature functions $\varphi(\bs \lambda_f)$ specified in \ref{sec:features} is used. 
The left side of the Figure \ref{fig:activeFeatures} shows the average number of activated features (i.e. with nonzero $\left<\gamma_j\right>^{-1}$) as  a function of the number of training data $N$, where error bars are due to randomization of training sets. As expected, the number of activated feature functions increases with the number of training data. 
The right part is an example of the final, converged values of $\left<\gamma_j\right>^{-1}$ for $N = 128$ training samples, where 5 features are activated (explained in the Figure caption).

\subsection{Predictions under different boundary conditions}
\label{sec:diffBC}
\begin{figure}[h]
\centering
\includegraphics[width=.8\textwidth]{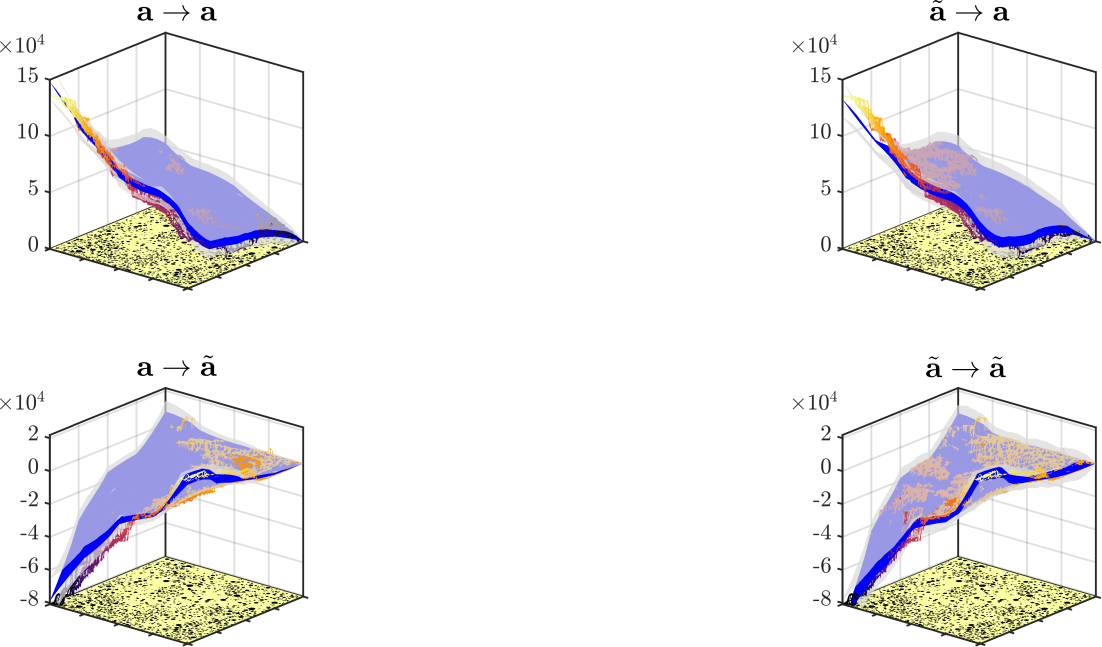}
\caption{Cross-prediction example on boundary conditions $\bs a = \{a_x = 1, a_y = 1, a_{xy} = 0 \}$ and $\tilde{\bs a} = \{\tilde{a}_x = 0, \tilde{a}_y = 0, \tilde{a}_{xy} = -1\}$ using a $4\times 4$ effective permeability field and $N = 128$ training samples. The colored surface is the true solution of a test sample, the blue is the predictive mean $\bs \mu_{\tx{pred}}$ of the surrogate and the transparent gray surfaces indicate the predictive error $\bs \mu_{\tx{pred}} \pm \bs \sigma_{\tx{pred}}$. The figures on the diagonal have identical boundary conditions $\bs a, \tilde{\bs a}$ on both training and test data. The top right example is trained on data with boundary conditions $\tilde{\bs a}$, but tested on data with boundary conditions $\bs a$, whereas the lower left is trained on $\bs a$, but tested on $\tilde{\bs a}$.}
\label{fig:cross_pred}
\end{figure}

\begin{table}[t]
\begin{small}
\begin{center}
\tbf{Coefficient of determination} $R^2$
\end{center}
\begin{center}
\begin{tabular}{r||c|c}
\begin{tabular}{c} trained on \\ \hline prediction on \end{tabular}
&$\bs a = \bsmat a_x = 1 & a_y = 1& a_{xy} = 0\esmat^T$ &$\tilde{\bs a} = \bsmat \tilde{a}_x = 0 & \tilde{a}_y = 0 & \tilde{a}_{xy} = -1\esmat^T$ \\\hline\hline
$\bs a = \bsmat a_x = 1 & a_y = 1& a_{xy} = 0\esmat^T$ & $0.930 \pm 1.3\cdot 10^{-3}$ & $0.863\pm 1.1\cdot 10^{-2}$  \\\hline
$\tilde{\bs a} = \bsmat \tilde{a}_x = 0 & \tilde{a}_y = 0 & \tilde{a}_{xy} = -1\esmat^T$ & $0.887\pm 7.5\cdot 10^{-3}$ & $0.903\pm 7.0\cdot 10^{-3}$ 
\end{tabular}
\vspace{2mm}
\end{center}
\begin{center}
\tbf{Mean log likelihood} $MLL$
\end{center}
\begin{center}
\begin{tabular}{r||c|c}
\begin{tabular}{c} trained on \\ \hline prediction on \end{tabular}
&$\bs a = \bsmat a_x = 1 & a_y = 1& a_{xy} = 0\esmat^T$ &$\tilde{\bs a} = \bsmat \tilde{a}_x = 0 & \tilde{a}_y = 0 & \tilde{a}_{xy} = -1\esmat^T$ \\\hline\hline
$\bs a = \bsmat a_x = 1 & a_y = 1& a_{xy} = 0\esmat^T$ & $-11.1\pm 7.6\cdot 10^{-3}$ & $-11.8\pm 9.3\cdot 10^{-3}$  \\\hline
$\tilde{\bs a} = \bsmat \tilde{a}_x = 0 & \tilde{a}_y = 0 & \tilde{a}_{xy} = -1\esmat^T$ & $-10.2\pm 1.3\cdot 10^{-2}$ & $-9.93\pm 2.3\cdot 10^{-2}$ 
\end{tabular}
\end{center}
\caption{Error metrics $R^2$ and $MLL$ corresponding to the cross-prediction example in section \ref{sec:diffBC} using distinct boundary conditions $\bs a$ and $\tilde{\bs a}$. Metrics are evaluated and averaged over 3 separate training sets of $N = 128$ size (indicated errors are due to variation of training data) and tested on $N_{\tx{test}} = 1024$ microstructural samples. Only slight deterioration is observed if boundary conditions of the training and test sets differ.}
\label{tab:crossPredTable}
\end{small}
\end{table}

A particularly useful quality of the proposed coarse-grained model is that boundary conditions can be imprinted deterministically onto the Darcy-based CGM as specified in equations \eqref{eq:Darcy} (c-d). As a result, a model trained on boundary conditions, say $\bs a$ (see section \ref{sec:data}), can be used  for predictions  on different boundary conditions, say $\tilde{\bs a} \neq \bs a$. Good predictive performance in such cases would suggest the ability of the proposed coarse-grained model to {\em extrapolate} which presupposes that it has correctly encoded salient physical information. 
This is a desirable trait in all  surrogates which cannot always be achieved with black-box statistical models that attempt to interpolate the input-output map \cite{Tripathy2018,Zhu2018}.

As an example, we use the boundary conditions $\bs a=\{a_x = a_y = 1, a_{xy} = 0\}$ (as in section \ref{sec:predPerformance}) and boundary conditions $\tilde{\bs a} =\{\tilde{a}_x = \tilde{a}_y = 0, \tilde{a}_{xy} = -1\}$. We generate $N = 128$ FGM input-output training pairs for each of the aforementioned boundary conditions and train a coarse-grained model with $\tx{dim}(\blc) = 16$ as previously described.  
In Table \ref{tab:crossPredTable}, we report the performance metrics $R^2$ and $MLL$ as evaluated on test data generated with each of the above boundary conditions. We observe that the off-diagonal entries which correspond to different BCs for training and testing are comparable with the diagonal entries which correspond to identical BCs for training and testing.
Indicative test samples and the corresponding predictions  are shown in Figure \ref{fig:cross_pred}. We observe again comparable predictive accuracy between the different combinations of training and test data.

\begin{figure}[t]
\centering
\includegraphics[width=\textwidth]{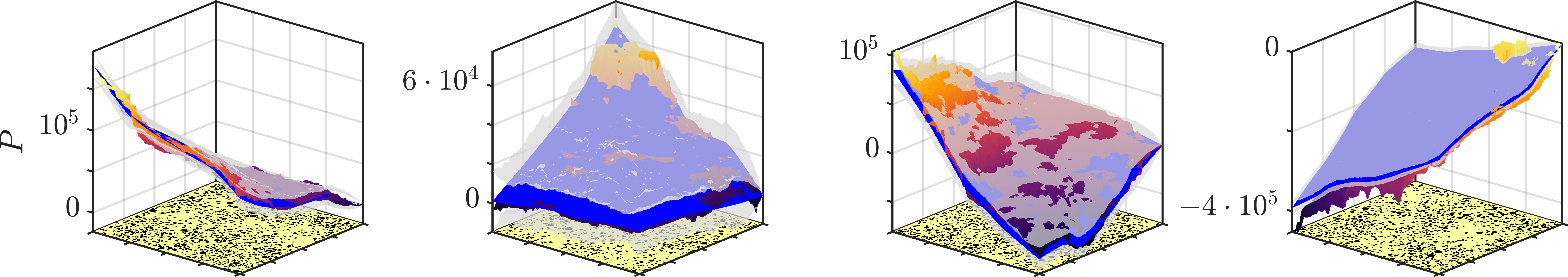}
\caption{Predictions on four randomly chosen test samples where boundary conditions in both training and test data are randomized according to $a_x \sim \mathcal N(0, 1)$, $a_y \sim \mathcal N(0, 1)$, $a_{xy} \sim \mathcal N(0, 1)$. The colored surface is the true Stokes pressure response of the test sample, the blue is the predictive mean $\bs \mu_{\tx{pred}}$ of the surrogate and the transparent gray surfaces show the predictive standard deviation $\bs \mu_{\tx{pred}} \pm \bs \sigma_{\tx{pred}}$.}
\label{fig:predRandBC}
\end{figure}

We also perform full randomization of boundary conditions in a further scenario, where different boundary conditions for every single test and training datum are drawn. In particular, we draw $a_x \sim \mathcal N(0, 1)$, $a_y \sim \mathcal N(0, 1)$, $a_{xy} \sim \mathcal N(0, 1)$ and generate $N = 128$ training data (i.e. FGM input-output pairs). Figure  \ref{fig:predRandBC} depicts predictions over four test cases with randomly selected BCs as above, where it is clearly seen that the surrogate is capable of producing accurate estimates both in terms of the mean as well as the breadth of the predictive uncertainty.
The error metrics $R^2$ and $MLL$ are evaluated on a $N_{\tx{test}} = 1024$ test set and yield the following values:
\be 
R^2 = 0.9776 \pm 0.0027 \quad  and \quad MLL = -11.07 \pm 0.089
\ee
which are comparable to the values provided earlier.

\subsection{Uncertainty propagation problem}
\begin{figure}[t]
\centering
\includegraphics[width=.7\textwidth]{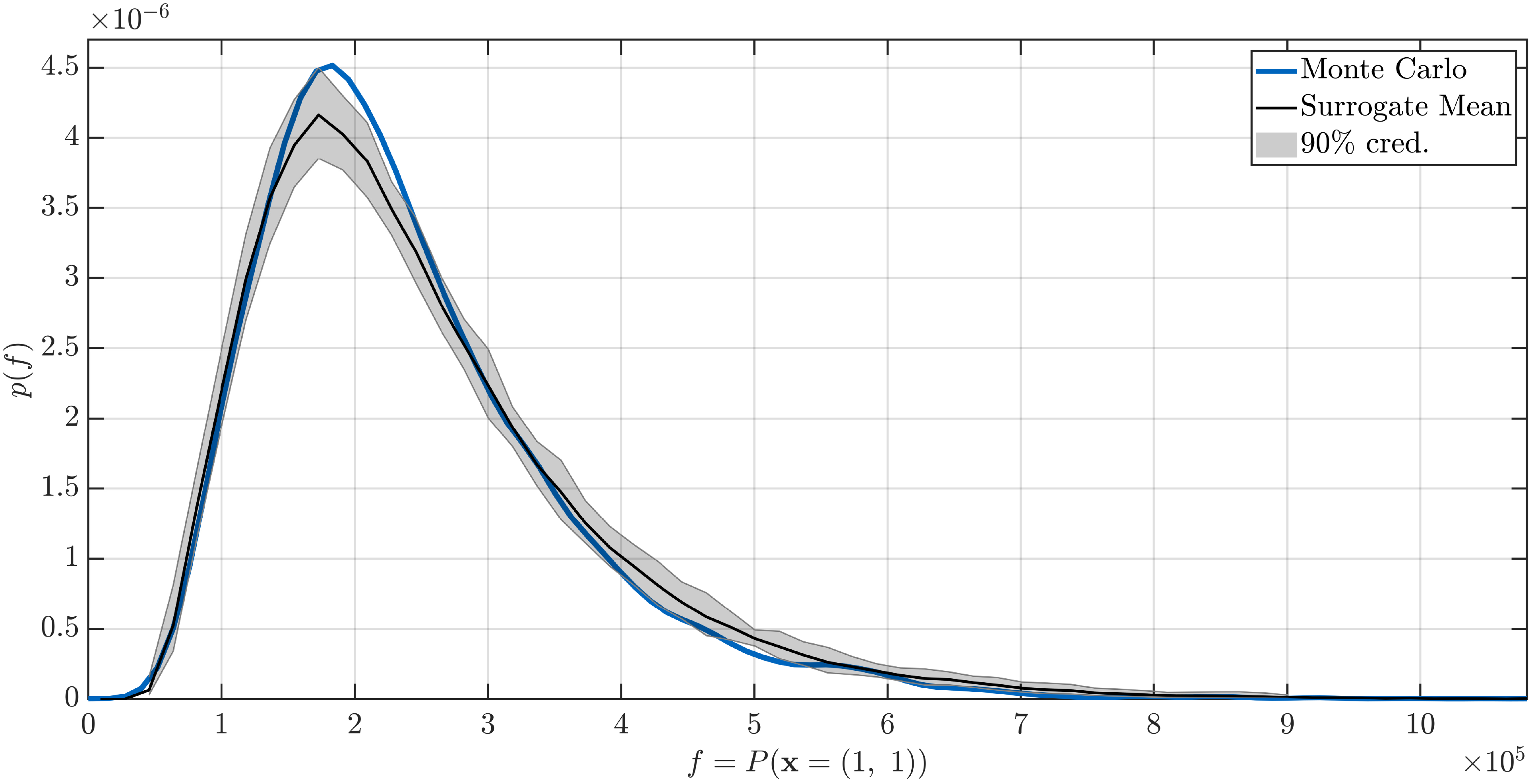}
\caption{Probability density function for the fine-scale Stokes pressure response at $\bs x = (1, ~~ 1)^T$ estimated using 10,000 Monte Carlo samples (blue line) and a surrogate model (black) trained on $N = 32$ training samples using an effective Darcy solver based on a $4\times 4$ permeability field discretization. Microstructures are sampled as in section \ref{sec:predPerformance}. The gray shaded area shows the $90\%$ credible intervals obtained by propagating the uncertainty of model parameters $\tilde{\bs \theta}_c$, $\bs \tau_c$ and $\bs \tau_{cf}$ as described in the text.}
\label{fig:UP32}
\end{figure}

The numerical experiment carried out in this section pertains to a simple uncertainty propagation (UP) problem. In a UP setting, one is interested in computing statistics of some quantity of interest (QoI) $f(\bs u_f)$ of the FGM output $\buf$  given some  density  $p(\bs \lambda_f)$ for the FGM input $\blf$. 
For the sake of illustration we assume that the QoI is scalar and therefore its density can be written as
\begin{equation}
p(f) = \int \delta(f - f(\bs u_f)) p(\bs u_f|\bs \lambda_f) p(\bs \lambda_f) d\bs \lambda_f d\bs u_f,
\end{equation}
where $p(\bs u_f|\bs \lambda_f) = \delta(\bs u_f - \bs u_f(\bs \lambda_f))$ if $\bs u_f$ is computed by running the  expensive FGM $\bs u_f(\bs \lambda_f)$. One can approximate  $p(f)$ with a histogram where the values in each bin are estimated with  Monte Carlo (MC) sampling. Due to slow MC convergence rates and the high cost of the FGM forward solver, an obvious strategy is to replace the FGM by the probabilistic coarse-grained model proposed and more specifically the predictive posterior  $p_{\tx{pred}}(\bs u_f|\bs \lambda_f, \mathcal D)$  (section \ref{sec:predictions}) as follows:
\begin{equation}
p(f |\mathcal D) = \int \delta(f - f(\bs u_f)) p_{\tx{pred}}(\bs u_f|\bs \lambda_f, \mathcal D) p(\bs \lambda_f) d\bs \lambda_f d\bs u_f.
\end{equation}
To illustrate the performance of such a scheme, we use as the QoI the FGM pressure $P(\bs x)$ at $\bs x = \bmat 1, & 1 \emat^T$ (i.e. top right corner of the problem domain $\Omega$). 
The blue line in Figure \ref{fig:UP32} shows a kernel density estimate of the MC-based histogram of 10,000 FGM runs. 
The black line is based on a trained coarse-grained model ($\tx{dim}(\blc) = 16$) with $N = 32$ training samples. 
The gray shaded area of Figure \ref{fig:UP32} aims to capture predictive uncertainties about $p(f)$ due to limited data. To that end, instead of $p_{\tx{pred}}(\bs u_f|\bs \lambda_f, \mathcal D)$ as defined in \refeq{eq:p_pred0}, we generate samples of the model parameters $\tilde{\bs \theta}_c^{(s)}, \bs \tau_c^{(s)}$ and $\bs \tau_{cf}^{(s)}$ from their approximate posteriors $Q_{\tilde{\bs \theta}_c}, Q_{\bs \tau_c}, Q_{\bs \tau_{cf}}$ and for each of those values we generate samples from
\begin{equation}
\begin{gathered}
    p(f|\tilde{\bs \theta}_c^{(s)}, \bs \tau_c^{(s)}, \bs \tau_{cf}^{(s)}, \mathcal D) = \\
    \int \delta(f - f(\bs u_f)) p_{\tx{pred}}(\bs u_f|\bs \lambda_f, \tilde{\bs \theta}_c^{(s)}, \bs \tau_c^{(s)}, \bs \tau_{cf}^{(s)}, \mathcal D) p(\bs \lambda_f) d\bs \lambda_f d\bs u_f.
\end{gathered}
\end{equation}
The mean and  $90\%$ credible intervals for the histogram value in each  bin are  subsequently  computed and plotted in Figure \ref{fig:UP32}. One can observe that the credible intervals envelop correctly even the tails of the true, MC-based histogram.

\subsection{Illustration of automatic adaptive refinement}
\label{sec:expref}
\begin{figure}[t]
\centering
\includegraphics[width=.9\textwidth]{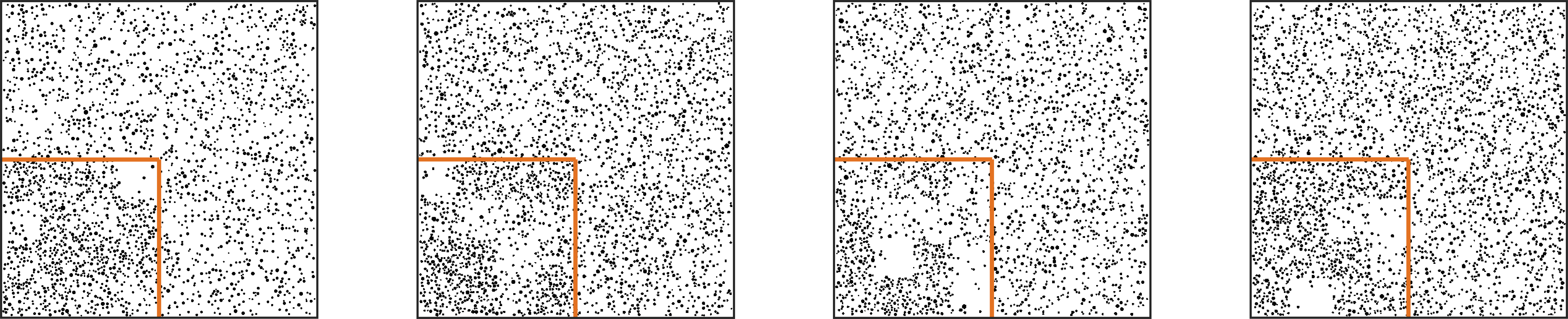}
\caption{`Tiled' microstructures, where spherical exclusions (solid phase) are distributed homogeneously except for the lower left corner $\Omega_{ll} = [0, 0.5] \times [0, 0.5]$ of the domain, where exclusions are homogeneously distributed (with differing random volume fraction) on $0.125 \times 0.125$ sub-cells. Such microstructures are used to enforce adaptive refinement in the lower left corner $\Omega_{ll}$ of the domain.}
\label{fig:tiledMicrostruct}
\end{figure}

\begin{figure}[t]
\centering
\includegraphics[width=.9\textwidth]{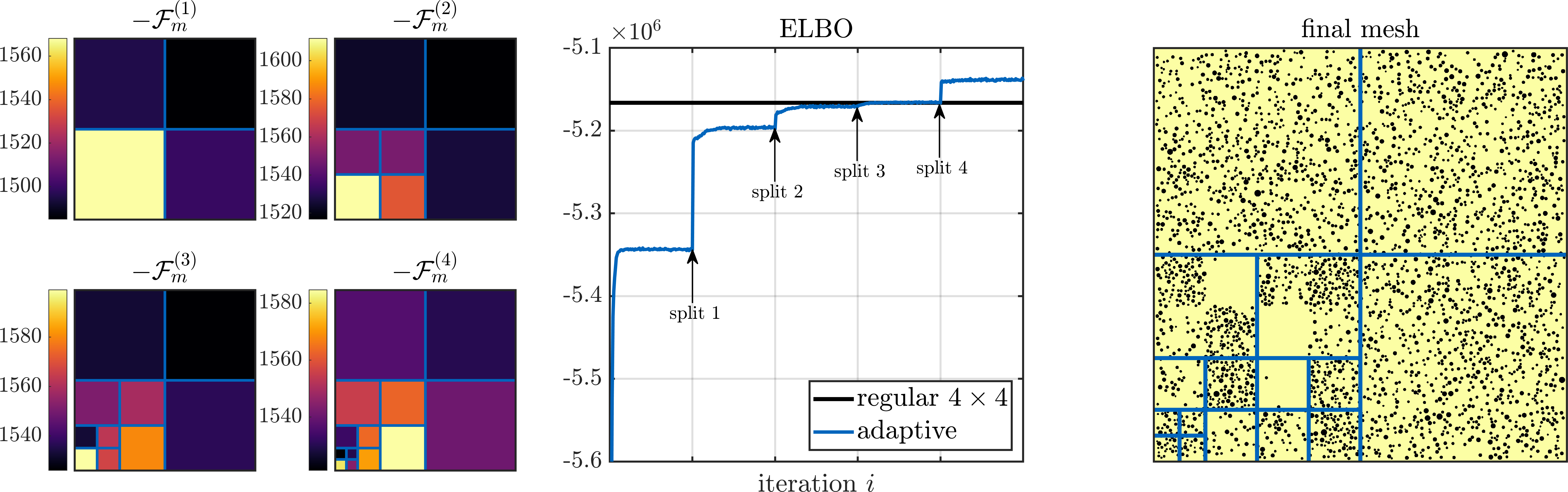}
\caption{Left: 
Cell scoring function $-\mathcal F_m^{(i)}$ given in equation \eqref{eq:cellScore} directly before every split (lighter color indicates smaller $\mathcal F_m^{(i)}$). Middle: 
evidence lower bound (ELBO) as given by equations \eqref{eq:elbo_exp}, \eqref{eq:elboComp}. The blue line shows the ELBO vs. training iteration of the adaptively refined model, the black line shows the final ELBO of a model with effective permeability discretized on a $4\times 4$ square grid. The right part of the Figure shows the final effective permeability field discretization (blue grid) with a representative microstructure underneath. }
\label{fig:split_plot}
\end{figure}

In this section, we illustrate the automatic adaptive refinement scheme of the coarse-grained model as explained in section \ref{sec:aar}.
We consider as the base model $\tx{dim}(\blc) = 4$ which corresponds to a regular  $2\times 2$ square grid discretization of the Darcy permeability field (see top left of Figure \ref{fig:split_plot}).
In order to better illustrate the capabilities of the proposed method, we draw the FGM microstructural samples, i.e. $\blf$, from a particular distribution and refer to the corresponding samples as `tiled' microstructures. These exhibit a homogeneous distribution of circular exclusions (solid phase) over the whole domain except for the lower left quadrant, i.e. $\Omega_{ll} = [0, 0.5] \times [0, 0.5]$. In this region, exclusions are distributed homogeneously on sub-cells of size $0.125 \times 0.125$, but with varying random volume fraction as can be seen in Figure \ref{fig:tiledMicrostruct} where four representative microstructures are depicted. Due to this localized inhomogeneity, it is natural to expect that refinements in  $[0, 0.5] \times [0, 0.5]$ would be most promising. The baseline $\tx{dim}(\bs \lambda_c) = 2\times 2$ model is trained on $N = 32$ data generated by the Stokes-based FGM.

After training, the cell scoring function defined in equation \eqref{eq:cellScore} indicates which of the $2\times 2 = 4$ cells of constant permeability should be split into four new  square sub-cells. 
After splitting that cell, training continues with all model parameters initialized to their previously found values. To facilitate the problem as much as possible, boundary conditions of type $a_x = 1$, $a_y = a_{xy} = 0$ are used, i.e. uniform expected velocity $\left<\bs V(\bs x)\right> = \left( 1, ~ 0\right)^T$ all over the domain. Given these boundary conditions and the `tiled' microstructures as described above, it should be beneficial to have a finer CGM permeability field discretization in $\Omega_{ll}$ than in the rest of the domain, where exclusions are distributed homogeneously.

Figure \ref{fig:split_plot} depicts the process of 4 successive splits each of which corresponds to dividing the selected cell into four equal-sized squares. Hence the number of cells/subregions and therefore the dimension of the associated $\tx{dim}(\blc)$ will become $7, 10, 13$ and $16$ after each of the aforementioned splits. 
The left part of Figure \ref{fig:split_plot} shows the cell scores $\mathcal F_m(Q)$ computed as in \refeq{eq:cellScore} right before each split. The cell that is subsequently subdivided is the one with the lowest $\mathcal F_m(Q)$. As expected, the splits are concentrated in the lower-left quadrant $\Omega_{ll}$ and the right part of the Figure shows the corresponding discretization of the permeability field after the 4 splits.
The tendency to refine close to the origin $\bs x = \bs 0$ (lower left corner) and along the lower boundary $y = 0$ may be explained by the fact that the microstructures are constructed such that solid phase exclusions have a minimum distance of $d = 0.003$ to the domain boundary, i.e. there always exists a narrow path along the boundaries where fluid can flow. As boundary conditions imply flux from left to right, it is beneficial to refine close to the lower domain boundary. The lower left corner is further distinguished by the fact that the pressure boundary condition $P(\bs x = 0) = 0$ is implied here.

More importantly, the middle part of \ref{fig:split_plot} shows the evolution of the ELBO over training iterations (blue line). The jumps that are observed after each split are a result of the improvement in the model which is to be expected due to the increased dimension $\tx{dim}(\blc)$.  We also compare the ELBO score after 4 splits (where $\dim(\blc)=16$) with a regular discretization of the Darcy permeability into $4 \times 4$ squares in which  $\tx{dim}(\blc) = 16$ as well, i.e. the dimension of the information bottleneck variables is the same. We note that the adaptive refinement proposed leads to an improved final model.

\section{Conclusions}
\label{sec:conclusion}
We have presented a physics-informed, fully Bayesian surrogate model to predict response fields of stochastic partial differential equations with high-dimensional stochastic input as those arising in random heterogeneous media. Employing an encoder-decoder architecture with a coarse-grained model (CGM) based on simplified physics as a stencil at its core, it is possible to provide tight and accurate probabilistic predictions even when the training data i.e. fine-grained model simulations, is scarce (i.e. of the order of $N \approx 10 \revtwomath{\sim} 100$) and the stochastic input dimension is large (i.e. $\sim10^4$ or larger). The encoding or coarse-graining step is capable of  extracting the salient features of the input random field which are most predictive of the fine-scale PDE-response, thereby drastically reducing the stochastic input dimension in an optimal way. This low-dimensional, latent representation then serves as the input to the CGM, the output of which is basal for the probabilistic reconstruction of the fine scale output.  Predictive uncertainty due to limited training data as well as due to the information loss unavoidably taking place during the coarse-graining step is accurately quantified by the fully probabilistic nature of the model. All parts of the model were trained simultaneously in a fully Bayesian way using Stochastic Variational Inference, without the need of any parameters to be specified by the analyst.

On the basis of the Stokes/Darcy similarity for flow through random porous media, we provided numerical evidence of the predictive capabilities under the aforementioned setting. Another important advantage of the proposed  formulation  is its ability to perform comparably well under {\em extrapolative} conditions. In the context of the models explored,  we demonstrated this by carrying out predictions on data under a considerably different (or even randomized) boundary conditions than those  used for training. Furthermore, we demonstrated one of the many  possible application of the proposed model for Uncertainty Propagation problems where not only accurate estimates of output statistics but the confidence in these predictions is quantified. Finally, we outlined a method for  automatically and adaptively refining the model and demonstrated its benefits in the context of the problems explored. 

Applications of the proposed model can be found in any multi-query setting with expensive forward model evaluations, as is often the case in design-, optimization- or inverse problems. Several model extensions can be contemplated. For coarse-graining of the high-dimensional stochastic inputs, a deep learning framework could be adopted, potentially benefiting from recent advances in computer vision with e.g. convolutional neural networks (CNNs) or deep Gaussian process models. We emphasize however that significant regularization would be needed to make those work in the {\em Small Data} setting considered. Furthermore, one might think of a model that not only passes information through the CGM, but also entails `bypassing' components, thereby effectively expanding the information bottleneck and allowing for more detailed reconstruction of fine scale output variability.

\appendix
\section{Optimizing the variational distributions \texorpdfstring{$Q$}{}}
\label{ap:optQ}
To maximize the ELBO functional $\mathcal F(Q)$ in equation \eqref{eq:elbo} w.r.t. the variational distributions $Q$ under the normalization constraint $\int Q_{\bs \theta_k}(\bs \theta_k)d\bs\theta_k = 1$, we maximize the functional $\mathcal J[Q; \bs \xi]$
\begin{equation}
\begin{split}
     \mathcal J[Q; \bs \xi] &= \int \prod_k\left(Q_{\bs \theta_k}(\bs\theta_k)\right) \log \frac{p(\bs \theta, \mathcal D)}{\prod_l\left(Q_i{\bs \theta_l}(\bs\theta_l) \right)}d\bs \theta \\
     &+ \sum_k \xi_k \left(\int Q_{\bs \theta_k}(\bs \theta_k)d\bs\theta_k - 1 \right) \\
     &= \int \prod_k\left(Q_{\bs \theta_k}(\bs\theta_k)\right) \log p(\bs \theta, \mathcal D)d\bs \theta - \sum_k \int Q_{\bs \theta_k}(\bs\theta_k) \log Q(\bs\theta_k)d\bs\theta_k \\
     &+ \sum_k \xi_k \left(\int Q_{\bs \theta_k}(\bs \theta_k)d\bs\theta_k - 1 \right)
\end{split}
\end{equation}
where $\bs \theta_k \subset \bs \theta = \left\{\left\{\bs \lambda_c^{(n)} \right\}_{n = 1}^N, \tilde{\bs \theta}_c, \bs \tau_c, \bs \tau_{cf}, \bs \gamma \right\}$ is a subset of the model parameters $\bs \theta$ and $\bs\xi$ are Lagrangean multipliers. The first order variation $\delta \mathcal J$ w.r.t. $Q_{\bs \theta_k}$ is
\begin{equation}
\begin{split}
\delta\mathcal J(\bs \theta_k; \xi_j) = \int \delta Q_{\bs \theta_k}(\bs \theta_k) \Bigg(\int &\prod_{i\neq k}\left(Q_{\bs \theta_i}(\bs \theta_{i}) \right) \log p(\bs \theta, \mathcal D)d\bs \theta_{-k} \\
&- \log Q_{\bs \theta_k}(\bs \theta_k) - 1 + \xi_k \Bigg)d\bs \theta_k,
\end{split}
\end{equation}
where $d\bs \theta_{-k} = \prod_{i \neq k} d\bs\theta_i$. As $\delta \mathcal J \stackrel{!}{=} 0$ for arbitrary $\delta Q_{\bs \theta_k}$,
\begin{equation}
\int \prod_{i\neq k}\left(Q_{\bs \theta_i}(\bs \theta_i) \right) \log p(\bs \theta, \mathcal D)d\bs \theta_{-k} - \log Q_{\bs \theta_k}(\bs \theta_k) - 1 + \xi_k = 0.
\end{equation}
or
\begin{align}
\log Q_{\bs \theta_k}(\bs \theta_k) &= \left<\log p(\bs \theta, \mathcal D) \right>_{i\neq k} + \xi_k - 1,
\label{eq:QVI}
\end{align}
where $\left<~.~\right>_{i\neq k}$ denotes the expected value w.r.t. $\prod_{i\neq k}\left(Q_{\bs \theta_i}(\bs \theta_{i}) \right)$. The normalization constraints are recaptured by $\nabla_{\xi} \mathcal J \stackrel{!}{=} 0$ and determine the constant $\xi_k - 1$ in \refeq{eq:QVI}. Thus, it is found that
\begin{equation}
Q_{\bs \theta_k}(\bs \theta_k) = \frac{\exp \left<\log p(\bs \theta, \mathcal D) \right>_{i\neq k}}{\int \exp \left<\log p(\bs \theta, \mathcal D) \right>_{i\neq k} d\theta_i}.
\end{equation}

\subsection{Black-box variational inference for \texorpdfstring{$Q_{\bs \lambda_c^{(n)}}$}{}}
\label{ap:bbvi}
We note that minimization of the KL divergence as given in equation \eqref{eq:optqlambda} is equivalent to
\begin{equation}
\begin{split}
\arg \min_{\bs \mu_{\bs \lambda_c}^{(n)}, \bs \sigma_{\bs \lambda_c}^{(n)}} & \tx{KL}\left(\left.\tilde{Q}_{\bs \lambda_c^{(n)}}(\bs \lambda_c^{(n)}|\bs \mu_{\bs \lambda_c}^{(n)}, \bs \sigma_{\bs \lambda_c}^{(n)}) \right|\left| Q_{\bs \lambda_c^{(n)}}(\bs \lambda_c^{(n)})\right.\right) \\
= \arg\min_{\bs \mu_{\bs \lambda_c}^{(n)}, \bs \sigma_{\bs \lambda_c}^{(n)}} & \int \tilde{Q}_{\bs \lambda_c^{(n)}}(\bs \lambda_c^{(n)}| \bs \mu_{\bs \lambda_c}^{(n)}, \bs \sigma_{\bs \lambda_c}^{(n)}) \log \frac{\tilde{Q}_{\bs \lambda_c^{(n)}}(\bs \lambda_c^{(n)}| \bs \mu_{\bs \lambda_c}^{(n)}, \bs \sigma_{\bs \lambda_c}^{(n)})}{Q_{\bs \lambda_c^{(n)}}(\bs \lambda_c^{(n)})} d\bs \lambda_c^{(n)} \\
= \arg\max_{\bs \mu_{\bs \lambda_c}^{(n)}, \bs \sigma_{\bs \lambda_c}^{(n)}} & \left( \vphantom{\sum_{m = 1}^{\tx{dim}(\bs \lambda_c)}} \right. \left<\log \mathcal N(\bs u_f^{(n)}| \bs W \bs u_c(\bs \lambda_c^{(n)}), \tx{diag}[\bs \tau_{cf}]^{-1}) \right>_{\tilde{Q}_{\bs \lambda_c^{(n)}}} \\
& + \sum_{m = 1}^{\tx{dim}(\bs \lambda_c)} \left<\log \mathcal N(\lambda_{c, m}^{(n)}|\tilde{\bs \theta}_{c, m}^T \bs\varphi_m(\bs \lambda_f^{(n)}), \tau_{c, m}^{-1}) \right>_{\tilde{Q}_{\bs \lambda_c^{(n)}}} \\
& \left. + H\left(\tilde{Q}_{\bs \lambda_c^{(n)}}(\bs \lambda_c^{(n)})\right) \vphantom{\sum_{m = 1}^{\tx{dim}(\bs \lambda_c)}} \right),
\end{split}
\label{eq:VI}
\end{equation}
where $H\left(\tilde{Q}_{\bs \lambda_c^{(n)}}(\bs \lambda_c^{(n)})\right)$ is the Shannon entropy of $\tilde{Q}_{\bs \lambda_c^{(n)}}$. Expected values of terms in \eqref{eq:VI} depending on the CGM $\bs u_c(\bs \lambda_c^{(n)})$ are analytically intractable and gradients w.r.t. $\bs \mu_{\bs \lambda_c}^{(n)}, \bs \sigma_{\bs \lambda_c}^{(n)}$ are therefore only accesible by noisy Monte Carlo estimates. To reduce the variance in those estimates, we apply the reparametrization trick \cite{Kingma2013}
\begin{equation}
\lambda_{c, m}^{(n)} = \mu_{\bs \lambda_c, m}^{(n)} + \sigma_{\bs \lambda_c, m}^{(n)} \epsilon_m^{(n)}, \qquad \epsilon_m^{(n)} \sim \mathcal N(0, 1)
\end{equation}
and express \refeq{eq:VI} as
\begin{equation}
\begin{gathered}
\bs \mu_{\bs \lambda_c}^{(n)}, \bs \sigma_{\bs \lambda_c}^{(n)} = \\
 \arg\max_{\bs \mu_{\bs \lambda_c}^{(n)}, \bs \sigma_{\bs \lambda_c}^{(n)}} \left( \vphantom{\sum_{m = 1}^{\tx{dim}(\bs \lambda_c)}} \right. \left<\log \mathcal N(\bs u_f^{(n)}| \bs W \bs u_c(\bs \mu_{\bs \lambda_c}^{(n)} + \bs \sigma_{\bs \lambda_c}^{(n)} \circ \bs \epsilon^{(n)}), \tx{diag}[\bs \tau_{cf}]^{-1}) \right>_{\mathcal N(\bs \epsilon^{(n)}| \bs 0, \bs I)}  \\
 + \sum_{m = 1}^{\tx{dim}(\bs \lambda_c)} \left<\log \mathcal N(\mu_{\bs \lambda_c, m}^{(n)} + \sigma_{\bs \lambda_c, m}^{(n)} \epsilon_m^{(n)}|\tilde{\bs \theta}_{c, m}^T \bs\varphi_m(\bs \lambda_f^{(n)}), \tau_{c, m}^{-1}) \right>_{\mathcal N(\bs \epsilon^{(n)}| \bs 0, \bs I)} \vphantom{\sum_{m = 1}^{\tx{dim}(\bs \lambda_c)}} \\
 + \left. \sum_{m = 1}^{\tx{dim}(\bs \lambda_c)} \log \sigma_{\bs \lambda_c, m}^{(n)}\right),
\label{eq:VIopt}
\end{gathered}
\end{equation}
where `$\circ$' denotes element-wise multiplication and we used the fact that \newline $H\left(\tilde{Q}_{\bs \lambda_c^{(n)}}(\bs \lambda_c^{(n)})\right) \vphantom{\sum_{m = 1}^{\tx{dim}(\bs \lambda_c)}} \propto \sum_{m = 1}^{\tx{dim}(\bs \lambda_c)} \log \sigma_{\bs \lambda_c, m}^{(n)}$. Derivatives w.r.t. $\bs \mu_{\bs \lambda_c}^{(n)}$, $\bs \sigma_{\bs \lambda_c}^{(n)}$ can be computed using the chain rule as
\begin{small}
\begin{align}
    \frac{\pa}{\pa \bs \mu_{\bs \lambda_c}^{(n)}}:~ &\left< \frac{\pa \bs \lambda_c^{(n)}}{\pa \bs \mu_{\bs \lambda_c}^{(n)}}\frac{\pa \bs u_c(\bs \lambda_c^{(n)})}{\pa \bs \lambda_c^{(n)}}\frac{\pa}{\pa \bs u_c}\log \mathcal N(\bs u_f^{(n)}| \bs W \bs u_c(\bs \lambda_c^{(n)}), \tx{diag}[\bs \tau_{cf}]^{-1}) \right>_{p(\bs \epsilon^{(n)})} \\ \nonumber
&+ \sum_{m = 1}^{\tx{dim}(\bs \lambda_c)} \left<\frac{\pa \lambda_{c, m}^{(n)}}{\pa \bs \mu_{\bs \lambda_c}^{(n)}}\frac{\pa}{\pa \lambda_{c, m}^{(n)}} \log \mathcal N(\lambda_{c, m}^{(n)}|\tilde{\bs \theta}_{c, m}^T \bs\varphi_m(\bs \lambda_f^{(n)}), \tau_{c, m}^{-1}) \right>_{p(\bs \epsilon^{(n)})} \vphantom{\sum_{m = 1}^{\tx{dim}(\bs \lambda_c)}}, \\
\frac{\pa}{\pa \bs \sigma_{\bs \lambda_c}^{(n)}}:~ &\left< \frac{\pa \bs \lambda_c^{(n)}}{\pa \bs \sigma_{\bs \lambda_c}^{(n)}}\frac{\pa \bs u_c(\bs \lambda_c^{(n)})}{\pa \bs \lambda_c^{(n)}}\frac{\pa}{\pa \bs u_c}\log \mathcal N(\bs u_f^{(n)}| \bs W \bs u_c(\bs \lambda_c^{(n)}), \tx{diag}[\bs \tau_{cf}]^{-1}) \right>_{p(\bs \epsilon^{(n)})}  \\ \nonumber
&+ \sum_{m = 1}^{\tx{dim}(\bs \lambda_c)} \left<\frac{\pa \lambda_{c, m}^{(n)}}{\pa \bs \sigma_{\bs \lambda_c}^{(n)}}\frac{\pa}{\pa \lambda_{c, m}^{(n)}} \log \mathcal N(\lambda_{c, m}^{(n)}|\tilde{\bs \theta}_{c, m}^T \bs\varphi_m(\bs \lambda_f^{(n)}), \tau_{c, m}^{-1}) \right>_{p(\bs \epsilon^{(n)})} \vphantom{\sum_{m = 1}^{\tx{dim}(\bs \lambda_c)}} \\
&+ (\bs \sigma_{\bs \lambda_c}^{(n)})^{-1},
\end{align}
\end{small}
where $\pa \bs \lambda_c^{(n)}/\pa \bs \mu_{\bs \lambda_c}^{(n)} = \bs I$ and $\pa \bs \lambda_c^{(n)}/\pa \bs \sigma_{\bs \lambda_c}^{(n)} = \tx{diag}[\bs \epsilon^{(n)}]$. To solve the optimization problem in \eqref{eq:VIopt}, the above gradients are passed to the \tit{adaptive moment estimation} optimizer (ADAM) using the default optimization parameters suggested in \cite{Kingma2014}.

\section{Feature functions}
\label{sec:features}
\newgeometry{left=12mm, right=12mm}
\begin{table}[h!]
\begin{center}
\tbf{Feature functions $\varphi$}
\vspace{1mm}

{\fontsize{5.5}{.2} \selectfont
\begin{tabular}{r|l|c}
\tbf{Index} $j$ & \tbf{Function} $\varphi_{jm}$ & \tbf{Comment} \\
\hline
1 & constant & $\varphi_j = 1$ \\
\hline
2 & pore fraction in $\Omega$ & pore fraction evaluated on full domain $\Omega$ \\
\hline
3 & $\log$ pore fraction in $\Omega$ & \\
\hline
4--7 & (pore fraction)$^{0.5\ldots 0.5\ldots 2.5}$ in $\Omega$ & \\
\hline
8 & $\exp$(pore fraction) & \\
\hline
9 &$\log$ SCA in $\Omega$ & $\log$ self-cons. approx. \cite{Torquato2001}, sec. 18.1.2, inf. contrast limit \\
\hline
10 & Maxwell-Approximation in $\Omega$ & inf. contrast limit, see \cite{Torquato2001}, sec. 18.2.1 \\
\hline
11 & $\log$ Maxwell-Approximation & \\
\hline
12--17 & $\log$ chord length density in $\Omega$, $d = 0.05\ldots 0$ & see \cite{Torquato2001} sec. 6.2.4 \\
\hline
18 & interface area in $\Omega$ &  \\
\hline
19 & $\log$ interface area in $\Omega$ \\
\hline
20--27 & $|\log$ interface area$|^{3/2, 1/2, 1/3, 1/4, 1/5, 2, 3, 4}$ in $\Omega$ \\
\hline
28--31 & (interface area)$^{1/3, 1/4, 1/5, 2}$ in $\Omega$ & \\
\hline
32 & mean distance edge in $\Omega$ & measured from excl. edge to edge \\
\hline
33 & $\log$ mean distance edge in $\Omega$ & measured from excl. edge to edge \\
\hline
34 & $\log^2$ mean distance edge in $\Omega$ & measured from excl. edge to edge \\
\hline
35 & $\log^3$ mean distance edge in $\Omega$ & measured from excl. edge to edge \\
\hline
36 & mean distance center in $\Omega$ & measured from excl. center to center \\
\hline
37 & min. distance center in $\Omega$ & measured from excl. center to center \\
\hline
38 & $\log$ min. distance center in $\Omega$ & measured from excl. center to center \\
\hline
39 & $\log^2$ min. distance center in $\Omega$ & measured from excl. center to center \\
\hline
40--43 & lineal path in $\Omega$ for $d = 0.025, 0.01, 0.005, 0.002$ & see \cite{Torquato2001}, sec. 2.4 \\
\hline
44--47 & $\log$ lineal path in $\Omega$ for $d = 0.025, 0.01, 0.005, 0.002$ & \\
\hline
48 & void nearest-neighbor pdf, $d = 0$, in $\Omega$ & see \cite{Torquato2001}, sec. 2.8 \\
\hline
49 & $\log$ void nearest-neighbor pdf, $d = 0$, in $\Omega$ & see \cite{Torquato2001}. sec. 2.8 \\
\hline
50 & pore size density, $d = 0$, in $\Omega$ & see \cite{Torquato2001}, sec. 2.6 \\
\hline
51 & $\log$ pore size density, $d = 0$, in $\Omega$ & see \cite{Torquato2001}, sec. 2.6 \\
\hline
52 & mean chord length  in $\Omega$  & see \cite{Torquato2001}, sec. 2.5 \\
\hline
53 & $\log$ mean chord length in $\Omega$ & see \cite{Torquato2001}, sec. 2.5 \\
\hline
54 & $\exp$ mean chord length in $\Omega$ & see \cite{Torquato2001}, sec. 2.5 \\
\hline
55 & (mean chord length)$^{0.5}$ in $\Omega$ & see \cite{Torquato2001}, sec. 2.5 \\
\hline
56--58 & $\left<r_{ex}^{0.2, 0.5, 1} \right>$ in $\Omega_m$ & expected exclusion radii moments \\
\hline
59--61 & $\log\left<r_{ex}^{0.2, 0.5, 1} \right>$ in $\Omega_m$ & $\log$ expected exclusion radii moments \\
\hline
62 & pore fraction in $\Omega_m$ \\
\hline
63 & $\log$ pore fraction in $\Omega_m$ \\
\hline
64 & $\exp$ pore fraction in $\Omega_m$ \\
\hline
65--68 & (pore fraction)$^{0.5, 1.5, 2, 2.5}$ in $\Omega_m$ \\
\hline
69 & $\log$ self-consistent approximation (inf. contrast) in $\Omega_m$ \\
\hline
70 & Maxwell Approximation in $\Omega_m$ \\
\hline
71 & $\log$ Maxwell Approximation in $\Omega_m$ \\
\hline
72--78 & \makecell{$\log$ chord length dens. in $\Omega_m$, \\ $d = (50, 25, 12.5, 6.25, 3, 1.5, 0)\cdot 10^{-4}$} \\
\hline
79 & interface area in $\Omega_m$ \\
\hline
80--88 & $|\log$ interface area$|^{1, 1.5, 2, 3, 4, 5, 1/2, 1/3, 1/4}$ in $\Omega_m$ \\
\hline
89--93 & $|$ interface area$|^{1/2, 1/3, 1/4, 1/5, 2}$ in $\Omega_m$ \\
\hline
94 & mean distance edge in $\Omega_m$ & measured from excl. edge to edge\\
\hline
95 & $\log$ mean distance edge in $\Omega_m$ & measured from excl. edge to edge\\
\hline
96 & max distance edge in $\Omega_m$ & measured from excl. edge to edge\\
\hline
97 & $\log$ max distance edge in $\Omega_m$ & measured from excl. edge to edge\\
\hline
98 & std distance edge in $\Omega_m$ & measured from excl. edge to edge\\
\hline
99 & $\log$ std distance edge in $\Omega_m$ & measured from excl. edge to edge\\
\hline
100--104 & square well potential, width = $(1, 2, 3, 4, 5)\cdot 10^{-2}$ in $\Omega_m$ & \\
\hline
105 & $\log^2$ mean distance  in $\Omega_m$ & measured from excl. edge to edge \\
\hline
106 & $\log^3$ mean distance in $\Omega_m$ & measured from excl. edge to edge \\
\hline
107 & mean distance center in $\Omega_m$ & measured from excl. center to center \\
\hline
108 & min. distance center in $\Omega_m$ & measured from excl. center to center \\
\hline
109 & $\log$ min. distance center in $\Omega_m$ & measured from excl. center to center \\
\hline
110 & $\log^2$ min. distance center in $\Omega_m$ & measured from excl. center to center \\
\hline
111--114 & lineal path in $\Omega_m$  for $d = 0.025, 0.01, 0.005, 0.002$ &  \\
\hline
115--118 & $\log$ lineal path in $\Omega_m$  for $d = 0.025, 0.01, 0.005, 0.002$ &  \\
\hline
119 & void nearest-neighbor pdf, $d = 0$, in $\Omega_m$ & see \cite{Torquato2001}, sec. 2.8 \\
\hline
120 & $\log$ void nearest-neighbor pdf, $d = 0$, in $\Omega_m$ & see \cite{Torquato2001}. sec. 2.8 \\
\hline
121 & pore size density, $d = 0$, in $\Omega_m$ & see \cite{Torquato2001}, sec. 2.6 \\
\hline
122 & $\log$ pore size density, $d = 0$, in $\Omega_m$ & see \cite{Torquato2001}, sec. 2.6 \\
\hline
123 & mean chord length in $\Omega_m$ & see \cite{Torquato2001}, sec. 2.5 \\
\hline
124 & $\log$ mean chord length in $\Omega_m$  & see \cite{Torquato2001}, sec. 2.5 \\
\hline
125 & $\exp$ mean chord length in $\Omega_m$  & see \cite{Torquato2001}, sec. 2.5 \\
\hline
126 & (mean chord length)$^{0.5}$  in $\Omega_m$ & see \cite{Torquato2001}, sec. 2.5 \\
\hline
127--129 & $\left<r_{ex}^{0.2, 0.5, 1} \right>$ in $\Omega_m$ & expected exclusion radii moments \\
\hline
130--132 & $\log\left<r_{ex}^{0.2, 0.5, 1} \right>$ in $\Omega_m$ & $\log$ expected exclusion radii moments \\
\hline
133 & length scale of exp. approx. to lin. path in $\Omega_m$ &  \\
\hline
134 & mean of euclidean dist. transform in $\Omega_m$ & see \cite{Soille1999} \\
\hline
135 & variance of euclidean dist. transform in $\Omega_m$ & see \cite{Soille1999} \\
\hline
136 & max. of euclidean dist. transform  in $\Omega_m$& see \cite{Soille1999} \\
\hline
137 & mean of chessboard dist. transform  in $\Omega_m$& see \cite{Soille1999} \\
\hline
138 & variance of chessboard dist. transform in $\Omega_m$ & see \cite{Soille1999} \\
\hline
139 & max. of chessboard dist. transform  in $\Omega_m$& see \cite{Soille1999} \\
\hline
140 & mean of cityblock dist. transform in $\Omega_m$ & see \cite{Soille1999} \\
\hline
141 & variance of cityblock dist. transform in $\Omega_m$ & see \cite{Soille1999} \\
\hline
142 & max. of cityblock dist. transform  in $\Omega_m$& see \cite{Soille1999} \\
\hline
143--145 & Gauss lin. filt. $d = 2, 5, 10$ pixels in $\Omega_m$ & see \cite{Grigo2019} \\
\hline
146 & Ising energy in $\Omega_m$&  \\
\hline
147 & shortest connected path, $x$-dir., Euclidean, in $\Omega_m$ & see \cite{Soille1999} \\
\hline
148 & shortest connected path, $y$-dir., Euclidean, in $\Omega_m$ & see \cite{Soille1999} \\
\hline
149 & shortest connected path, $x$-dir., cityblock, in $\Omega_m$ & see \cite{Soille1999} \\
\hline
150 & shortest connected path, $y$-dir., cityblock, in $\Omega_m$ & see \cite{Soille1999}
\end{tabular}
}
\vspace{2mm}
\end{center}
\caption{Set of 150 feature functions $\varphi$ applied in the numerical examples of section \ref{sec:experiments}.}
\label{featureTable}
\end{table}
\restoregeometry

Table \ref{featureTable} shows a list of the 150 feature functions used in the numerical examples of section \ref{sec:experiments}. Features 1--55 take the whole microstructure $\bs \lambda_f$ as input, features 56--150 use the subset $\bs \lambda_f^{(m)}$ pertaining to subdomain/cell $\Omega_m$ for which $\bs K_m(\bs x, \bs \lambda_c) = e^{\lambda_{c, m}} \bs I$.

\section{Proper orthogonal decomposition of output data \texorpdfstring{$\bs u_f$}{}}
\label{sec:POD}
\begin{figure}[h]
\centering
\includegraphics[width=.8\textwidth]{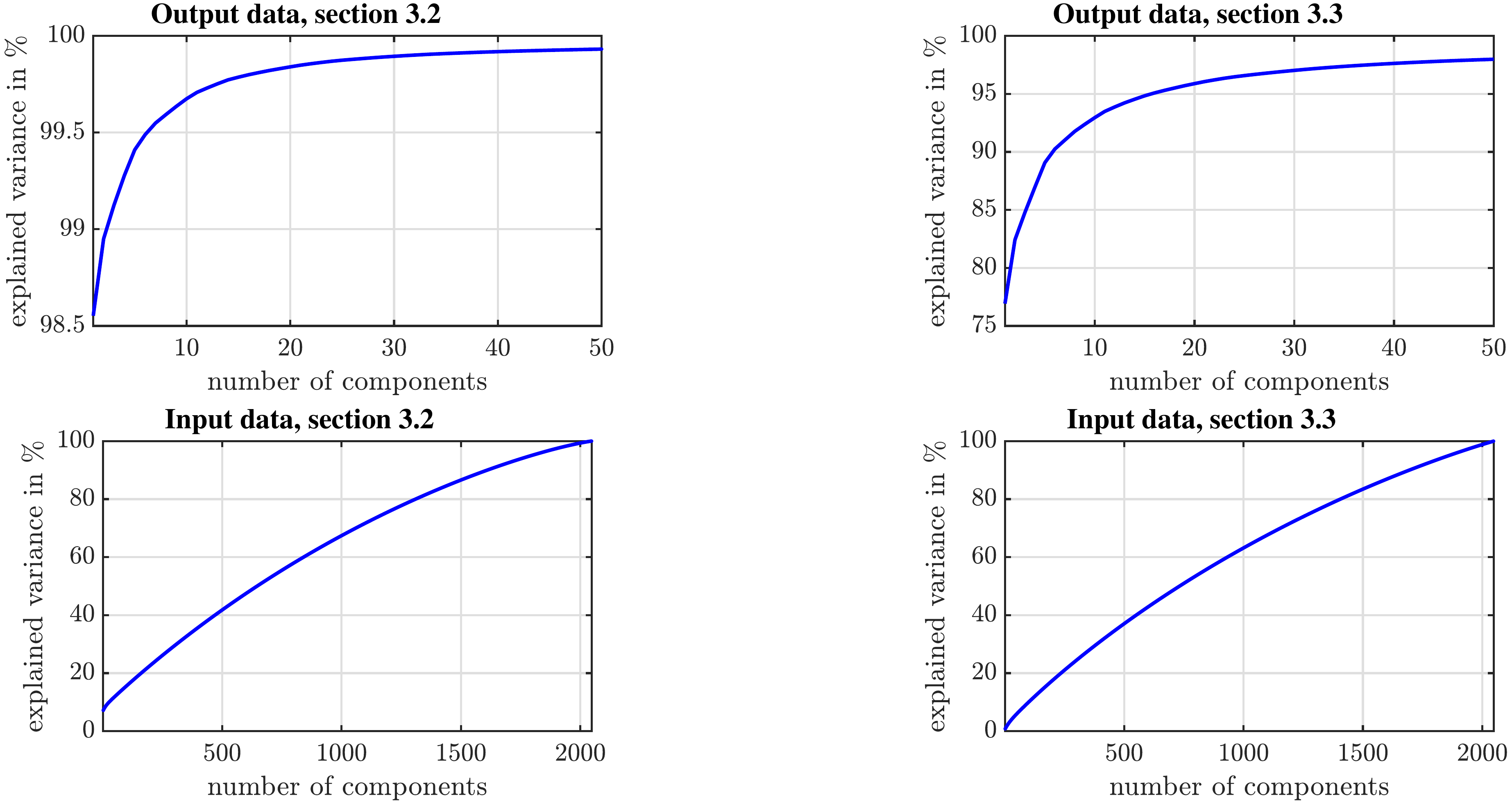}
\caption{Explained variance versus number of PCA components of the output $\bs u_f$ (top) and input $\bs \lambda_f$ (bottom) for the data used in sections \ref{sec:homLimit} (left) and \ref{sec:predPerformance} (right). All plots are based on a PCA analysis using 2048 random discretized input/output samples $\bs \lambda_f, \bs u_f$ drawn according to the distributions described in the text.}
\label{fig:POD}
\end{figure}
In section \ref{sec:homLimit}, it is mentioned that we observe very fast decay of POD eigenmodes for data with large variance in the pore space fraction $\tx{vol}(\Omega_f)/\tx{vol}(\Omega)$, which translates to high $\sigma_{\tx{ex}}^2$ as explained in section \ref{sec:data}. We therefore plot the \tit{explained PCA variance}, i.e. the cumulative sum of the $n$ largest variances of a PCA analysis performed on 2048 sample vectors $\mathcal D_{PCA} = \left\{\bs u_f(\bs \lambda_f^{(n)}) \right\}_{n = 1}^{2048}$ for the data used in sections \ref{sec:homLimit} and \ref{sec:predPerformance} in figure \ref{fig:POD}. We note that this fast decay in POD eigenmodes observed in section \ref{sec:homLimit} does \tbf{not} affect the effective input dimensionality of $\bs \lambda_f$. To show that, we discretize the 2048 microstructures from the distributions defined in sections \ref{sec:homLimit}, \ref{sec:predPerformance} to bitmap image vectors of $256\times 256 = 65,536$ dimensions and again plot the explained variance from the first $n$ components in the second row of figure \ref{fig:POD}. We observe that our claim of high-dimensional inputs is verified, as the explained variance is only slowly increasing with the number of POD components.

\section{Evidence Lower BOund (ELBO) \texorpdfstring{$\mathcal F(Q)$}{}}
\label{ap:elbo}
Using the expected values found in equations \eqref{eq:Qgamma}--\eqref{eq:expectedValues}, omitting constants and canceling terms, we find
\begin{equation}
\begin{split}
\mathcal F(Q) & = \left<\log p(\bs \theta, \mathcal D) - \log Q(\bs \theta)\right>  \\
& = - \tilde e \sum_{i = 1}^{\tx{dim}(\bs u_f)}\log \tilde f_i  +\sum_{m = 1}^{\tx{dim}(\bs \lambda_c)}  \sum_{n = 1}^N  \log \sigma_{\bs \lambda_c, m}^{(n)}   
 - \tilde c\sum_{m = 1}^{\tx{dim}(\bs \lambda_c)}\log \tilde d_m \\
 &    - \tilde a \sum_{m = 1}^{\tx{dim}(\bs \lambda_c)} \sum_{j = 1}^{\tx{dim}(\bs \gamma)}\log \tilde b_{jm} + \sum_{m = 1}^{\tx{dim}(\bs \lambda_c)}\sum_{j = 1}^{\tx{dim}(\bs \gamma)}\log \sigma_{\tilde{\theta}_{c, jm}}
\end{split}
\label{eq:elboComp}
\end{equation}
as the most compact form of the evidence lower bound (ELBO) $\mathcal F(Q)$. We note that with the exception of the first term, the remaining ones are automatically  separated into additive contributions from each cell $m$ or equivalently each $\lambda_{c,m}$.  Hence, in order to arrive at the cell-scoring function $\mathcal F_m(Q)$ given in Equation \eqref{eq:cellScore}, we pick out the terms in equation \eqref{eq:elboComp} which explicitly carry a cell index $m$. These are
\begin{equation*}
\mathcal F_m(Q) = \sum_{n = 1}^N \log \sigma_{\lambda_{c, m}}^{(n)} - \tilde{c} \log \tilde{d}_m  - \tilde a \sum_{j = 1}^{\tx{dim}(\bs \gamma)}\log \tilde b_{jm} + \sum_{j = 1}^{\tx{dim}(\bs \gamma)} \log \sigma_{\tilde{\theta}_{c, jm}}.
\end{equation*}
In the numerical experiments, we tie the precision parameters $\gamma_{jm}$ across different cells $m$ together i.e. $\gamma_{jm}=\gamma_{j}$ which allows sharing of information in terms of the microstructural features. As  a result $b_{jm}=b_j$ and the corresponding contribution from each cell $m$ becomes:
\be
\mathcal F_m(Q) = \sum_{n = 1}^N \log \sigma_{\lambda_{c, m}}^{(n)} - \tilde{c} \log \tilde{d}_m  + \sum_{j = 1}^{\tx{dim}(\bs \gamma)} \log \sigma_{\tilde{\theta}_{c, jm}}.
\ee

\bibliographystyle{model1-num-names}
\bibliography{refs}

\end{document}